\documentclass{article}

 \usepackage[main, final]{neurips_2026}

\usepackage[utf8]{inputenc} 
\usepackage[T1]{fontenc}    
\usepackage{hyperref}       
\usepackage{url}            
\usepackage{booktabs}       
\usepackage{amsfonts}       
\usepackage{nicefrac}       
\usepackage{microtype}      
\usepackage{xcolor}         

\usepackage{graphicx}
\usepackage[misc]{ifsym}

\usepackage{amsmath, amssymb}
\usepackage{caption}
\usepackage[justification=centering]{subcaption}
\usepackage{pifont}

\usepackage{multirow}
\usepackage{xr-hyper}
\usepackage{makeidx}  
\usepackage{algorithm}
\usepackage{algorithmic}
\usepackage{mathtools}
\usepackage{pdfpages}
\usepackage{pdflscape}
\usepackage{mathtools}
\usepackage{bbm}
\usepackage{lipsum}
\newcommand*\rot[1]{\rotatebox{90}{#1}}
\usepackage{mwe}

\title{Cross-Attentive Bayesian Low-Rank Adaptation for Multimodal Uncertainty Estimation}

\author{%
  Habibeh Naderi \\
  Faculty of Computer Science\\
  Dalhousie University\\
  \texttt{habibeh.naderi@dal.ca} \\
  \And
  Behrouz H. Soleimani \\
  Kinaxis \\
  Ottawa, ON \\
  \texttt{Behrouz@kinaxis.ca} \\
  \AND
  Stan Matwin \\
  Faculty of Computer Science\\
  Dalhousie University\\
  \texttt{stan@cs.dal.ca} \\
}

\begin{document}

\maketitle

\begin{abstract}
Parameter-efficient fine-tuning (PEFT) enables efficient adaptation of frozen language models, but most PEFT methods remain deterministic or unimodal, limiting their reliability in low-resource audio-text settings where uncertainty depends on both linguistic evidence and acoustic conditions. We introduce SPECTRA (Stochastic Posterior Estimation with Cross-modal Token-level Rank Adaptation), a multimodal Bayesian low-rank adaptation framework for uncertainty-aware audio-text learning. SPECTRA keeps the text and audio backbones frozen and confines stochasticity to a compact rank-$r$ latent matrix inside each LoRA adapter. At each transformer layer, text-derived low-rank token features query frame-level audio embeddings through lightweight cross-attention; the resulting token-specific acoustic context parameterizes the mean and variance of an amortized variational posterior over the adapter latent. This design treats audio not merely as an additional feature stream, but as a localized reliability signal that modulates both adaptation and confidence while preserving the scalability of PEFT. Posterior prediction is performed with Monte Carlo adapter samples, enabling a Bayesian uncertainty analysis that decomposes normalized predictive uncertainty into total, aleatoric, and adapter-space epistemic components and evaluates whether uncertainty distinguishes correct from misclassified predictions. Across IEMOCAP and clinical interview prediction tasks, SPECTRA is consistently competitive with or improves upon text-only Bayesian PEFT and conventional multimodal transfer-learning baselines, with token-level cross-attention yielding the most reliable gains. Additional modality-disagreement stress tests show that mismatched audio causes substantially larger degradation in AUC, likelihood, calibration, and Brier score than noisy but matched audio, highlighting the importance of modeling cross-modal reliability. These results suggest that Bayesian cross-modal conditioning in low-rank adapter space provides an efficient and principled mechanism for calibrated multimodal adaptation.

\end{abstract}

\section{Introduction}
\label{sec:introduction}

Pre-trained Large Language Models (LLMs) have become the dominant backbone for a wide range of prediction tasks \cite{llms_brown2020language,llms_minaee2024large}. However, adapting these large models to specialized domains remains challenging when labeled data are scarce and predictions must be reliable. This challenge is particularly pronounced in human-centered applications such as behavioral and clinical analysis, where downstream targets may include affect, cognitive state, and interaction quality. In such settings, models must simultaneously satisfy three requirements: (i) strong predictive performance under limited supervision, (ii) computationally efficient adaptation of large pretrained backbones, and (iii) reliable uncertainty estimates that reflect when model predictions should be trusted.

Parameter-efficient fine-tuning (PEFT) methods \cite{fu2023effectiveness_peft} address the first two requirements by adapting large models using small trainable modules while keeping the backbone weights frozen. Among these methods, Low-Rank Adaptation (LoRA) and related approaches introduce low-dimensional parameter updates that allow efficient adaptation with a small number of additional parameters \cite{hu2022lora,zhang2023adaptive_adalora,liu2022few_ia3}. PEFT techniques significantly reduce memory and training costs while often maintaining performance comparable to full fine-tuning. However, most PEFT approaches are deterministic and optimized primarily for predictive accuracy, which can lead to overconfident predictions in the presence of noisy inputs, domain shifts, or ambiguous examples \cite{uncertainty_leng2024taming,uncertainty_xiong2023can}. This limitation motivates uncertainty-aware PEFT methods that incorporate Bayesian principles to produce calibrated predictions \cite{wang2024blob,rahmati2025clora,laplacelora_yang2023bayesian}.

Bayesian variants of LoRA introduce uncertainty by placing probabilistic distributions over adapter parameters while keeping the backbone deterministic. Bayesian LoRA by Backpropagation (BLoB) \cite{wang2024blob} models epistemic uncertainty in LoRA weights using variational inference. More recently, Contextual LoRA (C-LoRA) \cite{rahmati2025clora} introduced input-dependent Bayesian adapters by conditioning the variational posterior on intermediate low-rank features. This contextualization produces heteroscedastic uncertainty, allowing the model to express varying confidence across different inputs while maintaining the computational efficiency of PEFT. Despite these advances, existing uncertainty-aware PEFT methods are largely unimodal and primarily condition uncertainty on internal language-model representations.

A second challenge arises when the predictive signal is inherently multimodal. Many real-world prediction tasks involve both lexical content and acoustic information. In speech-centered applications, text often provides strong semantic cues while audio carries complementary signals related to prosody, speaker state, and environmental conditions \cite{audiolm_borsos2023audiolm}. Multimodal transfer learning studies consistently show that these modalities can provide complementary evidence, especially when one modality becomes unreliable. However, incorporating multimodal information introduces additional modeling complexity. Naive feature fusion strategies may increase representational dimensionality, amplify over-parameterization in low-resource settings, and complicate calibration when modalities disagree. These issues become particularly critical when the reliability of one modality varies across time, such as when audio segments contain noise, silence, or variable recording conditions.

In many speech-based tasks, uncertainty is driven not only by the linguistic content but also by external acoustic conditions. For example, background noise, channel variability, or speaker-specific factors may reduce the reliability of audio segments that correspond to otherwise informative text tokens. Existing uncertainty-aware PEFT approaches do not explicitly incorporate such cross-modal reliability signals. Consequently, they may fail to adjust predictive confidence when acoustic evidence contradicts or weakens the textual signal.

\begin{figure}[t]
\centering
\makebox[\textwidth][c]{%
\includegraphics[width=\textwidth]{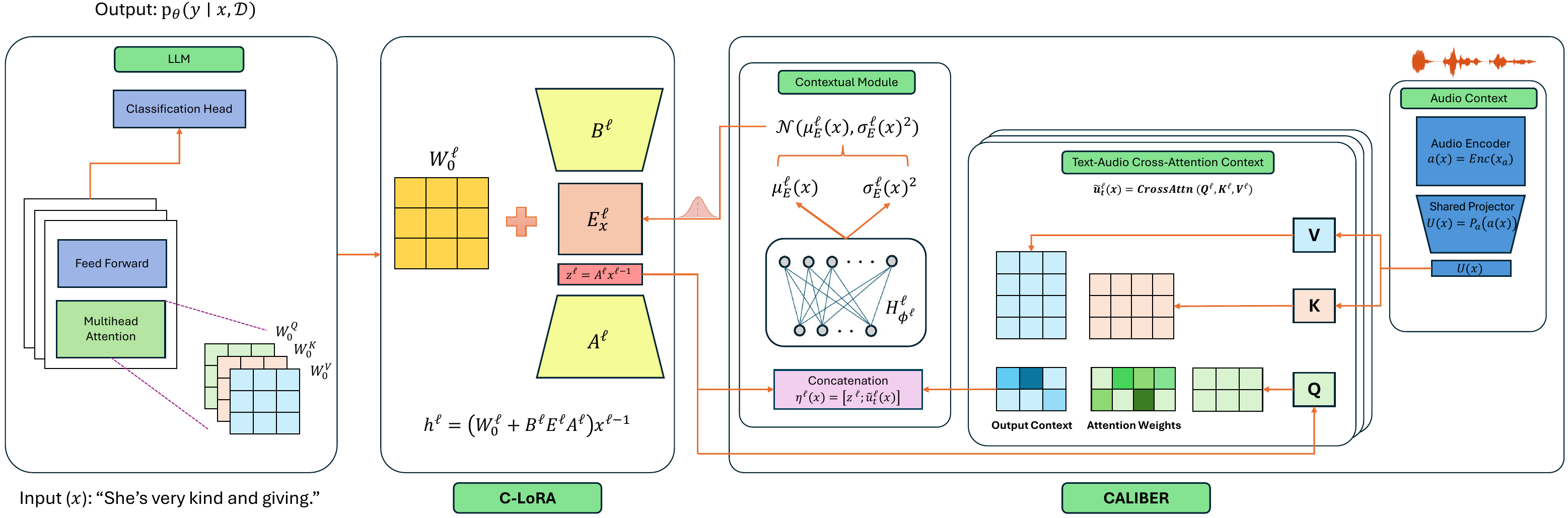}
}
\caption{\label{fig_architecture}Overview of the proposed SPECTRA architecture. Per-layer text-audio cross-attention conditions the variational adapter distribution, enabling uncertainty-aware multimodal low-rank adaptation. A frozen text backbone produces token-level hidden states that enter LoRA adapters at each transformer layer, while a frozen audio encoder produces frame-level acoustic embeddings. The audio frames are first projected into a shared context space and then consumed by lightweight per-layer cross-attention modules, where LoRA text features act as queries and audio frames serve as keys and values. The resulting token-conditioned cross-modal context is concatenated with the local low-rank feature to parameterize a variational posterior over the compact latent matrix $E^\ell_{x,t}$. Samples from this posterior modulate the low-rank update $B^\ell E^\ell_{x,t} A^\ell$, enabling context-aware Bayesian adaptation with calibrated uncertainty while preserving the efficiency of PEFT.}
\end{figure}

In this work, we introduce Stochastic Posterior Estimation with Cross-modal Token-level Rank Adaptation (SPECTRA), a multimodal uncertainty-aware PEFT framework that integrates token-level cross-modal context into Bayesian low-rank adaptation (Figure~\ref{fig_architecture}). SPECTRA extends contextual Bayesian adapters by conditioning the variational posterior over the low-rank latent space on cross-attention between text tokens and frame-level audio representations. Each transformer layer forms a token-specific view of the audio sequence through lightweight cross-attention modules. The resulting cross-modal context acts as a reliability signal that modulates both the mean and variance of the adapter posterior.

Conceptually, SPECTRA combines three complementary research directions. From the perspective of PEFT, the method preserves the scalability advantages of LoRA by restricting trainable parameters to lightweight adapters and keeping the backbone frozen. From the uncertainty modeling perspective, SPECTRA extends contextual Bayesian LoRA approaches by conditioning the posterior distribution on multimodal evidence, enabling uncertainty estimates that respond to cross-modal interactions. From the multimodal learning perspective, SPECTRA introduces a lightweight alternative to high-dimensional feature fusion: instead of merging modalities directly in the representation space, audio information influences the distribution over adapter behavior in the low-rank space. This design allows the model to dynamically adjust its confidence based on temporally localized acoustic cues while maintaining parameter efficiency.

Another key design principle of SPECTRA is to treat audio as a contextual reliability signal rather than a primary feature stream. Each transformer layer computes token-conditioned cross-attention over the audio frames, enabling depth-specific alignment between text tokens and acoustic segments. These cross-attended features are then used to parameterize the variational posterior of a compact latent matrix within the low-rank adapter space. Because stochasticity is confined to this low-dimensional latent component, the Bayesian complexity scales with the adapter rank rather than the full backbone dimension, making the approach practical for large transformer models.

Overall, SPECTRA aims to address a fundamental problem in multimodal low-resource prediction: how to adapt large language models efficiently while producing uncertainty estimates that respond to temporally localized cross-modal reliability cues. Our contributions can be summarized as follows:

\begin{itemize}
    \item We propose SPECTRA, a multimodal, uncertainty-aware PEFT framework that conditions adapter uncertainty on external audio evidence in addition to internal low-rank text features, yielding cross-modal, heteroscedastic uncertainty.

    \item We introduce a per-layer, token-level text-audio cross-attention mechanism that allows the variational posterior to depend on temporally localized acoustic information, enabling depth-specific and token-specific cross-modal conditioning without heavy multimodal fusion.

    \item We preserve lightweight Bayesian complexity by restricting stochasticity to a compact latent component in the rank-$r$ space, maintaining the scalability advantages of contextual LoRA-style Bayesianization.

    \item We provide a unified perspective connecting PEFT, uncertainty-aware LoRA (e.g., BLoB and C-LoRA), and multimodal transfer learning, showing how cross-modal context can be incorporated as a principled uncertainty signal rather than only as a feature-fusion signal.
\end{itemize}

\section{Proposed Method}
\label{sec:caliber}

\subsection{Preliminaries: Low-Rank and Bayesian Adapter Tuning}

\paragraph{Low-rank adaptation.}
Low-Rank Adaptation (LoRA) adapts a frozen linear transformation $W_0\in\mathbb{R}^{d\times k}$ by learning a low-rank update rather than modifying all entries of $W_0$ \cite{hu2022lora}. For an input $x\in\mathbb{R}^{k}$, the adapted forward pass is
{\small
\begin{equation}
h=(W_0+\Delta W)x=(W_0+BA)x,
\qquad
B\in\mathbb{R}^{d\times r},\; A\in\mathbb{R}^{r\times k},\; r\ll \min(d,k).
\label{eq:lora_standard_caliber}
\end{equation}
}
Only $A$ and $B$ are trained, reducing the number of trainable parameters from $dk$ to $r(d+k)$ while keeping the pretrained backbone fixed. However, standard LoRA is deterministic and therefore provides only point estimates of the adapter parameters.
Bayesian LoRA methods address this limitation by placing a variational posterior over adapter parameters and optimizing an evidence lower bound (ELBO) \cite{wang2024blob}. Contextual LoRA (C-LoRA) further amortizes this posterior from input-dependent low-rank features, yielding heteroscedastic uncertainty while keeping Bayesian inference in a compact adapter subspace \cite{rahmati2025clora}. SPECTRA builds on this idea but conditions the adapter posterior on both text-derived low-rank features and token-level audio context.

\subsection{SPECTRA: Stochastic Posterior Estimation with Cross-modal Token-level Rank Adaptation}

SPECTRA is a multimodal uncertainty-aware PEFT framework for audio-text prediction. The text transformer and the audio encoder are kept frozen. Instead of fusing a high-dimensional audio representation with the final text embedding, SPECTRA uses audio as a contextual reliability signal that shapes the distribution over low-rank adapter behavior at each layer. Figure~\ref{fig_architecture} illustrates the architecture, and pseudocode is provided in Appendix~\ref{app:sec_supplementary}.

\paragraph{Setup and stochastic low-rank adapter.}
Let $\mathcal{D}=\{(x_i,y_i)\}_{i=1}^{N}$ be the training set, where each text input $x_i$ is paired with an audio sequence $a(x_i)$. Consider a frozen transformer with $L$ layers. In a linear sub-layer at layer $\ell$, let $W_0^\ell\in\mathbb{R}^{d\times k}$ be the frozen weight, $x_t^{\ell-1}\in\mathbb{R}^{k}$ the hidden state at token position $t\in[T_x]$, and $h_t^\ell\in\mathbb{R}^{d}$ the output. SPECTRA uses deterministic LoRA factors $A^\ell\in\mathbb{R}^{r\times k}$ and $B^\ell\in\mathbb{R}^{d\times r}$, but inserts a stochastic latent matrix $E_{x,t}^\ell\in\mathbb{R}^{r\times r}$ in the rank-$r$ space:
{\small
\begin{equation}
z_t^\ell=A^\ell x_t^{\ell-1},
\qquad
h_t^\ell
=
W_0^\ell x_t^{\ell-1}+B^\ell E_{x,t}^\ell z_t^\ell,
\qquad
\Delta W_t^\ell(x)=B^\ell E_{x,t}^\ell A^\ell .
\label{eq:caliber_forward}
\end{equation}
}
Thus, stochasticity is confined to $E_{x,t}^\ell$, whose size is $r^2$ per token and layer, while $A^\ell$ and $B^\ell$ remain deterministic trainable parameters. This design preserves the efficiency of LoRA and avoids placing a posterior over the full backbone or over all entries of the LoRA factors.

\paragraph{Per-layer text-audio cross-attention.}
For each sample, a frozen audio encoder produces frame-level embeddings $a(x)=[a_1(x),\dots,a_{T_a}(x)]$, with $a_s(x)\in\mathbb{R}^{d_a}$. We first project each frame into a low-dimensional context space,
{\small
\begin{equation}
u_s(x)=P_a(a_s(x))\in\mathbb{R}^{c},
\qquad
U(x)=[u_1(x),\dots,u_{T_a}(x)]\in\mathbb{R}^{T_a\times c}.
\label{eq:caliber_audio_proj}
\end{equation}
}
At layer $\ell$, the local low-rank text feature $z_t^\ell$ queries the projected audio frames to obtain a token-specific acoustic context: $\tilde u_t^\ell(x)=\mathrm{CrossAttn}^\ell(q_t^\ell,K^\ell(x),V^\ell(x))\in\mathbb{R}^{r}$. Let $q_t^\ell = W_Q^\ell z_t^\ell$, $K^\ell(x) = U(x)W_K^\ell$, and $V^\ell(x) = U(x)W_V^\ell$. Then: $\tilde u_t^\ell(x)
=
W_O^\ell
\left[
\mathrm{softmax}
\!\left(
\frac{q_t^\ell (K^\ell(x))^\top}{\sqrt{d_c}}
\right)
V^\ell(x)
\right]$

This per-layer cross-attention lets different transformer layers attend to different temporal regions of the audio. The resulting context is used only to parameterize the adapter posterior, so SPECTRA does not increase the dimensionality of the downstream classifier.

\paragraph{Context-conditioned variational posterior.}
SPECTRA models $E_{x,t}^\ell$ with an input-dependent diagonal Gaussian posterior. The posterior is conditioned on both the local text adapter feature and the token-specific audio context:
{\small
\begin{align}
\eta_t^\ell(x) &= [z_t^\ell;\tilde u_t^\ell(x)]\in\mathbb{R}^{2r},
\label{eq:caliber_concat}\\
(\mu_{E,t}^\ell(x),\log v_{E,t}^\ell(x)) &= H_{\phi^\ell}^\ell(\eta_t^\ell(x)),
\qquad
H_{\phi^\ell}^\ell:\mathbb{R}^{2r}\rightarrow\mathbb{R}^{2r^2},\nonumber\\
q_{\phi^\ell}(\mathrm{vec}(E_{x,t}^\ell)\mid x)
&=
\mathcal{N}\!\left(
\mu_{E,t}^\ell(x),
\mathrm{diag}\big(\sigma_{E,t}^\ell(x)^2\big)
\right),
\label{eq:caliber_q}\\
\sigma_{E,t}^\ell(x)
&=
\varepsilon\,\mathrm{softplus}\!\left(\log v_{E,t}^\ell(x)\right)+\delta,
\qquad
\varepsilon>0,
\;\delta\approx 10^{-6}.
\label{eq:caliber_sigma}
\end{align}
}
Because both the posterior mean and variance depend on $\eta_t^\ell(x)$, SPECTRA induces heteroscedastic uncertainty: the same text activation can lead to different adapter uncertainty depending on the acoustic frames attended by that token.

\paragraph{Reparameterized adapter sampling.}
During training and Monte Carlo inference, we sample $E_{x,t}^\ell$ using the reparameterization trick:
{\small
\begin{align}
\mathrm{vec}(E_{x,t}^\ell)
&=
\mu_{E,t}^\ell(x)+\sigma_{E,t}^\ell(x)\odot \xi_t^\ell,
\qquad
\xi_t^\ell\sim\mathcal{N}(0,I_{r^2}),
\label{eq:caliber_reparam}\\
\Delta h_t^\ell(x)
&=
B^\ell E_{x,t}^\ell z_t^\ell
=
B^\ell E_{x,t}^\ell A^\ell x_t^{\ell-1}.
\label{eq:caliber_delta_h}
\end{align}
}
Thus, each stochastic draw changes the effective low-rank update while keeping the pretrained backbone fixed. The Bayesian overhead scales with $O(r^2)$ in the latent adapter space rather than with the full hidden dimension of the transformer.

\paragraph{Prior and training objective.}
We use a factorized Gaussian prior over the compact latent matrix,
{\small
\begin{equation}
p(\mathrm{vec}(E^\ell))=\mathcal{N}(0,\beta^2 I_{r^2}),
\label{eq:caliber_prior}
\end{equation}
}
where $\beta$ controls the prior scale. Let $\theta$ denote the deterministic trainable parameters, including $\{A^\ell,B^\ell\}_{\ell=1}^L$, $P_a$, the cross-attention projections, and the task head, and let $\phi=\{\phi^\ell\}_{\ell=1}^{L}$ denote the variational inference heads. SPECTRA maximizes
{\small
\begin{equation}
\mathcal{L}(\theta,\phi)
=
\sum_{i=1}^{N}
\left[
\mathbb{E}_{q_\phi(E_{x_i}\mid x_i)}
\log p_\theta(y_i\mid x_i,E_{x_i})
-
\gamma
\sum_{\ell=1}^{L}
\frac{1}{T_{x_i}}
\sum_{t=1}^{T_{x_i}}
\mathrm{KL}\!\left(
q_{\phi^\ell}(E_{x_i,t}^\ell\mid x_i)\,\|\,p(E^\ell)
\right)
\right].
\label{eq:caliber_elbo}
\end{equation}
}
The KL term is averaged over token positions so that its magnitude does not grow trivially with sequence length. In practice, we minimize the negative ELBO and use one Monte Carlo draw per training example,
{\small
\begin{equation}
\mathbb{E}_{q_\phi(E_{x_i}\mid x_i)}
\log p_\theta(y_i\mid x_i,E_{x_i})
\approx
\log p_\theta(y_i\mid x_i,E_{x_i}^{(1)}),
\qquad
E_{x_i}^{(1)}\sim q_\phi(E_{x_i}\mid x_i).
\label{eq:caliber_mc_train}
\end{equation}
}
Only the adapters, cross-attention modules, audio projection, inference heads, and classifier are updated; the text and audio backbones remain frozen.

\paragraph{Posterior prediction and uncertainty scores.}
At test time, SPECTRA can use either posterior means or Monte Carlo model averaging. For a test input $x^\ast$, the posterior predictive distribution is approximated as
{\small
\begin{equation}
p(y^\ast\mid x^\ast,\mathcal{D})
\approx
\frac{1}{M}\sum_{m=1}^{M}
p_\theta(y^\ast\mid x^\ast,E_{x^\ast}^{(m)}),
\qquad
E_{x^\ast}^{(m)}\sim q_\phi(E_{x^\ast}\mid x^\ast).
\label{eq:caliber_ppd}
\end{equation}
}
For classification with $C$ classes, let $\mathbf{p}_i^{(m)}=p_\theta(y_i\mid x_i,E_i^{(m)})$ and $\bar{\mathbf{p}}_i=M^{-1}\sum_{m=1}^{M}\mathbf{p}_i^{(m)}$. We report total predictive uncertainty, aleatoric uncertainty, and adapter-space epistemic uncertainty as
{\small
\begin{equation}
U_{\mathrm{total}}(x_i)=H[\bar{\mathbf{p}}_i],
\qquad
U_{\mathrm{alea}}(x_i)=\frac{1}{M}\sum_{m=1}^{M}H[\mathbf{p}_i^{(m)}],
\qquad
U_{\mathrm{epi}}(x_i)=U_{\mathrm{total}}(x_i)-U_{\mathrm{alea}}(x_i),
\label{eq:caliber_uncertainty}
\end{equation}
}
where $H[\mathbf{p}]=-\sum_{c=1}^{C}p_c\log(p_c+\epsilon_h)$. Since our tasks include both binary and multi-class labels, we use normalized scores $\widetilde{U}_{b}(x_i)=U_b(x_i)/\log C$ for $b\in\{\mathrm{total},\mathrm{alea},\mathrm{epi}\}$. In SPECTRA, $U_{\mathrm{epi}}$ measures disagreement over the stochastic adapter latent rather than uncertainty over all backbone parameters; it is therefore best interpreted as adapter-space epistemic uncertainty. High epistemic uncertainty indicates that different context-conditioned adapter samples induce different predictions, while high aleatoric uncertainty indicates that individual sampled adapters remain uncertain.

\paragraph{Relation to prior work.}
SPECTRA differs from BLoB by using an amortized input-dependent posterior instead of a fixed posterior over adapter weights. It differs from C-LoRA by conditioning this posterior not only on internal low-rank text features but also on per-layer, token-level audio context. Consequently, SPECTRA preserves the lightweight Bayesian structure of contextual LoRA while allowing uncertainty and adaptation to respond to temporally localized cross-modal evidence.

\section{Dataset}
\label{sec_data}

\begin{table}[tp]
\centering
\caption{Dataset information for the parent and offspring data.}
\label{tbl_dataset_info}
\setlength\tabcolsep{2pt}
\scalebox{0.8}{
\begin{tabular}{llcccccccc}
\toprule
 & \textbf{Task} & \textbf{\# Samples} & \textbf{\# Classes} & \textbf{Imbalanced (\%)} & \textbf{Label 0} & \textbf{Label 1} & \textbf{Label 2} & \textbf{Label 3} \\
\midrule
Parent & objective & 23693 & 2 & 19.08 & 13096 & 10597 & - & - \\
Parent & sentiment & 23479 & 3 & 57.39 & 4725 & 11088 & 7666 & - \\
Parent & anger & 23694 & 2 & 92.22 & 21984 & 1710 & - & - \\
Parent & fear & 23694 & 2 & 95.29 & 22628 & 1066 & - & - \\
Parent & joy & 23690 & 2 & 66.58 & 17756 & 5934 & - & - \\
Parent & sadness & 23694 & 2 & 92.70 & 22081 & 1613 & - & - \\
Parent & neutral & 23693 & 2 & 16.28 & 10797 & 12896 & - & - \\
Parent & cohesion & 23694 & 2 & 81.32 & 19964 & 3730 & - & - \\
Parent & rumination & 23694 & 2 & 98.95 & 23447 & 247 & - & - \\
Parent & overinclusive & 23694 & 2 & 97.20 & 23049 & 645 & - & - \\
Parent & worry & 23694 & 2 & 91.76 & 21890 & 1804 & - & - \\
Parent & criticism & 23694 & 2 & 88.42 & 21235 & 2459 & - & - \\
\midrule
Offspring & objective & 10319 & 2 & 1.90 & 5209 & 5110 & - & - \\
Offspring & sentiment & 10222 & 3 & 52.51 & 2404 & 5062 & 2756 & - \\
Offspring & richness & 10273 & 3 & 97.08 & 1366 & 8654 & 253 & - \\
Offspring & reference & 10328 & 4 & 77.89 & 2072 & 4970 & 2187 & 1099 \\
Offspring & irrelevance & 10301 & 2 & 87.98 & 9196 & 1105 & - & - \\
Offspring & anger & 10329 & 2 & 95.55 & 9889 & 440 & - & - \\
Offspring & fear & 10329 & 2 & 99.04 & 10231 & 98 & - & - \\
Offspring & joy & 10329 & 2 & 94.49 & 9790 & 539 & - & - \\
Offspring & sadness & 10329 & 2 & 94.62 & 9802 & 527 & - & - \\
Offspring & neutral & 10328 & 2 & 80.92 & 1655 & 8673 & - & - \\
Offspring & coherence & 10329 & 2 & 68.35 & 7846 & 2483 & - & - \\
Offspring & rumination & 10329 & 2 & 99.01 & 10228 & 101 & - & - \\
Offspring & worry & 10329 & 2 & 99.13 & 10240 & 89 & - & - \\
Offspring & anxiousness & 10317 & 2 & 96.51 & 9969 & 348 & - & - \\
Offspring & aggression & 2182 & 2 & 99.49 & 2171 & 11 & - & - \\
Offspring & criticism & 8090 & 2 & 95.04 & 7708 & 382 & - & - \\
Offspring & self-criticism & 10329 & 2 & 96.77 & 10006 & 323 & - & - \\
\bottomrule
\end{tabular}
}
\end{table}

\paragraph{FORBOW.} The data used in this work consists of audio speech samples from 369 subjects participating in the Families Overcoming Risks and Building Opportunities for Well Being (FORBOW) research project \cite{forbow_uher2014familial}. Participants are parents, 266 mothers and 103 fathers, in the age range of 28-51 years. In these clinical interviews, parents were asked to talk about their children for five minutes without interruption. Out of these subjects, 149 were diagnosed with Major Depressive Disorder (MDD), 66 were diagnosed with Bipolarity Disorder (BD), 19 were diagnosed with Schizophrenia, and 129 were the control group with no major mood disorders. In addition to the parents' interview files, FORBOW research project collected interviews with the children themselves. The audio interviews of children consists of 3 parts: 1) a three minute interview where children talk about themselves, 2) a two minute interview talking about a positive experience they had, and 3) a two minute interview where they talk about a negative experience they had. All three interviews are uninterrupted with a total of 7 minutes of speech from each child. We transcribed and broke down each sample into multiple segments based on changes in emotion, sentiment, objectivity/subjectivity, etc. Average word count in a segment is 17 and average audio length for a segment is 6.47 seconds.Table \ref{tbl_dataset_info} summarizes the dataset statistics, including the set of prediction tasks at different levels of granularity (segment-level, document-level, and psychological and cognitive tasks), as well as the label distributions and the degree of class imbalance.




\paragraph{IEMOCAP.} In addition to our defined set of prediction tasks, we evaluate SPECTRA on IEMOCAP as the most widely used public dataset for emotion recognition \cite{busso2008iemocap}. The IEMOCAP corpus contains a total of 5 sessions and 10 different speakers, with each session being a conversation of two exclusive speakers. We follow the conventional evaluation protocol: merge `excited' with `happy' to better balance the size of each emotion class and drop the unbalanced emotion classes to leave the final four classes (neutral, happy, sad, angry) with a similar amount of data points and cross-validate on five folds of the standard splits.

\section{Experiments}
\label{sec_experiments}

We evaluate SPECTRA on emotion recognition and clinical interview prediction to test whether audio-conditioned Bayesian low-rank adaptation improves parameter-efficient multimodal learning. The main paper reports compact summaries of the experimental grid; full per-backbone, per-audio-encoder, per-task results are provided in Appendix~\ref{app:full_results}.

\paragraph{Protocol.}
We use six frozen text encoders, nliRoBERTa, nliDistilRoBERTa, allRoBERTaLarge, allDistilRoBERTa, allMiniLM12, and paraTinyBERT, and three frozen audio encoders, whisperMedium, wav2vec2LargeFT, and hubertLargeFT. We compare three method families: (i) text-only PEFT using LoRA and C-LoRA, (ii) multimodal transfer learning, which fuses a pooled audio embedding with the text representation and trains a lightweight MLP head, and (iii) SPECTRA, which uses audio as a context signal for Bayesian low-rank adaptation. SPECTRA-G conditions the adapters on a global audio embedding, while SPECTRA-X uses token-to-frame cross-modal conditioning. On IEMOCAP we also evaluate a Shared-KV version of SPECTRA-X, where audio key/value projections are shared across layers to reduce overhead.

All PEFT methods use rank $r=8$ and scaling $\alpha=32$. Text and audio backbones are frozen; only adapter and task-head parameters are trained. Models are optimized for 50 epochs with AdamW using learning rate $10^{-3}$ and weight decay $10^{-3}$. For the Bayesian adapter, we use $\beta=0.2$, $\varepsilon=0.05$, and $\gamma=0.008$, and estimate predictive uncertainty with $M=10$ Monte Carlo forward passes. All clinical experiments use 5-fold cross-validation with speaker-level separation. We report AUC (\%) for predictive performance because the clinical labels are imbalanced.

\subsection{Main Predictive Results}
\label{sec:exp_results}

\begin{table}[tp]
\centering
\caption{Unimodal emotion recognition results on IEMOCAP (AUC \%).}
\label{tbl_iemocap_unimodal_auc}
\setlength\tabcolsep{3.0pt}
\scalebox{0.95}{
\begin{tabular}{llccc}
\toprule
Modality & Model & Transfer & LoRA & C-LoRA \\
\midrule
\multirow[t]{6}{*}{Text} & nliRoBERTa & 86.58$\pm$0.46 & 87.67$\pm$0.62 & \textbf{88.08$\pm$0.50} \\
 & nliDistilRoBERTa & 86.35$\pm$0.63 & 87.33$\pm$0.71 & \textbf{87.38$\pm$1.12} \\
 & allRoBERTaLarge & 86.88$\pm$0.64 & 84.12$\pm$0.54 & \textbf{88.40$\pm$1.07} \\
 & allDistilRoBERTa & 85.64$\pm$0.57 & 87.28$\pm$0.49 & \textbf{87.89$\pm$0.60} \\
 & allMiniLM12 & 83.05$\pm$0.88 & 75.36$\pm$0.83 & \textbf{87.19$\pm$0.71} \\
 & paraTinyBERT & 84.69$\pm$0.36 & \textbf{86.24$\pm$0.76} & 86.13$\pm$1.06 \\
\cline{1-5}
\multirow[t]{2}{*}{Audio} & whisperMedium & 93.00$\pm$0.44 & - & - \\
 & hubertLargeFT & 87.86$\pm$0.87 & - & - \\
\bottomrule
\end{tabular}
}
\end{table}

\begin{table}[tp]
\centering
\caption{Multimodal emotion recognition results on IEMOCAP (AUC \%).}
\label{tbl_iemocap_multimodal_auc}
\setlength\tabcolsep{3.0pt}
\scalebox{0.8}{
\begin{tabular}{llcccc}
\toprule
\multirow[c]{2}{*}{\textbf{Text Model}} & \multirow[c]{2}{*}{\textbf{Audio Model}} & \multirow[c]{2}{*}{\textbf{Transfer}} & \textbf{SPECTRA-G} & \multicolumn{2}{c}{\textbf{SPECTRA-X}} \\
 &  &  & \textbf{Audio Ctx} & \textbf{Shared-KV} & \textbf{Cross-Attn} \\
\midrule
\multirow[t]{2}{*}{nliRoBERTa} & whisperMedium & 93.40$\pm$0.63 & 93.27$\pm$0.60 & \textbf{94.01$\pm$0.53} & 93.05$\pm$0.99 \\
 & hubertLargeFT & 89.93$\pm$0.46 & 90.40$\pm$0.54 & 90.20$\pm$0.69 & \textbf{91.06$\pm$0.70} \\
\cline{1-6}
\multirow[t]{2}{*}{nliDistilRoBERTa} & whisperMedium & 93.39$\pm$0.75 & 92.91$\pm$0.64 & 93.15$\pm$0.19 & \textbf{93.99$\pm$0.95} \\
 & hubertLargeFT & \textbf{89.76$\pm$0.54} & 89.21$\pm$0.72 & 89.46$\pm$0.59 & 89.51$\pm$0.97 \\
\cline{1-6}
\multirow[t]{2}{*}{allRoBERTaLarge} & whisperMedium & \textbf{94.64$\pm$0.49} & 92.48$\pm$0.76 & 92.59$\pm$0.95 & 94.05$\pm$0.33 \\
 & hubertLargeFT & 88.06$\pm$0.14 & \textbf{89.21$\pm$1.67} & 88.49$\pm$2.18 & 88.81$\pm$1.39 \\
\cline{1-6}
\multirow[t]{2}{*}{allDistilRoBERTa} & whisperMedium & 94.48$\pm$0.38 & 92.56$\pm$0.81 & 93.14$\pm$0.76 & \textbf{94.68$\pm$0.58} \\
 & hubertLargeFT & 90.28$\pm$0.52 & 90.00$\pm$0.43 & 89.62$\pm$0.82 & \textbf{91.09$\pm$0.55} \\
\cline{1-6}
\multirow[t]{2}{*}{allMiniLM12} & whisperMedium & \textbf{93.32$\pm$0.57} & 92.93$\pm$0.47 & 92.84$\pm$0.64 & 92.99$\pm$0.67 \\
 & hubertLargeFT & 89.75$\pm$1.00 & 89.37$\pm$0.60 & \textbf{89.78$\pm$0.83} & 89.62$\pm$1.17 \\
\cline{1-6}
\multirow[t]{2}{*}{paraTinyBERT} & whisperMedium & 92.51$\pm$0.49 & 91.62$\pm$0.61 & 92.27$\pm$0.70 & \textbf{93.36$\pm$0.52} \\
 & hubertLargeFT & 89.13$\pm$0.74 & 88.32$\pm$0.51 & \textbf{89.62$\pm$0.42} & 88.97$\pm$0.77 \\
\bottomrule
\end{tabular}
}
\end{table}

\begin{table}[!tp]
\centering
\caption{AUC (\%) on a 5 fold cross-validation on segment-level predictions on parent data.}
\label{tbl_parent}
\setlength\tabcolsep{2.75pt}
\scalebox{0.75}{
\begin{tabular}{lllcccccccccccc}
\toprule
 & & & \rot{\textbf{objective}} & \rot{\textbf{sentiment}} & \rot{\textbf{anger}} & \rot{\textbf{fear}} & \rot{\textbf{joy}} & \rot{\textbf{sadness}} & \rot{\textbf{neutral}} & \rot{\textbf{cohesion}} & \rot{\textbf{rumination}} & \rot{\textbf{overinclude}} & \rot{\textbf{worry}} & \rot{\textbf{criticism}} \\
Method & LLM & Audio Encoder &  &  &  &  &  &  &  &  &  &  &  &  \\
\midrule
LoRA & nRoBERTa & - & 88.02 & 92.38 & 90.87 & 89.81 & 89.77 & 89.92 & 79.49 & 78.30 & 59.01 & 85.76 & 89.99 & 91.93 \\
C-LoRA & nRoBERTa & - & 86.82 & 91.42 & 88.76 & 75.32 & 87.74 & 87.95 & 77.39 & 66.40 & 55.68 & 65.06 & 86.94 & 88.91 \\
\cmidrule[0.05pt](lr){1-3}
\multirow[t]{3}{*}{Transfer} & \multirow[t]{3}{*}{nRoBERTa} & whisperM & 85.70 & 90.39 & 89.56 & 89.10 & 89.13 & 89.51 & 78.28 & 79.63 & 82.40 & 83.55 & 89.03 & 90.15 \\
 &  & wav2vec2LFT & 85.15 & 91.07 & 89.20 & 88.43 & 88.69 & 88.76 & 77.36 & 76.89 & 79.16 & 82.25 & 88.47 & 90.15 \\
 &  & hubertLFT & 85.08 & 90.94 & 88.31 & 87.87 & 88.56 & 88.95 & 76.92 & 77.06 & 79.70 & 82.31 & 88.26 & 89.67 \\
\cmidrule[0.05pt](lr){1-3}
\multirow[t]{3}{*}{SPECTRA-G} & \multirow[t]{3}{*}{nRoBERTa} & whisperM & 90.03 & 94.86 & 89.84 & 88.52 & 91.65 & 91.52 & 80.53 & 78.15 & 59.43 & 82.45 & 91.09 & 92.34 \\
 &  & wav2vec2LFT & 90.11 & 94.58 & 88.54 & 88.64 & 91.64 & 91.09 & 76.32 & 74.71 & 66.43 & 76.74 & 90.52 & \textbf{93.46} \\
 &  & hubertLFT & 89.91 & 94.57 & 90.92 & 85.92 & 91.69 & 87.55 & 80.93 & 76.15 & 65.35 & 74.84 & 90.78 & 93.43 \\
\cmidrule[0.05pt](lr){1-3}
\multirow[t]{3}{*}{SPECTRA-X} & \multirow[t]{3}{*}{nRoBERTa} & whisperM & 91.08 & 95.09 & 90.34 & 89.71 & 92.08 & \textbf{93.17} & \textbf{81.92} & 78.51 & 60.41 & 82.86 & 91.09 & 93.17 \\
 &  & wav2vec2LFT & 90.74 & 94.89 & 89.11 & 88.71 & 92.65 & 91.4 & 77.05 & 75.53 & 67.42 & 78.01 & 91.36 & 93.23 \\
 &  & hubertLFT & 90.68 & 94.71 & \textbf{91.55} & 86.26 & 91.94 & 88.34 & 81.71 & 76.71 & 66.12 & 75.21 & \textbf{91.5} & 92.95 \\
\midrule
LoRA & pTinyBERT & - & 86.91 & 90.23 & 74.97 & 53.94 & 87.82 & 75.71 & 77.12 & 75.47 & 50.00 & 52.10 & 75.31 & 84.09 \\
C-LoRA & pTinyBERT & - & 85.91 & 89.98 & 78.39 & 65.44 & 87.24 & 82.61 & 76.00 & 72.43 & 54.66 & 62.36 & 80.94 & 86.21 \\
\cmidrule[0.05pt](lr){1-3}
\multirow[t]{3}{*}{Transfer} & \multirow[t]{3}{*}{pTinyBERT} & whisperM & 84.07 & 88.75 & 86.82 & 86.52 & 86.95 & 87.00 & 71.37 & 79.08 & \textbf{83.61} & 83.07 & 87.11 & 87.77 \\
 &  & wav2vec2LFT & 83.04 & 88.42 & 86.18 & 84.52 & 86.21 & 86.69 & 73.06 & 76.84 & 80.53 & 79.95 & 85.88 & 86.73 \\
 &  & hubertLFT & 83.11 & 88.10 & 85.11 & 85.01 & 86.07 & 86.06 & 72.77 & 76.57 & 78.33 & 80.44 & 85.93 & 86.92 \\
\cmidrule[0.05pt](lr){1-3}
\multirow[t]{3}{*}{SPECTRA-G} & \multirow[t]{3}{*}{pTinyBERT} & whisperM & 89.86 & 94.09 & 90.14 & 85.94 & 91.51 & 88.70 & 80.27 & 79.08 & 57.04 & 66.70 & 79.42 & 91.54 \\
 &  & wav2vec2LFT & 90.54 & 94.23 & 90.60 & 78.13 & 91.31 & 89.00 & 80.03 & 81.70 & 58.44 & 77.19 & 84.65 & 87.78 \\
 &  & hubertLFT & 90.51 & 93.78 & 89.11 & 75.13 & 91.59 & 86.15 & 80.09 & 81.87 & 60.90 & 75.91 & 90.02 & 91.07 \\
\cmidrule[0.05pt](lr){1-3}
\multirow[t]{3}{*}{SPECTRA-X} & \multirow[t]{3}{*}{pTinyBERT} & whisperM & 90.15 & 94.63 & 90.28 & 86.52 & 92.88 & 89.67 & 80.36 & 79.71 & 57.55 & 67.4 & 81.05 & 92.14 \\
 &  & wav2vec2LFT & 91.73 & 94.31 & 91.49 & 78.5 & 91.58 & 89.34 & 80.95 & \textbf{82.64} & 58.6 & 77.97 & 84.98 & 88.59 \\
 &  & hubertLFT & 90.96 & 93.87 & 90.14 & 75.49 & 92.54 & 87.13 & 80.59 & 82.0 & 61.01 & 76.97 & 90.77 & 92.29 \\
\midrule
LoRA & aMiniLM12 & - & 86.91 & 90.16 & 58.83 & 51.26 & 87.16 & 50.74 & 77.14 & 71.31 & 50.00 & 50.00 & 68.33 & 72.70 \\
C-LoRA & aMiniLM12 & - & 86.38 & 90.10 & 60.29 & 53.76 & 79.97 & 55.29 & 75.86 & 62.41 & 51.39 & 52.36 & 54.23 & 60.67 \\
\cmidrule[0.05pt](lr){1-3}
\multirow[t]{3}{*}{Transfer} & \multirow[t]{3}{*}{aMiniLM12} & whisperM & 82.68 & 83.32 & 80.24 & 81.81 & 84.06 & 81.81 & 72.29 & 78.50 & 78.86 & 80.40 & 82.44 & 80.56 \\
 &  & wav2vec2LFT & 81.28 & 84.12 & 80.56 & 78.22 & 83.01 & 80.08 & 69.64 & 75.65 & 72.78 & 75.07 & 80.36 & 82.11 \\
 &  & hubertLFT & 80.88 & 83.60 & 77.45 & 78.13 & 81.97 & 78.77 & 68.50 & 74.87 & 69.31 & 74.70 & 79.53 & 79.78 \\
\cmidrule[0.05pt](lr){1-3}
\multirow[t]{3}{*}{SPECTRA-G} & \multirow[t]{3}{*}{aMiniLM12} & whisperM & 91.94 & \textbf{95.30} & 64.82 & 57.31 & 91.38 & 79.54 & 81.35 & 75.48 & 55.00 & 58.86 & 66.89 & 91.72 \\
 &  & wav2vec2LFT & 91.11 & 94.92 & 65.59 & 60.84 & 91.88 & 56.38 & 81.26 & 71.78 & 55.70 & 61.57 & 76.15 & 81.24 \\
 &  & hubertLFT & 91.59 & 95.13 & 55.46 & 56.76 & 92.17 & 75.05 & 79.96 & 70.35 & 55.14 & 63.50 & 72.13 & 89.06 \\
\cmidrule[0.05pt](lr){1-3}
\multirow[t]{3}{*}{SPECTRA-X} & \multirow[t]{3}{*}{aMiniLM12} & whisperM & \textbf{92.69} & 95.12 & 65.26 & 57.99 & 92.52 & 80.33 & 81.9 & 75.77 & 55.2 & 59.76 & 67.2 & 92.15 \\
 &  & wav2vec2LFT & 91.71 & 94.7 & 66.66 & 61.55 & \textbf{93.15} & 56.99 & 81.68 & 72.43 & 56.57 & 61.97 & 76.76 & 81.82 \\
 &  & hubertLFT & 91.98 & 94.86 & 55.51 & 57.61 & 92.94 & 76.05 & 80.98 & 71.06 & 55.84 & 64.25 & 72.88 & 89.84 \\
\midrule
LoRA & aDRoBERTa & - & 87.39 & 91.31 & 89.70 & 88.26 & 88.94 & 88.95 & 78.30 & 77.73 & 78.73 & 84.76 & 89.29 & 90.40 \\
C-LoRA & aDRoBERTa & - & 86.40 & 90.70 & 88.09 & 86.44 & 88.21 & 87.70 & 76.74 & 76.75 & 68.47 & 80.21 & 87.64 & 89.57 \\
\cmidrule[0.05pt](lr){1-3}
\multirow[t]{3}{*}{Transfer} & \multirow[t]{3}{*}{aDRoBERTa} & whisperM & 82.99 & 82.35 & 82.86 & 82.70 & 84.83 & 83.49 & 72.47 & 78.90 & 81.43 & 80.39 & 84.48 & 83.41 \\
 &  & wav2vec2LFT & 81.77 & 85.53 & 81.74 & 80.99 & 84.01 & 82.14 & 70.76 & 75.81 & 75.23 & 76.46 & 82.26 & 83.41 \\
 &  & hubertLFT & 81.73 & 84.80 & 80.38 & 79.58 & 83.01 & 81.52 & 68.97 & 75.77 & 68.52 & 75.51 & 81.40 & 82.23 \\
\cmidrule[0.05pt](lr){1-3}
\multirow[t]{3}{*}{SPECTRA-G} & \multirow[t]{3}{*}{aDRoBERTa} & whisperM & 88.31 & 92.69 & 89.91 & 88.49 & 89.54 & 90.51 & 78.98 & 79.73 & 76.13 & 83.36 & 89.19 & 91.38 \\
 &  & wav2vec2LFT & 88.30 & 92.99 & 90.34 & 89.94 & 90.39 & 89.49 & 79.08 & 79.23 & 75.62 & 84.56 & 89.51 & 91.08 \\
 &  & hubertLFT & 88.93 & 92.89 & 88.74 & 82.09 & 90.08 & 90.25 & 78.43 & 81.11 & 77.65 & 85.93 & 89.39 & 90.46 \\
\cmidrule[0.05pt](lr){1-3}
\multirow[t]{3}{*}{SPECTRA-X} & \multirow[t]{3}{*}{aDRoBERTa} & whisperM & 88.81 & 93.27 & 90.77 & 88.88 & 90.01 & 92.01 & 79.89 & 79.99 & 76.55 & 85.05 & 89.83 & 92.26 \\
 &  & wav2vec2LFT & 88.84 & 94.1 & 91.03 & \textbf{90.95} & 90.75 & 90.81 & 79.39 & 79.6 & 76.81 & 85.62 & 90.94 & 91.68 \\
 &  & hubertLFT & 90.63 & 93.71 & 89.12 & 83.23 & 90.89 & 90.97 & 79.28 & 82.12 & 78.42 & \textbf{87.59} & 89.98 & 91.27 \\
\bottomrule
\end{tabular}
}
\end{table}

On IEMOCAP, the results indicate that the task is already strongly driven by the acoustic modality, with whisperMedium achieving the best unimodal performance overall at 93\% AUC, substantially higher than all text-only models (Table~\ref{tbl_iemocap_unimodal_auc}). This suggests that emotional content in IEMOCAP is heavily reflected in prosodic and paralinguistic cues, which is consistent with the dataset's acted conversational setting. Nevertheless, the text-only comparison remains informative: C-LoRA is consistently competitive and usually outperforms deterministic LoRA, yielding the best text-only result for five of the six language backbones. This trend supports the value of contextual Bayesian low-rank adaptation even in the unimodal setting, where input-dependent uncertainty appears to improve robustness relative to deterministic PEFT. At the same time, the gap between the best text-only result (88.40\% for allRoBERTaLarge) and the best audio-only result highlights the importance of incorporating acoustic evidence when the target is emotion recognition.

The multimodal results in Table~\ref{tbl_iemocap_multimodal_auc} show that SPECTRA is most beneficial when the text backbone and audio backbone are reasonably complementary, rather than when one modality already dominates the prediction. Across the 12 multimodal model combinations, the best SPECTRA variant outperforms the transfer-learning baseline in 9 cases, with the gains most often achieved by the token-level SPECTRA-X variants. In particular, the full cross-attention version performs best for several pairings, suggesting that fine-grained token-to-frame conditioning can provide useful localized acoustic context beyond simple global fusion. The shared-KV variant also performs strongly, achieving the best overall multimodal result of 94.01\%, which indicates that much of the benefit of SPECTRA can be retained with reduced parameter overhead. However, improvements are not universal: for combinations already dominated by a very strong audio backbone, such as allRoBERTaLarge+whisperMedium, the transfer baseline remains best. This pattern suggests that SPECTRA is especially effective when cross-modal conditioning helps resolve uncertainty or complement weaker textual representations, whereas its advantage is naturally smaller when a single modality already provides near-saturated predictive performance.

Table~\ref{tbl_parent} shows a clear and consistent advantage of multimodal contextualized adaptation over both text-only PEFT and conventional transfer-based fusion on the parent segment-level tasks. Overall, the strongest results are typically achieved by the SPECTRA variants, with SPECTRA-X most often providing the best or near-best performance, which suggests that token-level text-audio conditioning is more effective than relying only on a global audio summary. This trend is especially visible on relatively well-structured affective and discourse-related tasks such as objective, sentiment, joy, sadness, neutral, and criticism, where SPECTRA substantially improves over both LoRA/C-LoRA and the transfer baselines across multiple text encoders. These gains indicate that, in this clinical interview setting, acoustic cues provide useful localized evidence that can refine the adapter behavior beyond what deterministic text-only adaptation or pooled multimodal fusion can capture. Importantly, the improvements are not limited to the strongest text models: even weaker backbones such as paraTinyBERT and allMiniLM12 benefit substantially from SPECTRA, suggesting that cross-modal Bayesian conditioning can compensate for limited textual representational capacity. The results on offspring data is provided in the Appendix, Table~\ref{app:tbl_offspring}.

\subsection{Modality-Disagreement Stress Test}
\label{sec:modality_disagreement}

Standard multimodal evaluation uses correctly paired text and audio, but this does not test whether a model reacts appropriately when one modality becomes unreliable. We therefore evaluate the same trained SPECTRA model under four held-out conditions: clean matched text-audio pairs, clean text with noisy audio, clean text with mismatched audio from another example, and corrupted text with clean audio. The label is kept fixed in all conditions, so changes in performance reflect sensitivity to modality reliability rather than changes in the task. We report AUC, negative log-likelihood (NLL), expected calibration error (ECE), and Brier score in Table \ref{tab:modality_disagreement}. Detailed perturbation definitions, calibration equations, reliability diagrams, and uncertainty decomposition results are provided in Appendix~\ref{app:stress_test_details}.

\begin{table}[t]
\centering
\caption{Modality-disagreement stress test. Higher AUC is better; lower NLL, ECE, and Brier score are better. AUC is reported on the $[0,1]$ scale.}
\label{tab:modality_disagreement}
\setlength{\tabcolsep}{5pt}
\scalebox{0.9}{
\begin{tabular}{lcccc}
\toprule
\textbf{Condition} & \textbf{AUC} $\uparrow$ & \textbf{NLL} $\downarrow$ & \textbf{ECE} $\downarrow$ & \textbf{Brier} $\downarrow$ \\
\midrule
Clean text + matched audio     & 0.835 & 0.419 & 0.042 & 0.260 \\
Clean text + noisy audio       & 0.812 & 0.435 & 0.021 & 0.278 \\
Clean text + mismatched audio  & 0.716 & 0.560 & 0.087 & 0.346 \\
Corrupted text + clean audio   & 0.795 & 0.458 & 0.049 & 0.290 \\
\bottomrule
\end{tabular}
}
\end{table}

The clean matched condition performs best overall. Noisy but correctly paired audio causes only a moderate AUC drop, from 0.835 to 0.812, and the model remains well calibrated. In contrast, mismatched audio is substantially more harmful: AUC drops to 0.716, while NLL, ECE, and Brier score all increase sharply. This shows that semantically inconsistent audio is more damaging than degraded but still matched audio, and it exposes a residual overconfidence failure mode under cross-modal contradiction. Corrupted text with clean audio leads to an intermediate degradation, suggesting that valid acoustic evidence can partially compensate when lexical evidence is degraded. These stress-test results support the central motivation of SPECTRA: audio is not merely an additional feature stream, but a contextual signal that can shape both adaptation and predictive confidence.

\begin{figure*}[!t]
	\centering
    \makebox[0.8\linewidth][c]{%
    	\begin{subfigure}[b]{0.4\columnwidth}
    		\centering
    		\includegraphics[width=\linewidth]{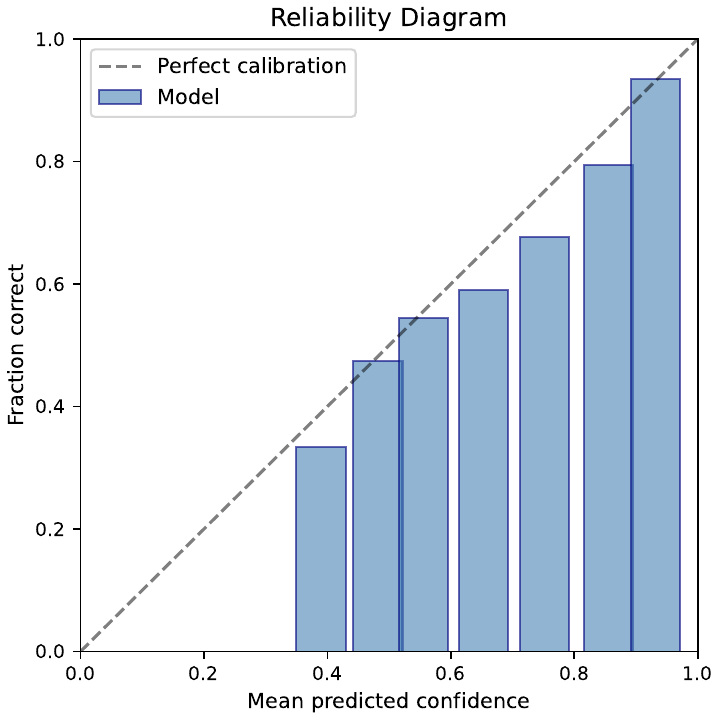}
            \caption{\label{fig_calibration}Calibration diagram}
    	\end{subfigure}
    	\begin{subfigure}[b]{0.6\columnwidth}
    		\centering
    		\includegraphics[width=\linewidth]{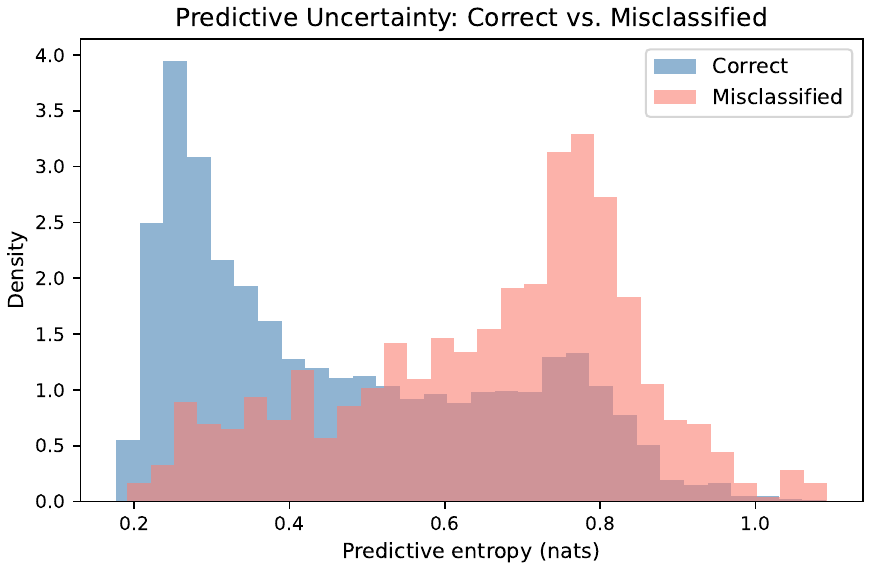}
            \caption{\label{fig_uncertainty_hist}Predictive uncertainty}
    	\end{subfigure}
    }
	\caption{\label{fig_reliability}Calibration and predictive uncertainty analysis.}
\end{figure*}

Figure \ref{fig_reliability} provides complementary evidence about the uncertainty behavior of the model. The reliability diagram in Figure \ref{fig_calibration} shows that the predicted confidence and empirical accuracy follow the diagonal trend very well, indicating that the model is well calibrated overall. Figure \ref{fig_uncertainty_hist} further shows that predictive entropy is informative about error likelihood: correctly classified samples are concentrated at lower uncertainty values, while misclassified samples tend to shift toward higher entropy. This separation suggests that the model's uncertainty estimates are meaningful, as larger predictive uncertainty is generally associated with a greater chance of incorrect predictions.

\section{Conclusion}
\label{sec:conclusion}




We introduced SPECTRA, a multimodal uncertainty-aware PEFT framework that integrates token-level audio-text context into Bayesian low-rank adaptation. By conditioning a variational posterior in the rank-$r$ adapter space on per-layer cross-attended acoustic context, SPECTRA provides a lightweight mechanism for incorporating multimodal reliability cues without modifying the frozen backbone or relying on heavy feature-fusion architectures. This design preserves the scalability advantages of PEFT while enabling heteroscedastic, input-dependent uncertainty that reflects localized cross-modal interactions.

Experimental results on multiple datasets show that SPECTRA is often superior to text-only PEFT baselines and conventional multimodal transfer-learning approaches. In particular, the token-level cross-attention variant yields the most consistent improvements, suggesting that fine-grained cross-modal contextualization is more effective than global audio conditioning in many low-resource prediction settings.

Architecturally, SPECTRA's Bayesian core is not audio-specific. It consumes a text query sequence and a masked sequence of context tokens projected to a small common dimension. Image patches, video clips, physiological measurements, or other temporal sensor features can replace audio frames as keys/values while leaving $q_\phi\!\left(E_{x,t}^{\ell}\mid[z_t^\ell;\tilde{u}_t^\ell]\right)$ and $B^\ell E_{x,t}^\ell A^\ell$ unchanged. The implementation separates encoder interfaces, shared projection, cross-attention, and the modality-agnostic posterior head. Multiple context modalities could be concatenated as a typed K/V memory or attended separately and gated before the posterior head.

Overall, our findings highlight that multimodal information can play a dual role: not only as a source of predictive features, but also as a signal for modulating model confidence and adaptation. This provides a principled and parameter-efficient direction for uncertainty-aware multimodal learning with large frozen backbones. Future work may extend SPECTRA to additional modalities, longer temporal contexts, and other high-stakes low-resource applications where reliable uncertainty estimation is essential.

\bibliographystyle{plainnat}
\bibliography{CanAI2026-CoCoLoRA}

@inproceedings{rahmati2025clora,
title={C-Lo{RA}: Contextual Low-Rank Adaptation for Uncertainty Estimation in Large Language Models},
author={Amir Hossein Rahmati and Sanket Jantre and Weifeng Zhang and Yucheng Wang and Byung-Jun Yoon and Nathan Urban and Xiaoning Qian},
booktitle={NeurIPS},
year={2025}
}

@article{hu2022lora,
  title={Lora: Low-rank adaptation of large language models.},
  author={Hu, Edward J and Shen, Yelong and Wallis, Phillip and Allen-Zhu, Zeyuan and Li, Yuanzhi and Wang, Shean and Wang, Liang and Chen, Weizhu and others},
  journal={Iclr},
  volume={1},
  number={2},
  pages={3},
  year={2022}
}

@article{wang2024blob,
  title={Blob: Bayesian low-rank adaptation by backpropagation for large language models},
  author={Wang, Yibin and Shi, Haizhou and Han, Ligong and Metaxas, Dimitris and Wang, Hao},
  journal={NeurIPS},
  volume={37},
  pages={67758--67794},
  year={2024}
}

@article{laplacelora_yang2023bayesian,
  title={Bayesian low-rank adaptation for large language models},
  author={Yang, Adam X and Robeyns, Maxime and Wang, Xi and Aitchison, Laurence},
  journal={ICLR},
  year={2024}
}

@article{llms_brown2020language,
  title={Language models are few-shot learners},
  author={Brown, Tom and Mann, Benjamin and Ryder, Nick and Subbiah, Melanie and Kaplan, Jared D and Dhariwal, Prafulla and Neelakantan, Arvind and Shyam, Pranav and Sastry, Girish and Askell, Amanda and others},
  journal={Advances in neural information processing systems},
  volume={33},
  pages={1877--1901},
  year={2020}
}

@article{llms_minaee2024large,
  title={Large language models: A survey},
  author={Minaee, Shervin and Mikolov, Tomas and Nikzad, Narjes and Chenaghlu, Meysam and Socher, Richard and Amatriain, Xavier and Gao, Jianfeng},
  journal={arXiv preprint arXiv:2402.06196},
  year={2024}
}

@article{audiolm_borsos2023audiolm,
  title={Audiolm: a language modeling approach to audio generation},
  author={Borsos, Zal{\'a}n and Marinier, Rapha{\"e}l and Vincent, Damien and Kharitonov, Eugene and Pietquin, Olivier and Sharifi, Matt and Roblek, Dominik and Teboul, Olivier and Grangier, David and Tagliasacchi, Marco and others},
  journal={IEEE/ACM transactions on audio, speech, and language processing},
  volume={31},
  pages={2523--2533},
  year={2023},
  publisher={IEEE}
}

@article{uncertainty_xiong2023can,
  title={Can llms express their uncertainty? an empirical evaluation of confidence elicitation in llms},
  author={Xiong, Miao and Hu, Zhiyuan and Lu, Xinyang and Li, Yifei and Fu, Jie and He, Junxian and Hooi, Bryan},
  journal={ICLR},
  year={2024}
}

@article{uncertainty_leng2024taming,
  title={Taming overconfidence in llms: Reward calibration in rlhf},
  author={Leng, Jixuan and Huang, Chengsong and Zhu, Banghua and Huang, Jiaxin},
  journal={ICLR},
  year={2025}
}

@inproceedings{fu2023effectiveness_peft,
  title={On the effectiveness of parameter-efficient fine-tuning},
  author={Fu, Zihao and Yang, Haoran and So, Anthony Man-Cho and Lam, Wai and Bing, Lidong and Collier, Nigel},
  booktitle={AAAI},
  volume={37},
  pages={12799--12807},
  year={2023}
}

@article{liu2022few_ia3,
  title={Few-shot parameter-efficient fine-tuning is better and cheaper than in-context learning},
  author={Liu, Haokun and Tam, Derek and Muqeeth, Mohammed and Mohta, Jay and Huang, Tenghao and Bansal, Mohit and Raffel, Colin A},
  journal={NeurIPS},
  volume={35},
  pages={1950--1965},
  year={2022}
}

@inproceedings{zhang2023adaptive_adalora,
  title={Adaptive Budget Allocation for Parameter-Efficient Fine-Tuning },
  author={Qingru Zhang and Minshuo Chen and Alexander Bukharin and Pengcheng He and Yu Cheng and Weizhu Chen and Tuo Zhao},
  booktitle={ICLR},
  year={2023}
}

@article{forbow_uher2014familial,
  title={A familial risk enriched cohort as a platform for testing early interventions to prevent severe mental illness},
  author={Uher, Rudolf and Cumby, Jill and MacKenzie, Lynn E and Morash-Conway, Jessica and Glover, Jacqueline M and Aylott, Alice and Propper, Lukas and Abidi, Sabina and Bagnell, Alexa and Pavlova, Barbara and others},
  journal={BMC psychiatry},
  volume={14},
  number={1},
  pages={344},
  year={2014},
  publisher={Springer}
}

@article{muller1994critical_icc,
  title={A critical discussion of intraclass correlation coefficients},
  author={M{\"u}ller, Reinhold and B{\"u}ttner, Petra},
  journal={Statistics in medicine},
  volume={13},
  number={23-24},
  pages={2465--2476},
  year={1994},
  publisher={Wiley Online Library}
}

@article{busso2008iemocap,
  title={IEMOCAP: Interactive emotional dyadic motion capture database},
  author={Busso, Carlos and Bulut, Murtaza and Lee, Chi-Chun and Kazemzadeh, Abe and Mower, Emily and Kim, Samuel and Chang, Jeannette N and Lee, Sungbok and Narayanan, Shrikanth S},
  journal={Language resources and evaluation},
  volume={42},
  number={4},
  pages={335--359},
  year={2008},
  publisher={Springer}
}
\label{sec_references}

\clearpage
\appendix

\section*{Technical appendices and supplementary material}
\label{app:sec_supplementary}
In this appendix, we present further details, higher-resolution figures/tables, and additional experimental results that are not included in the main text due to space constraints.

\section{Proposed Method: SPECTRA Algorithm}

The pseudo-code of SPECTRA is provided in Algorithm \ref{alg:caliber}. A frozen text backbone produces token-level hidden states that enter LoRA adapters at each transformer layer, while a frozen audio encoder produces frame-level acoustic embeddings. The audio frames are first projected into a shared context space and then consumed by lightweight per-layer cross-attention modules, where LoRA text features act as queries and audio frames serve as keys and values. The resulting token-conditioned cross-modal context is concatenated with the local low-rank feature to parameterize a variational posterior over the compact latent matrix $E^\ell_{x,t}$. Samples from this posterior modulate the low-rank update $B^\ell E^\ell_{x,t} A^\ell$, enabling context-aware Bayesian adaptation with calibrated uncertainty while preserving the efficiency of PEFT.

\begin{algorithm}[h]
\caption{Training SPECTRA}
\label{alg:caliber}
\begin{algorithmic}[1]
\STATE \textbf{Input:} Training set $\mathcal{D}=\{(x_i,y_i)\}_{i=1}^N$, frozen transformer backbone, frozen audio encoder
\STATE Initialize low-rank parameters $\{A^\ell,B^\ell\}_{\ell=1}^L$
\STATE Initialize audio projection $P_a$, per-layer cross-attention parameters $\{W_Q^\ell,W_K^\ell,W_V^\ell,W_O^\ell\}_{\ell=1}^L$, and inference heads $\{H^\ell_{\phi^\ell}\}_{\ell=1}^L$
\FOR{each minibatch $(x,y)$}
    \STATE Encode the text sequence with the frozen transformer and obtain hidden states $\{x_t^{\ell-1}\}$
    \STATE Encode the accompanying audio and obtain frame-level embeddings $\{a_s(x)\}_{s=1}^{T_a}$
    \STATE Project audio frames: $u_s(x)=P_a(a_s(x))$
    \FOR{each layer $\ell=1,\dots,L$}
        \FOR{each token position $t=1,\dots,T_x$}
            \STATE Compute local LoRA feature: $z_t^\ell = A^\ell x_t^{\ell-1}$
            \STATE Form token-conditioned audio context $\tilde u_t^\ell(x)$ using Eqs.~\eqref{eq:caliber_crossattn}-\eqref{eq:caliber_crossattn_expanded}
            \STATE Form contextual summary $\eta_t^\ell(x)=[z_t^\ell;\tilde u_t^\ell(x)]$
            \STATE Compute posterior parameters $(\mu^\ell_{E,t}(x),\sigma^\ell_{E,t}(x))$ using Eqs.~\eqref{eq:caliber_q}-\eqref{eq:caliber_sigma}
            \STATE Sample $E^\ell_{x,t}$ via Eq.~\eqref{eq:caliber_reparam}
            \STATE Compute stochastic adapter update:
            \[
            \Delta h_t^\ell(x)=B^\ell E^\ell_{x,t} A^\ell x_t^{\ell-1}
            \]
        \ENDFOR
    \ENDFOR
    \STATE Compute task likelihood and ELBO objective in Eq.~\eqref{eq:caliber_elbo}
    \STATE Update only trainable parameters $\theta,\phi$ by gradient ascent on the ELBO (or equivalently gradient descent on the negative ELBO)
\ENDFOR
\end{algorithmic}
\end{algorithm}

\subsection{Preliminaries}

\paragraph{Low-Rank Adaptation (LoRA)} Low-Rank Adaptation (LoRA) is a parameter-efficient fine-tuning method designed to adapt large pre-trained models to downstream tasks while keeping the backbone weights frozen \cite{hu2022lora}. The core assumption behind LoRA is that task-specific weight updates lie in a low-dimensional subspace. Instead of updating the full weight matrix, LoRA learns a low-rank decomposition of the weight update.

Consider a linear transformation with frozen pre-trained weights $W_0 \in \mathbb{R}^{d \times k}$. The adapted forward pass is defined as:
{\small
\begin{equation}
h \;=\; (W_0 + \Delta W)x \;=\; (W_0 + BA)x,
\label{eq:lora_standard_caliber}
\end{equation}
}
where $x \in \mathbb{R}^{k}$ is the input, $h \in \mathbb{R}^{d}$ is the output, and $\Delta W = BA$ is the low-rank update. Here, $B \in \mathbb{R}^{d \times r}$ and $A \in \mathbb{R}^{r \times k}$ are trainable matrices with rank $r \ll \min(d,k)$. This factorization reduces the number of trainable parameters from $d\times k$ to $r\times(d+k)$, yielding substantial savings in memory and computation while often matching full fine-tuning performance.

\paragraph{Bayesian Uncertainty-Aware Low-Rank Adaptation} While LoRA provides efficient deterministic adaptation, it does not model uncertainty in the learned parameters. Full Bayesian inference over all backbone weights is computationally infeasible for modern transformer models. A scalable alternative is to restrict uncertainty modeling to the low-rank adapters.

Bayesian LoRA by Backpropagation (BLoB) introduces uncertainty into the LoRA parameters while keeping the frozen backbone intact. Given the LoRA decomposition $\Delta W = BA$, BLoB maintains $B$ as deterministic and places a variational posterior over $A$. Specifically, a mean-field variational inference approximation is used: $q(A) = \mathcal{N}\big(A \mid \mu_A, \Omega_A^2\big)$, where $\mu_A$ and $\Omega_A^2$ denote the variational mean and variance parameters. The variational parameters are learned by maximizing the Evidence Lower Bound (ELBO):
{\small
\[
\mathcal{L}' 
= \mathbb{E}_{q} \big[ \log p(\mathcal{D} \mid A, B) \big]
- \mathrm{KL}\big[q(A) \,\|\, p(A)\big].
\]
}

The first term corresponds to the expected log-likelihood under the variational posterior, while the second term regularizes the solution via the Kullback-Leibler divergence to the prior. Optimization is performed using the reparameterization trick, enabling gradient-based training of both mean and variance parameters.

By confining Bayesian inference to the low-rank adapters, BLoB preserves the scalability benefits of LoRA while providing principled uncertainty estimation. This makes it an effective compromise between deterministic fine-tuning and fully Bayesian neural networks.

\subsection{SPECTRA: Stochastic Posterior Estimation with Cross-modal Token-level Rank Adaptation}

The preliminaries above highlight a practical gap in low-resource, high-stakes settings. Deterministic PEFT methods such as LoRA are efficient, but they can be overconfident when supervision is scarce or inputs are ambiguous. Bayesianizing adapters, as in BLoB, introduces epistemic uncertainty, yet its posterior is typically input-independent, limiting its ability to respond to sample-specific noise. C-LoRA addresses this by amortizing a contextual variational posterior in the low-rank space, producing heteroscedastic (input-dependent) uncertainty, but it conditions that uncertainty primarily on internal language-model features. In multimodal speech-centered applications, however, predictive uncertainty is often driven by external factors such as background noise, channel variability, prosodic variation, or speaker state, which are observable in audio but not necessarily recoverable from text alone.

We therefore propose SPECTRA (Stochastic Posterior Estimation with Cross-modal Token-level Rank Adaptation), a multimodal, uncertainty-aware PEFT framework that extends contextual Bayesian adapters by conditioning the variational posterior over the low-rank latent on per-layer text-audio cross-attention. A visual representation of the framework is shown in Figure~\ref{fig_architecture}. The key design principle is to use audio as a contextual reliability signal rather than as a heavy feature-fusion stream. Concretely, we keep the transformer backbone frozen and avoid expanding representation dimensionality at the classifier level. Instead, we allow each layer to form a lightweight token-conditioned view of the audio sequence and use that cross-modal context to shape uncertainty and adaptation within the rank-$r$ space.

\paragraph{Setup and notation.} Let $\mathcal{D}=\{(x_i,y_i)\}_{i=1}^N$ denote a dataset of $N$ i.i.d.\ samples. We consider a pre-trained transformer with $L$ layers. In a linear sub-layer at layer $\ell\in[L]$, let $W^\ell_0\in\mathbb{R}^{d\times k}$ be the frozen pre-trained weight. For sequence inputs, let $x_t^{\ell-1}\in\mathbb{R}^{k}$ denote the hidden representation at token position $t\in[T_x]$, and let $h_t^\ell\in\mathbb{R}^{d}$ denote the corresponding output. LoRA parameterizes a low-rank update $\Delta W^\ell$ via matrices $B^\ell\in\mathbb{R}^{d\times r}$ and $A^\ell\in\mathbb{R}^{r\times k}$ with rank $r\ll d$. Following the lightweight factorization introduced in C-LoRA \cite{rahmati2025clora}, we insert an additional matrix $E^\ell_{x,t}$ in the low-dimensional space:
{\small
\begin{equation}
h_t^\ell
\;=\;
\bigl(W^\ell_0 + \Delta W_t^\ell(x)\bigr)x_t^{\ell-1}
\;=\;
W^\ell_0 x_t^{\ell-1} + B^\ell E^\ell_{x,t} A^\ell x_t^{\ell-1},
\qquad
\Delta W_t^\ell(x) \;=\; B^\ell E^\ell_{x,t} A^\ell .
\label{eq:caliber_forward}
\end{equation}
}
The LoRA intermediate feature is defined token-wise as: $z_t^\ell = A^\ell x_t^{\ell-1}\in\mathbb{R}^{r}$.


\paragraph{Where uncertainty lives and what cross-modal context controls.} SPECTRA restricts stochasticity to the compact latent $E^\ell_{x,t}\in\mathbb{R}^{r\times r}$, keeping $A^\ell$ and $B^\ell$ deterministic. This preserves the scalability of PEFT while enabling uncertainty estimation through variational inference. The key distinction from prior work is what the posterior conditions on. Instead of relying only on internal text-derived features, SPECTRA conditions the posterior on a token-specific cross-modal context obtained by letting text query the accompanying audio frames. As a result, uncertainty can increase when the audio suggests unreliability (e.g., noisy or weakly informative regions) and decrease when the acoustic signal is temporally aligned and consistent with the text.

\paragraph{Per-layer text-audio cross-attention context.} Each input sample $x$ is accompanied by a sequence of frame-level audio embeddings
{\small
\[
a(x)=\big[a_1(x),\dots,a_{T_a}(x)\big], \qquad a_s(x)\in\mathbb{R}^{d_a},
\]
}
obtained from a frozen or separately trained audio encoder, where $T_a$ denotes the number of audio frames. We first project each audio frame into a shared context space of dimension $c \ll d_a$:
{\small
\begin{equation}
u_s(x) = P_a\!\big(a_s(x)\big)\in\mathbb{R}^{c},
\qquad s\in[T_a],
\label{eq:caliber_audio_proj}
\end{equation}
}
and collect the projected audio sequence as $U(x)=\big[u_1(x),\dots,u_{T_a}(x)\big]\in\mathbb{R}^{T_a\times c}$. At layer $\ell$, SPECTRA forms a token-conditioned audio context by cross-attending the local LoRA text features $\{z_t^\ell\}_{t=1}^{T_x}$ to the projected audio frames. Specifically, for each token position $t$, we define
{\small
\begin{equation}
\tilde u_t^\ell(x)
=
\mathrm{CrossAttn}^\ell\!\bigl(q_t^\ell,\,K^\ell(x),\,V^\ell(x)\bigr)
\in\mathbb{R}^{r},
\label{eq:caliber_crossattn}
\end{equation}
}
where the query is derived from the local text feature, $q_t^\ell = W_Q^\ell z_t^\ell \in \mathbb{R}^{d_c}$,
and the keys and values are derived from the projected audio sequence,
{\small
\begin{equation}
K^\ell(x) = U(x)W_K^\ell \in \mathbb{R}^{T_a\times d_c},
\qquad
V^\ell(x) = U(x)W_V^\ell \in \mathbb{R}^{T_a\times d_c}.
\label{eq:caliber_kv}
\end{equation}
}
Using scaled dot-product attention, the token-conditioned context is
{\small
\begin{equation}
\tilde u_t^\ell(x)
=
W_O^\ell
\left[
\mathrm{softmax}
\!\left(
\frac{q_t^\ell (K^\ell(x))^\top}{\sqrt{d_c}}
\right)
V^\ell(x)
\right],
\label{eq:caliber_crossattn_expanded}
\end{equation}
}
where $W_O^\ell$ maps the attention output back to $\mathbb{R}^{r}$. This per-layer design allows different transformer layers to attend to different temporal regions and acoustic cues, yielding a depth-specific cross-modal reliability signal without high-dimensional multimodal fusion.

\paragraph{Context-conditioned variational posterior over $\mathbf{E^\ell_{x,t}}$.} We model $E^\ell_{x,t}\in\mathbb{R}^{r\times r}$ as a latent random matrix with an input-dependent Gaussian variational posterior. The posterior is conditioned on (i) the local LoRA feature $z_t^\ell$, which captures how the current text activation enters the low-rank adapter, and (ii) the cross-attended audio context $\tilde u_t^\ell(x)$, which captures token-specific external reliability information. We concatenate these to form the contextual summary
{\small
\begin{equation}
\eta_t^\ell(x)
=
[\,z_t^\ell;\tilde u_t^\ell(x)\,]
\in \mathbb{R}^{2r}.
\label{eq:caliber_concat}
\end{equation}
}
Let $H^\ell_{\phi^\ell}:\mathbb{R}^{2r}\to\mathbb{R}^{2r^2}$ output a mean vector $\mu^\ell_{E,t}(x)\in\mathbb{R}^{r^2}$ and a log-variance vector $\log v^\ell_{E,t}(x)\in\mathbb{R}^{r^2}$. Then
{\small
\begin{equation}
q_{\phi^\ell}\!\left(\mathrm{vec}(E^\ell_{x,t})\mid x\right)
\;=\;
\mathcal{N}\!\left(\mu^\ell_{E,t}(x),\,\mathrm{diag}\!\bigl(\sigma^\ell_{E,t}(x)^2\bigr)\right),
\label{eq:caliber_q}
\end{equation}
}
where $\sigma^\ell_{E,t}(x)\in\mathbb{R}^{r^2}_{>0}$ is parameterized using a scaled softplus:
{\small
\begin{equation}
\sigma^\ell_{E,t}(x)
=
\varepsilon\,\mathrm{softplus}\!\bigl(\log v^\ell_{E,t}(x)\bigr)+\delta,
\qquad
\varepsilon>0,\;\delta\approx 10^{-6}.
\label{eq:caliber_sigma}
\end{equation}
}
Equation~\eqref{eq:caliber_q} induces heteroscedastic uncertainty because both $\mu^\ell_{E,t}(x)$ and $\sigma^\ell_{E,t}(x)$ depend on the local text feature and the token-specific cross-modal context. Consequently, the same lexical pattern can yield different posterior variance depending on which acoustic frames it attends to and how informative or reliable those frames appear.

\paragraph{Reparameterized sampling and stochastic low-rank update.} We sample $E^\ell_{x,t}$ via the reparameterization trick. Let $\xi_t^\ell\sim\mathcal{N}(0,I_{r^2})$:
{\small
\begin{equation}
\mathrm{vec}(E^\ell_{x,t})
=
\mu^\ell_{E,t}(x) + \sigma^\ell_{E,t}(x)\odot \xi_t^\ell,
\label{eq:caliber_reparam}
\end{equation}
}
and reshape to $E^\ell_{x,t}\in\mathbb{R}^{r\times r}$. The resulting stochastic adapter contribution is
{\small
\begin{equation}
\Delta h_t^\ell(x)
=
B^\ell E^\ell_{x,t} z_t^\ell
=
B^\ell E^\ell_{x,t} A^\ell x_t^{\ell-1}.
\label{eq:caliber_delta_h}
\end{equation}
}
Importantly, stochasticity remains confined to $r\times r$ variables, keeping the Bayesian complexity per token and per layer at $O(r^2)$ rather than scaling with the backbone dimension $d$.

\paragraph{Prior and ELBO objective.} We place a factorized Gaussian prior on $\mathrm{vec}(E^\ell)$:
{\small
\begin{equation}
p\!\left(\mathrm{vec}(E^\ell)\right)
=
\mathcal{N}\!\left(0,\beta^2 I_{r^2}\right),
\label{eq:caliber_prior}
\end{equation}
}
where $\beta>0$ controls the prior scale. Let $\theta$ collect all deterministic parameters, including $\{A^\ell,B^\ell\}_{\ell=1}^L$, the audio projection $P_a$, and the per-layer cross-attention parameters $\{W_Q^\ell,W_K^\ell,W_V^\ell,W_O^\ell\}_{\ell=1}^L$. Let $\phi=\{\phi^\ell\}_{\ell=1}^L$ denote the variational parameters of the inference heads. SPECTRA maximizes the ELBO:
{\small
\begin{equation}
\mathcal{L}(\theta,\phi)
=
\sum_{i=1}^N
\Biggl[
\mathbb{E}_{q_\phi(E_{x_i}\mid x_i)}
\log p_\theta\!\left(y_i\mid x_i,E_{x_i}\right)
-
\gamma
\sum_{\ell=1}^L
\frac{1}{T_{x_i}}
\sum_{t=1}^{T_{x_i}}
\mathrm{KL}\!\left(
q_{\phi^\ell}\!\left(E^\ell_{x_i,t}\mid x_i\right)
\,\|\, p(E^\ell)
\right)
\Biggr],
\label{eq:caliber_elbo}
\end{equation}
}
with $\gamma>0$ a KL reweighting coefficient. The complete learning objective is expressed as a summation over all samples: $\mathcal{L}= \sum_{(x,y)\in\mathcal{D}}\mathcal{L}(x,y)$. We optimize the model by minimizing the negative ELBO with respect to all trainable parameters, where the deterministic adapter parameters ${A^\ell,B^\ell}_{\ell=1}^L$ are included in $\theta$. Equivalently, training minimizes the expected negative log-likelihood (NLL) with the KL regularization term. Averaging the KL across token positions prevents the regularization magnitude from growing trivially with sequence length, while still encouraging the contextual posterior to remain close to the prior unless the data support deviation from it. The conditioning on the cross-attended context $\tilde u_t^\ell(x)$ means that this deviation can systematically reflect the reliability of the accompanying audio signal.

\paragraph{Stochastic optimization.} Using one Monte Carlo draw per sample (as in C-LoRA), we approximate the expected log-likelihood term as
{\small
\begin{equation}
\mathbb{E}_{q_\phi(E_{x_i}\mid x_i)}
\log p_\theta\!\left(y_i\mid x_i,E_{x_i}\right)
\;\approx\;
\log p_\theta\!\left(y_i\mid x_i,E^{(1)}_{x_i}\right),
\qquad
E^{(1)}_{x_i}\sim q_\phi(E_{x_i}\mid x_i).
\label{eq:caliber_mc_train}
\end{equation}
}
Optimization is performed end-to-end using the reparameterization trick, updating only the low-rank adapter parameters, cross-attention projections, and variational inference heads, while keeping the backbone frozen. Monte Carlo sampling provides uncertainty estimates that reflect variability induced by both the contextual conditioning and the stochastic adapter representation.

\paragraph{Posterior predictive inference and cross-modal uncertainty.} At test time, SPECTRA supports either a deterministic approximation using posterior means or Monte Carlo model averaging. For a test input $x^\ast$ with accompanying frame-level audio sequence $a(x^\ast)$:
{\small
\begin{equation}
p(y^\ast\mid x^\ast,\mathcal{D})
\;\approx\;
\frac{1}{M}\sum_{m=1}^M
p_\theta\!\left(y^\ast\mid x^\ast,E^{(m)}_{x^\ast}\right),
\qquad
E^{(m)}_{x^\ast}\sim q_\phi(E_{x^\ast}\mid x^\ast).
\label{eq:caliber_ppd}
\end{equation}
}
Because $q_\phi(E_{x^\ast}\mid x^\ast)$ is conditioned on both the local text features $\{z_t^\ell\}$ and the per-layer cross-attended audio context $\{\tilde u_t^\ell(x^\ast)\}$, predictive uncertainty becomes explicitly multimodal, token-sensitive, and heteroscedastic. Practically, this provides a principled mechanism for modulating confidence based on temporally localized acoustic evidence without changing the frozen backbone or introducing heavy multimodal fusion blocks.

\paragraph{Predictive uncertainty decomposition.}
\label{sec:uncertainty_decomposition}

SPECTRA produces predictive uncertainty through Monte Carlo marginalization over the
context-conditioned stochastic adapter latent variables. For a held-out input $x_i$, let
\[
E_i^{(m)} =
\left\{ E_{i,t}^{\ell,(m)} \right\}_{\ell=1,t=1}^{L,T_i}
\sim q_{\phi}(E_i \mid x_i),
\qquad m=1,\ldots,M,
\]
denote the $m$-th draw from the variational posterior over the collection of stochastic
low-rank adapter latents across layers and token positions. Each draw induces a categorical
predictive distribution
\[
\mathbf{p}_i^{(m)}
=
p_{\theta}(y_i \mid x_i, E_i^{(m)})
=
\operatorname{softmax}
\left(
f_{\theta}(x_i, E_i^{(m)})
\right)
\in \Delta^{C-1},
\]
where $C$ is the number of classes and $\Delta^{C-1}$ denotes the probability simplex.
The posterior predictive distribution is approximated by Monte Carlo averaging:
\[
\bar{\mathbf{p}}_i
=
p(y_i \mid x_i,\mathcal{D})
\approx
\frac{1}{M}
\sum_{m=1}^{M}
\mathbf{p}_i^{(m)}.
\]
The final prediction is: $\hat{y}_i = \arg\max_{c \in \{1,\ldots,C\}} \bar{p}_{ic}$. The predictive entropy of the averaged distribution measures the total predictive
uncertainty:
\[
U_{\mathrm{total}}(x_i)
=
H\!\left[\bar{\mathbf{p}}_i\right]
=
-\sum_{c=1}^{C}
\bar{p}_{ic}
\log
\left(
\bar{p}_{ic} + \epsilon_h
\right),
\]
where $\epsilon_h$ is a small constant used only for numerical stability. This quantity
captures the overall uncertainty in the final prediction after marginalizing over the
stochastic adapter posterior. Following the standard Bayesian classification decomposition, we further separate total
predictive uncertainty into aleatoric and epistemic components. The aleatoric component is
estimated as the expected entropy of the class distribution under posterior samples:
\[
U_{\mathrm{alea}}(x_i)
=
\mathbb{E}_{q_{\phi}(E_i \mid x_i)}
\left[
H\!\left[
p_{\theta}(y_i \mid x_i,E_i)
\right]
\right]
\approx
\frac{1}{M}
\sum_{m=1}^{M}
H\!\left[
\mathbf{p}_i^{(m)}
\right],
\]
with: $H\!\left[
\mathbf{p}_i^{(m)}
\right]
=
-\sum_{c=1}^{C}
p_{ic}^{(m)}
\log
\left(
p_{ic}^{(m)} + \epsilon_h
\right)$. This term is large when individual stochastic adapter samples are themselves uncertain,
which typically corresponds to intrinsically ambiguous examples, overlapping classes, or
label noise. The epistemic component is measured by the mutual information between the prediction and
the stochastic adapter latent:
\[
U_{\mathrm{epi}}(x_i)
=
I(y_i,E_i \mid x_i,\mathcal{D})
=
H\!\left[
\bar{\mathbf{p}}_i
\right]
-
\mathbb{E}_{q_{\phi}(E_i \mid x_i)}
\left[
H\!\left[
p_{\theta}(y_i \mid x_i,E_i)
\right]
\right].
\]
Using the Monte Carlo approximation, this becomes
\[
U_{\mathrm{epi}}(x_i)
\approx
U_{\mathrm{total}}(x_i)
-
U_{\mathrm{alea}}(x_i).
\]
Intuitively, $U_{\mathrm{epi}}(x_i)$ is high when different stochastic adapter samples
produce confident but disagreeing predictions, indicating uncertainty about the appropriate
adapter behavior for the input. In SPECTRA, this quantity should be interpreted as
adapter-space epistemic uncertainty, since stochasticity is confined to the compact
latent matrix $E_i$ in the low-rank adapter space rather than placed over all backbone
parameters.

Because SPECTRA conditions $q_{\phi}(E_i \mid x_i)$ on both the local text-derived
low-rank features and the cross-attended audio context, this decomposition provides a
fine-grained view of multimodal uncertainty. When the audio-text evidence is ambiguous,
weakly aligned, or unreliable, the posterior over $E_i$ may induce greater variability
across stochastic adapter samples, increasing $U_{\mathrm{epi}}$. In contrast, when the
input is intrinsically ambiguous but posterior samples agree in their uncertainty,
$U_{\mathrm{alea}}$ is expected to dominate. For tasks with different numbers of classes, we use normalized uncertainty scores:
\[
\widetilde{U}_{\mathrm{total}}(x_i)
=
\frac{U_{\mathrm{total}}(x_i)}{\log C},
\qquad
\widetilde{U}_{\mathrm{alea}}(x_i)
=
\frac{U_{\mathrm{alea}}(x_i)}{\log C},
\qquad
\widetilde{U}_{\mathrm{epi}}(x_i)
=
\frac{U_{\mathrm{epi}}(x_i)}{\log C}.
\]
This normalization maps all entropy-based scores to $[0,1]$, making uncertainty values
comparable across binary and multi-class prediction tasks.

\paragraph{Epistemic and aleatoric uncertainty analysis.}
In addition to reporting predictive entropy, we decompose the uncertainty of SPECTRA into
aleatoric and epistemic components using the same $M=10$ Monte Carlo stochastic forward
passes used for posterior predictive inference. For each held-out example, we compute the
total uncertainty $U_{\mathrm{total}}$, the expected predictive entropy
$U_{\mathrm{alea}}$, and the mutual-information-based adapter-space epistemic uncertainty
$U_{\mathrm{epi}}$. Since our benchmark contains both binary and multi-class tasks, all
entropy-based quantities are normalized by $\log C$ before aggregation.

This decomposition allows us to distinguish between two qualitatively different sources of
uncertainty. High aleatoric uncertainty indicates that individual sampled adapters remain
uncertain, suggesting intrinsically ambiguous, noisy, or difficult-to-label examples. High
epistemic uncertainty indicates disagreement across stochastic adapter samples, suggesting
that the model is uncertain about the appropriate low-rank adaptation for the given
audio-text input. In the context of SPECTRA, this is particularly useful because the
posterior over the adapter latent is conditioned on token-level cross-modal audio context.
Thus, increased epistemic uncertainty can reflect cases where the text and audio provide
weak, conflicting, or unreliable evidence for adaptation. Let
\[
\mathcal{I}_{\mathrm{corr}}
=
\{i : \hat{y}_i = y_i\},
\qquad
\mathcal{I}_{\mathrm{err}}
=
\{i : \hat{y}_i \neq y_i\}.
\]
For each uncertainty type
$b \in \{\mathrm{total},\mathrm{alea},\mathrm{epi}\}$, we summarize the mean uncertainty
over correctly and incorrectly classified examples:
\[
\bar{U}_{b}^{\mathrm{corr}}
=
\frac{1}{|\mathcal{I}_{\mathrm{corr}}|}
\sum_{i \in \mathcal{I}_{\mathrm{corr}}}
\widetilde{U}_{b}(x_i),
\]
\[
\bar{U}_{b}^{\mathrm{err}}
=
\frac{1}{|\mathcal{I}_{\mathrm{err}}|}
\sum_{i \in \mathcal{I}_{\mathrm{err}}}
\widetilde{U}_{b}(x_i).
\]
We also compute the uncertainty gap
\[
\Delta U_b
=
\bar{U}_{b}^{\mathrm{err}}
-
\bar{U}_{b}^{\mathrm{corr}},
\]
where larger positive values indicate that the uncertainty score assigns higher uncertainty
to misclassified examples. To evaluate whether uncertainty predicts model failure, we compute error-detection AUROC:
\[
\mathrm{AUROC}_{\mathrm{err}}(U_b)
=
\mathrm{AUROC}
\left(
\left\{
\mathbb{I}[\hat{y}_i \neq y_i]
\right\}_{i=1}^{N},
\left\{
\widetilde{U}_b(x_i)
\right\}_{i=1}^{N}
\right),
\]
where $U_b$ is one of $U_{\mathrm{total}}$, $U_{\mathrm{alea}}$, or
$U_{\mathrm{epi}}$. A higher value indicates that the corresponding uncertainty component
better identifies examples on which the model is likely to be incorrect.

\paragraph{Summary of novelties and differences to prior work.} SPECTRA preserves C-LoRA's lightweight Bayesianization by restricting stochasticity to $E^\ell_{x,t}\in\mathbb{R}^{r\times r}$, keeping $A^\ell$ and $B^\ell$ deterministic and avoiding the $O(d)$-scaled Bayesian overhead typical of Bayesianizing full LoRA factors. Unlike BLoB, the posterior is amortized and input-dependent, yielding sample-specific heteroscedastic uncertainty. 
Unlike C-LoRA, which conditions the contextual posterior only on internal low-rank features $z^\ell$, SPECTRA additionally incorporates per-layer, token-level text-audio cross-attention over audio frames, enabling uncertainty to reflect temporally localized cross-modal evidence while remaining parameter-efficient.

\section{Further Experimental Results}
\label{app:full_results}


This section describes the experimental protocol used to evaluate the proposed SPECTRA framework and its relationship to prior contextualized low-rank adaptation methods. Our goal is to assess whether token-level cross-modal contextualization improves parameter-efficient multimodal adaptation compared to both text-only PEFT methods and global audio conditioning approaches.

\paragraph{Models and Modalities.} We evaluate SPECTRA using multiple combinations of pre-trained text and audio encoders to ensure robustness across backbone architectures. For the text modality, we consider six sentence-level transformer encoders widely used for semantic representation learning: nliRoBERTa, paraTinyBERT, allMiniLM12, nliDistilRoBERTa, allDistilRoBERTa, and allRoBERTaLarge. These models provide diverse trade-offs between representational capacity and computational cost.

For the audio modality, we employ three large pre-trained speech representation models: whisperMedium, hubertLargeFT, and wav2vec2LargeFT. These encoders produce high-level acoustic embeddings that capture prosody, speaker characteristics, and paralinguistic cues.

\paragraph{Compared Methods.} We compare three families of approaches representing different levels of multimodal integration and parameter-efficient adaptation:

\begin{enumerate}
    \item \textbf{Text-only PEFT.} such as LoRA and C-LoRA that adapt the frozen LLM using low-rank updates on text only (no audio).
    \item \textbf{Multimodal transfer learning with feature fusion.} we obtain a pooled audio embedding from an audio encoder and fuse it with the text representation (e.g., concatenation), followed by a task-specific classification head (3-layer MLP, 32 units followed by 16 units followed by a softmax). This is a conventional lightweight multimodal baseline that injects audio at the representation/head level.
    \item \textbf{Multimodal context-conditioned PEFT (ours).} instead of heavy feature fusion, we use audio as context that conditions the variational low-rank adapter uncertainty and adaptation throughout the transformer.
\end{enumerate} 

\paragraph{SPECTRA Variants and Ablations.}
\label{sec:caliber_variants}

To better understand the contribution of the proposed contextualized adapter mechanism, we evaluate several variants of SPECTRA that differ in how audio information is incorporated into the LoRA adaptation process. These ablations allow us to isolate the effect of cross-modal contextualization and identify which architectural components contribute most to performance improvements.

\begin{enumerate}
    \item \textbf{SPECTRA-X (Cross-Attention).} The primary variant uses token-level cross-attention between text tokens and frame-level audio representations within each LoRA layer (as in Figure \ref{fig_architecture}). Text token representations act as queries while audio frames provide keys and values. The resulting contextual signal is projected into the LoRA latent space and used to condition the variational adapter parameters. This design enables fine-grained alignment between linguistic tokens and acoustic events.
    \item \textbf{SPECTRA-X (Shared-KV).} To reduce parameter overhead, we evaluate a variant where the key and value projections of the cross-attention module are shared across transformer layers. In this configuration, the audio representation is projected once at the model level and reused by all LoRA layers, while each layer maintains its own query projection and output mapping. This design significantly reduces the number of additional parameters while preserving layer-specific contextualization.
    \item \textbf{SPECTRA-G (Global Context).} We use a single global audio embedding to condition the adapters. Comparing SPECTRA-G and -X isolates the benefit of moving from global contextualization to token-level cross-modal alignment.
\end{enumerate}

Together, these variants allow us to analyze the impact of architectural choices on performance, parameter efficiency, and multimodal representation quality.

\paragraph{Training and Evaluation Protocol.} All PEFT methods, including LoRA, C-LoRA, and the SPECTRA variants, use rank $r=8$ and scaling factor $\alpha=32$. Only the adapter parameters are updated, while all text and audio backbone encoders remain frozen. Models are trained for 50 epochs using AdamW with learning rate $10^{-3}$ and weight decay $10^{-3}$. For SPECTRA-G, pooled audio embeddings are projected into a 16-dimensional context space using a shared MLP, followed by layer-specific linear mappings into the 8-dimensional latent adapter space. For SPECTRA-X, frame-level audio representations are used instead of pooled embeddings, and a lightweight multi-head cross-attention module performs token-level alignment between text tokens and audio frames in a low-dimensional latent space. The Bayesian adapter formulation follows the same variational parameterization as C-LoRA, with prior and regularization hyperparameters set to $\beta=0.2$, $\varepsilon=0.05$, and $\gamma=0.008$. During inference, predictive uncertainty is estimated using Monte Carlo sampling with $M=10$ stochastic forward passes. All experiments are conducted using 5-fold cross-validation with speaker-level separation to prevent leakage between training and test folds. Performance is reported using area under the ROC curve (AUC\%), which is appropriate for the imbalanced label distributions in the clinical interview dataset. Final results are averaged across the five folds for each task.

\subsection{Experimenal Results}
\label{sec:exp_results}

\paragraph{Modality disagreement and calibration stress tests.}
\label{app:stress_test_details}

A central motivation of SPECTRA is that audio should not be treated merely as an additional feature stream, but also as a contextual reliability signal that modulates adaptation and predictive confidence. In ordinary multimodal evaluation, however, the text and audio modalities are naturally paired and usually contain mutually consistent evidence. Such evaluation measures predictive performance under matched conditions, but it does not directly test whether the model responds appropriately when one modality becomes unreliable or contradicts the other. To evaluate this behavior, we introduce a modality-disagreement stress test that perturbs the relationship between the textual and acoustic inputs at test time. Let the held-out evaluation set be $\mathcal{D}_{\mathrm{test}} = \{(x_i^{\mathrm{text}},x_i^{\mathrm{audio}},y_i)\}_{i=1}^{N}$, where \(x_i^{\mathrm{text}}\) denotes the transcript, \(x_i^{\mathrm{audio}}\) denotes the corresponding audio segment, and \(y_i\) is the original label. We evaluate the same trained SPECTRA model under four input conditions: $\mathcal{C} = \{c_{\mathrm{clean}}, c_{\mathrm{audio\text{-}noise}}, c_{\mathrm{audio\text{-}mismatch}}, c_{\mathrm{text\text{-}corrupt}}\}$.

The clean condition preserves the original paired input as $c_{\mathrm{clean}}: (x_i^{\mathrm{text}},x_i^{\mathrm{audio}})$. The noisy-audio condition preserves the correct text-audio pairing but corrupts the acoustic signal as $c_{\mathrm{audio\text{-}noise}}: (x_i^{\mathrm{text}},\tau_{\mathrm{noise}}(x_i^{\mathrm{audio}}))$, where \(\tau_{\mathrm{noise}}(\cdot)\) denotes the audio corruption operator. This condition tests robustness to degraded but still semantically matched acoustic evidence. The mismatched-audio condition preserves the clean transcript but replaces the corresponding audio with an audio segment from another held-out example as $c_{\mathrm{audio\text{-}mismatch}}: (x_i^{\mathrm{text}},x_{\pi(i)}^{\mathrm{audio}})$ with $\pi(i) \neq i$. This condition intentionally destroys the correspondence between linguistic and acoustic evidence, thereby testing whether the model over-relies on misleading audio context. Finally, the corrupted-text condition perturbs the transcript while keeping the original audio clean as $c_{\mathrm{text\text{-}corrupt}}: (\tau_{\mathrm{text}}(x_i^{\mathrm{text}}),x_i^{\mathrm{audio}})$, where \(\tau_{\mathrm{text}}(\cdot)\) denotes the text corruption operator. This condition evaluates whether clean acoustic information can partially compensate when the lexical stream is degraded. Importantly, the target label \(y_i\) is kept fixed in all conditions. Therefore, performance differences reflect the model's sensitivity to modality reliability rather than changes in the prediction task.

For each condition \(c\), we compute the posterior predictive distribution using Monte Carlo model averaging over the stochastic adapter latent: $p_{i,c}^{(m)} = p_{\theta}\left(y \mid x_{i,c}^{\mathrm{text}}, x_{i,c}^{\mathrm{audio}}, E_{i,c}^{(m)}\right)$, with $E_{i,c}^{(m)} \sim q_{\phi}\left(E \mid x_{i,c}^{\mathrm{text}}, x_{i,c}^{\mathrm{audio}}\right)$, and $\bar{p}_{i,c} = \frac{1}{M}\sum_{m=1}^{M} p_{i,c}^{(m)}$. We then evaluate discrimination, probabilistic accuracy, and calibration using AUC, negative log-likelihood (NLL), expected calibration error (ECE), and Brier score: $\mathrm{NLL}_{c} = -\frac{1}{N}\sum_{i=1}^{N} \log \bar{p}_{i,c,y_i}$ and $\mathrm{Brier}_{c} = \frac{1}{N}\sum_{i=1}^{N}\sum_{k=1}^{K}\left(\bar{p}_{i,c,k} - \mathbbm{1}\{y_i=k\}\right)^2$.

For calibration, we define the predicted confidence as $\mathrm{conf}_{i,c} = \max_{k}\bar{p}_{i,c,k}$ and partition the predictions into \(B\) confidence bins \(\{\mathcal{B}_b\}_{b=1}^{B}\). The expected calibration error is $\mathrm{ECE}_{c} = \sum_{b=1}^{B}\frac{|\mathcal{B}_b|}{N}\left|\mathrm{acc}(\mathcal{B}_b) - \mathrm{conf}(\mathcal{B}_b)\right|$, where $\mathrm{acc}(\mathcal{B}_b) = \frac{1}{|\mathcal{B}_b|}\sum_{i \in \mathcal{B}_b}\mathbbm{1}\left\{y_i = \arg\max_k \bar{p}_{i,c,k}\right\}$ and $\mathrm{conf}(\mathcal{B}_b) = \frac{1}{|\mathcal{B}_b|}\sum_{i \in \mathcal{B}_b}\mathrm{conf}_{i,c}$.

\paragraph{Uncertainty decomposition.}
To determine whether modality perturbations primarily affect irreducible data uncertainty or model uncertainty, we decompose predictive uncertainty into total, aleatoric, and epistemic components. Let $H(p) = -\sum_{k=1}^{K} p_k \log p_k$ denote predictive entropy. The normalized total uncertainty is $U^{\mathrm{total}}_{i,c} = \frac{H(\bar{p}_{i,c})}{\log K}$. The normalized aleatoric component is estimated by the expected entropy of individual stochastic predictions as $U^{\mathrm{alea}}_{i,c} = \frac{1}{M}\sum_{m=1}^{M}\frac{H(p_{i,c}^{(m)})}{\log K}$. The epistemic component is then estimated by the mutual information between predictions and the stochastic adapter latent as $U^{\mathrm{epi}}_{i,c} = U^{\mathrm{total}}_{i,c} - U^{\mathrm{alea}}_{i,c}$.
This decomposition is particularly relevant for SPECTRA, because the model's stochasticity is concentrated in the low-rank adapter-space latent \(E\). Hence, increases in \(U^{\mathrm{epi}}\) indicate that the model's predictions are more sensitive to posterior samples of the adapter, while increases in \(U^{\mathrm{alea}}\) indicate that the individual sampled predictors are themselves less confident.

We also measure the aggregate posterior dispersion of the adapter latent as $U^{E}_{i,c} = \frac{1}{L T_i r^2}\sum_{\ell=1}^{L}\sum_{t=1}^{T_i}\sum_{j=1}^{r^2}\sigma^{\ell}_{E,t,j}\left(x_{i,c}^{\mathrm{text}}, x_{i,c}^{\mathrm{audio}}\right)$,
where \(\sigma^{\ell}_{E,t,j}\) denotes the posterior standard deviation of the \(j\)-th entry of \(\mathrm{vec}(E^{\ell}_{i,t})\). This quantity measures whether modality perturbations directly increase the inferred posterior scale in the Bayesian adapter space.

\begin{table}[t]
\centering
\caption{
Modality-disagreement stress-test results. Higher AUC is better, while lower NLL, ECE, and Brier score are better. The same held-out examples are evaluated under each condition.
}
\label{app:tab:modality_disagreement}
\begin{tabular}{lcccc}
\toprule
\textbf{Condition} & \textbf{AUC} \(\uparrow\) & \textbf{NLL} \(\downarrow\) & \textbf{ECE} \(\downarrow\) & \textbf{Brier} \(\downarrow\) \\
\midrule
Clean text + matched audio     & 0.835 & 0.419 & 0.042 & 0.260 \\
Clean text + noisy audio       & 0.812 & 0.435 & 0.021 & 0.278 \\
Clean text + mismatched audio  & 0.716 & 0.560 & 0.087 & 0.346 \\
Corrupted text + clean audio   & 0.795 & 0.458 & 0.049 & 0.290 \\
\bottomrule
\end{tabular}
\end{table}

\begin{figure}[t]
\centering
\includegraphics[width=0.9\textwidth]{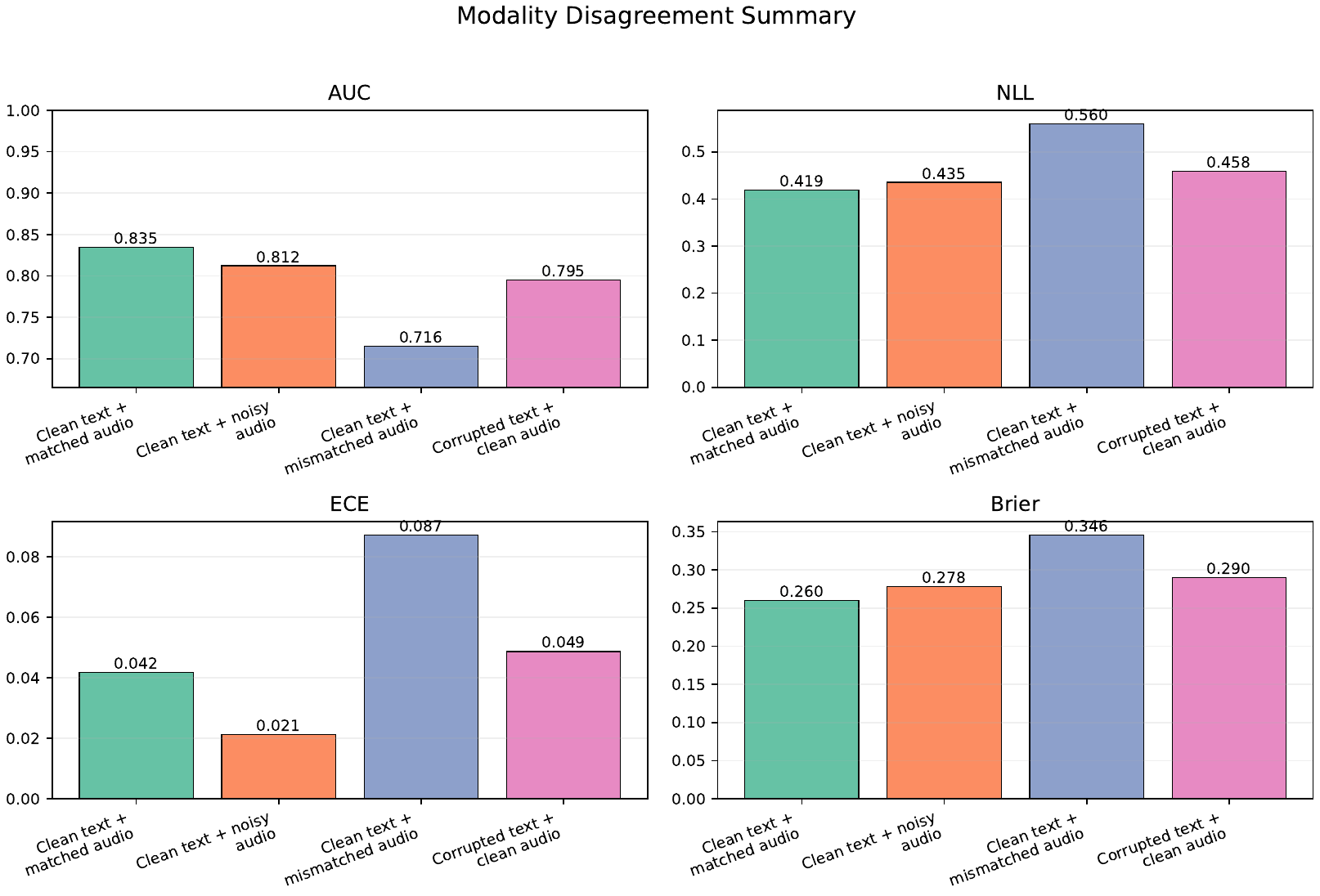}
\caption{
Summary of predictive performance under modality-disagreement stress tests on FORBOW clinical dataset. Clean matched text-audio pairs achieve the best overall performance. Noisy audio causes a moderate decrease in AUC and mild increases in NLL and Brier score. Mismatched audio produces the largest degradation across all metrics, indicating that contradictory cross-modal evidence is substantially more harmful than simple acoustic degradation. Corrupted text with clean audio leads to an intermediate degradation, suggesting that clean acoustic information partially compensates for degraded lexical input.
}
\label{app:fig:modality_disagreement_summary}
\end{figure}

\paragraph{Discrimination and probabilistic performance.}
Table~\ref{app:tab:modality_disagreement} and Figure~\ref{app:fig:modality_disagreement_summary} show that the clean matched condition achieves the strongest overall performance, with an AUC of \(0.835\), NLL of \(0.419\), ECE of \(0.042\), and Brier score of \(0.260\). This represents the standard evaluation setting where textual and acoustic evidence are correctly paired.

When the audio is corrupted with noise but remains correctly paired with the text, AUC decreases moderately from \(0.835\) to \(0.812\), and both NLL and Brier score increase slightly. This indicates that acoustic degradation reduces discriminative and probabilistic quality, but the model remains relatively robust because the linguistic stream is still clean and the audio retains its original semantic correspondence. Interestingly, ECE decreases from \(0.042\) to \(0.021\). This should not be interpreted as an overall improvement in prediction quality, since AUC, NLL, and Brier score all deteriorate. Rather, the lower ECE suggests that the model becomes more conservative under noisy audio, producing confidence estimates that are better aligned with empirical accuracy even though the predictions themselves are less accurate.

The most severe degradation occurs under clean text with mismatched audio. AUC drops to \(0.716\), NLL increases to \(0.560\), ECE increases to \(0.087\), and Brier score increases to \(0.346\). This condition is substantially harder than noisy audio because the acoustic modality is not merely degraded; it is actively misleading. The model receives clean lexical evidence but audio context that belongs to another example. The large drop in AUC and the sharp increases in NLL and Brier score show that incorrect cross-modal pairing can harm both ranking quality and probability estimation. The increase in ECE further indicates that mismatched audio produces residual overconfidence: the model sometimes assigns high confidence to predictions whose empirical accuracy is lower than expected.

Under corrupted text with clean audio, performance degrades less severely than under mismatched audio. AUC decreases from \(0.835\) to \(0.795\), NLL increases to \(0.458\), ECE increases slightly to \(0.049\), and Brier score rises to \(0.290\). This suggests that clean acoustic information can partially compensate for degraded lexical evidence, but not fully recover the performance of the clean matched condition. The asymmetry between the mismatched-audio and corrupted-text conditions is informative: misleading audio is more damaging than degraded text in this setting, suggesting that the model uses audio context strongly enough that incorrect acoustic evidence can distort the posterior adapter behavior.

\begin{figure}[t]
\centering
\includegraphics[width=0.8\textwidth]{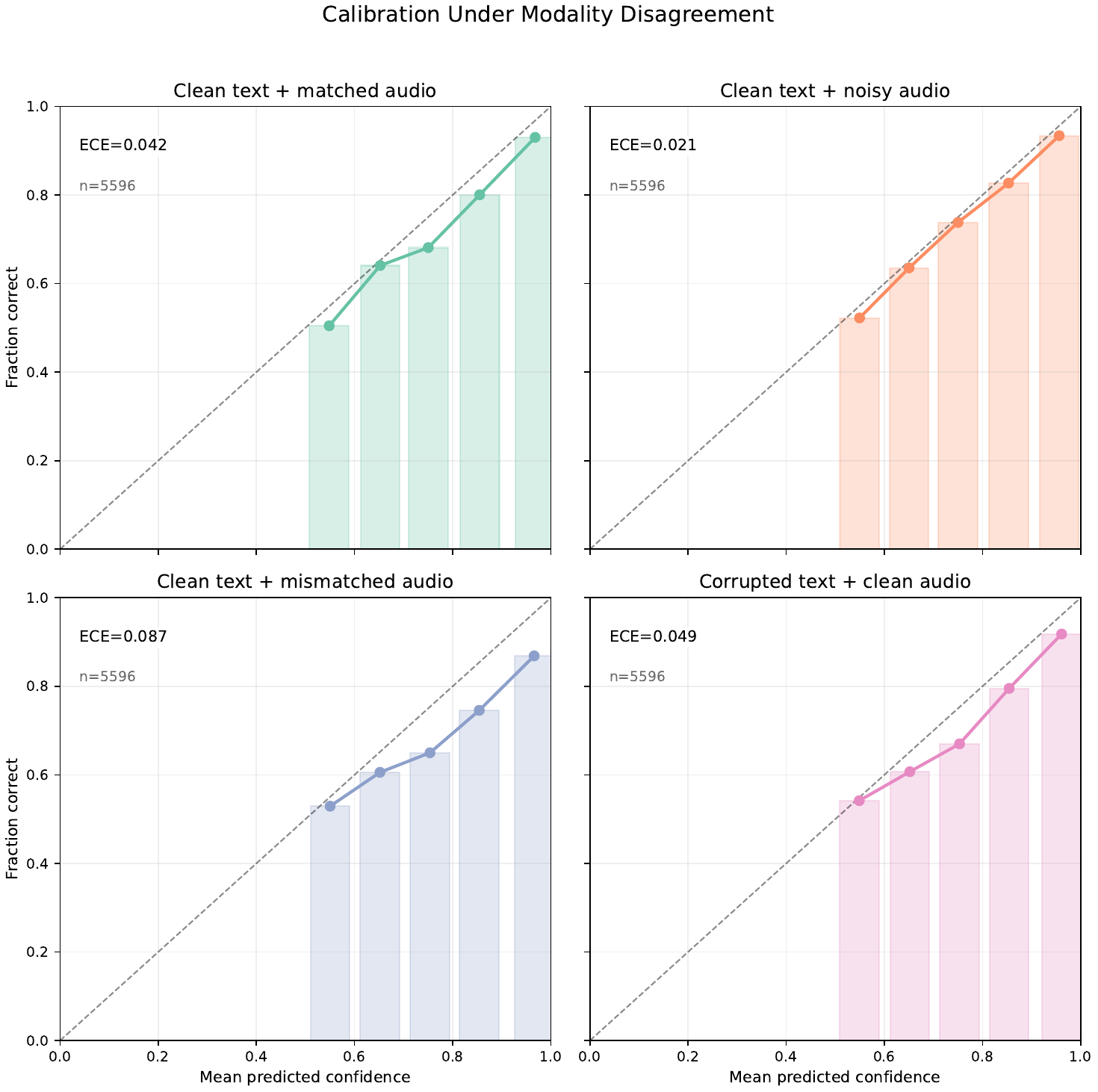}
\caption{
Reliability diagrams under modality disagreement on FORBOW clinical dataset. Each panel reports calibration over the same \(N=5596\) held-out examples. The clean matched condition is well calibrated overall. Noisy audio produces the lowest ECE, suggesting that the model becomes more conservative under acoustic degradation. Mismatched audio leads to the largest ECE and clear overconfidence in high-confidence bins, showing that semantically inconsistent audio is more harmful to calibration than noisy but correctly paired audio. Corrupted text with clean audio produces only a mild calibration degradation.
}
\label{app:fig:calibration_disagreement}
\end{figure}

\paragraph{Calibration under modality disagreement.}
Figure~\ref{app:fig:calibration_disagreement} shows reliability diagrams for all four conditions. In the clean matched setting, predicted confidence generally follows empirical accuracy, yielding an ECE of \(0.042\). This indicates that SPECTRA's posterior predictive probabilities are reasonably calibrated when the modalities are correctly paired.

The noisy-audio condition achieves the lowest ECE, \(0.021\), despite lower AUC and higher NLL than the clean condition. The reliability curve remains close to the diagonal, suggesting that the model appropriately reduces confidence when the acoustic signal becomes less reliable. This behavior is desirable in uncertainty-aware multimodal learning: degradation in one modality should not only affect the predicted label, but should also be reflected in predictive confidence.

The mismatched-audio condition exhibits the worst calibration, with ECE increasing to \(0.087\). The reliability curve lies noticeably below the diagonal in high-confidence regions, indicating overconfidence under cross-modal contradiction. This is an important finding because it reveals a failure mode that is not visible from standard clean evaluation alone. Although SPECTRA incorporates audio as a posterior-conditioning signal, the model can still be miscalibrated when the audio is plausible but semantically inconsistent with the text. Therefore, mismatched audio provides a stronger test of cross-modal reliability than noisy audio.

The corrupted-text condition yields ECE \(0.049\), only slightly worse than the clean matched condition. This suggests that the model remains moderately calibrated when the lexical stream is degraded but the audio remains valid. The reliability curve is closer to the diagonal than in the mismatched-audio setting, supporting the interpretation that valid acoustic evidence helps stabilize predictions when text quality decreases.

\begin{figure}[t]
\centering
\includegraphics[width=0.98\textwidth]{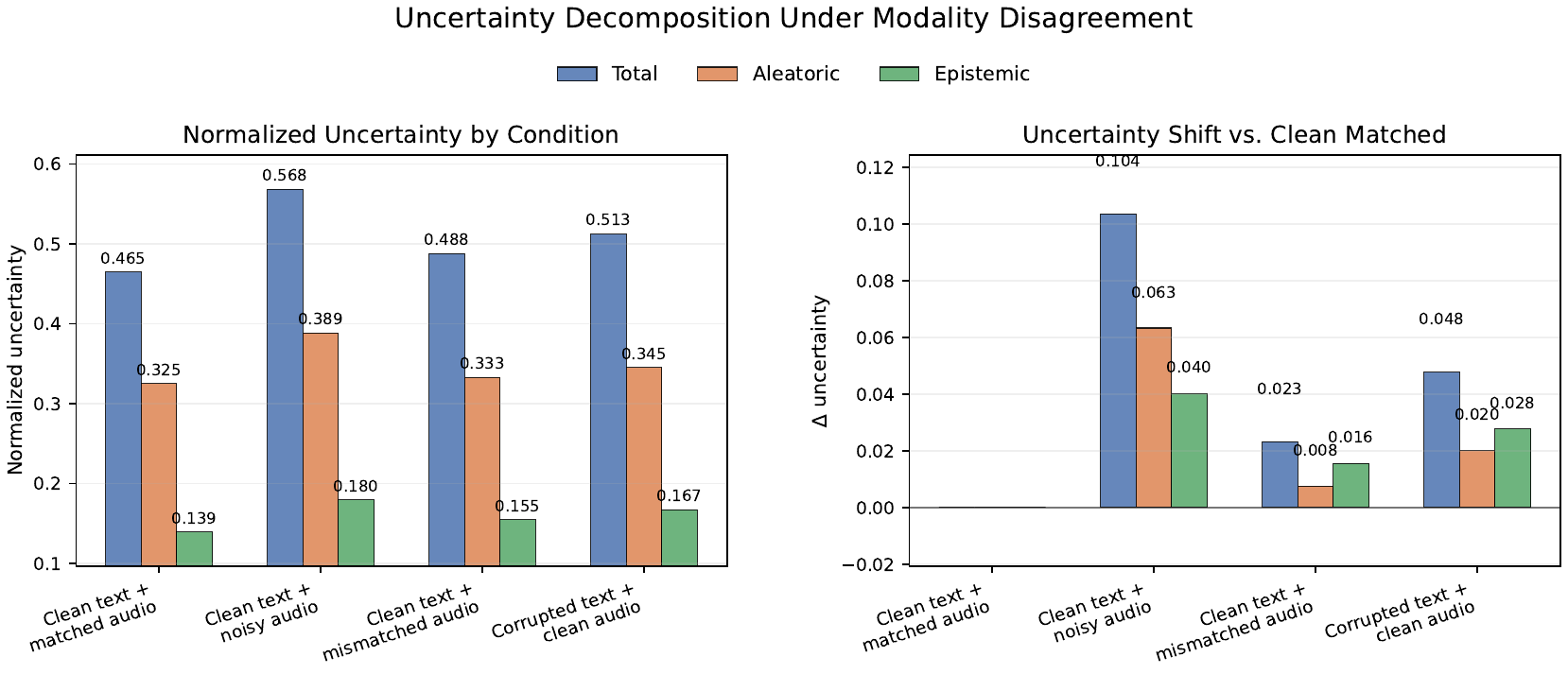}
\caption{
Uncertainty decomposition under modality disagreement on FORBOW clinical dataset. Left: normalized total, aleatoric, and epistemic uncertainty for each condition. Right: uncertainty shift relative to the clean matched condition. Noisy audio produces the largest total uncertainty increase, with increases in both aleatoric and epistemic components. Corrupted text with clean audio yields an intermediate increase, especially in epistemic uncertainty. Mismatched audio produces the largest performance degradation but only a modest uncertainty increase, revealing a residual overconfidence failure mode under semantically inconsistent audio.
}
\label{app:fig:uncertainty_decomposition}
\end{figure}

\paragraph{Uncertainty decomposition.}
Figure~\ref{app:fig:uncertainty_decomposition} decomposes predictive uncertainty into total, aleatoric, and epistemic components. In the clean matched condition, normalized total uncertainty is \(0.465\), with aleatoric uncertainty \(0.325\) and epistemic uncertainty \(0.139\). These values provide the reference point for interpreting the perturbation-induced uncertainty shifts.

Noisy audio produces the largest increase in total uncertainty, from \(0.465\) to \(0.568\). This corresponds to a shift of \(+0.104\), decomposed into \(+0.063\) aleatoric and \(+0.040\) epistemic uncertainty. This pattern is consistent with the nature of the perturbation: noise reduces the clarity of the acoustic evidence, making individual sampled predictors less confident, while also increasing disagreement across posterior adapter samples. The simultaneous increase in aleatoric and epistemic uncertainty suggests that the model recognizes noisy audio as both less informative and less stable for adaptation.

Mismatched audio increases total uncertainty more modestly, from \(0.465\) to \(0.488\), with an epistemic shift of \(+0.016\). This indicates that the model detects some additional posterior uncertainty under cross-modal conflict, but the increase is smaller than the degradation observed in AUC, NLL, ECE, and Brier score. This mismatch between performance degradation and uncertainty increase is important: it suggests that semantically incorrect audio can mislead the model without always producing a sufficiently large uncertainty response. In other words, mismatched audio exposes a residual overconfidence failure mode.

Corrupted text with clean audio produces an intermediate increase in total uncertainty, from \(0.465\) to \(0.513\), with aleatoric and epistemic shifts of \(+0.020\) and \(+0.028\), respectively. The larger epistemic component relative to the aleatoric shift suggests that corrupting the lexical input increases uncertainty about how the adapter should adapt, while clean audio still provides useful evidence that prevents the larger degradation observed under mismatched audio.

Overall, the decomposition shows that all modality perturbations increase predictive uncertainty relative to the clean matched condition. However, the type of uncertainty depends on the perturbation. Noisy audio primarily increases both aleatoric and epistemic uncertainty, corrupted text produces a moderate epistemic increase, and mismatched audio causes the largest performance and calibration degradation despite only a modest uncertainty increase.

\paragraph{Predictive entropy and adapter posterior response.}
Perturbed conditions shift the uncertainty distribution relative to the clean matched baseline, consistent with the aggregate decomposition in Figure~\ref{app:fig:uncertainty_decomposition}.

The right panel reports the mean posterior standard deviation of the stochastic adapter latent \(E\). The aggregate value remains nearly constant across conditions, approximately \(0.047\)-\(0.048\). This result is informative. It suggests that the observed changes in predictive uncertainty are not primarily driven by a large global increase in the average posterior scale \(\sigma_E\). Instead, the uncertainty response may arise from changes in posterior means, downstream class probabilities, token-specific effects that are averaged out by the global statistic, or localized layer-wise changes in \(\sigma_E\) that are not visible after averaging over all layers, tokens, and rank-space dimensions.

This finding also highlights an important direction for improving SPECTRA. Although the current model produces meaningful predictive uncertainty shifts under modality perturbations, the aggregate adapter posterior variance does not strongly separate clean, noisy, mismatched, and text-corrupted inputs. A more explicit reliability-aware posterior could encourage the adapter variance itself to respond monotonically to cross-modal reliability. For example, one could introduce a learned reliability score \(\rho_{t}^{\ell}(x)\in[0,1]\) and define $\sigma_{E,t}^{\ell}(x)^2 = \sigma_{\phi,t}^{\ell}(x)^2 + \lambda\left(1-\rho_{t}^{\ell}(x)\right)$, so that unreliable or contradictory cross-modal evidence directly increases the posterior variance in the adapter space. The present stress test therefore serves two purposes: it validates that SPECTRA's predictive uncertainty responds to modality degradation, and it reveals that future versions should make the adapter posterior variance more explicitly reliability-sensitive.


\paragraph{Discussion.}
The modality-disagreement experiments provide a stronger test of uncertainty-aware multimodal adaptation than standard clean evaluation. The results support three main conclusions. First, SPECTRA is robust to moderate acoustic degradation: noisy audio reduces AUC and increases NLL and Brier score, but the model remains comparatively well calibrated and increases predictive uncertainty. Second, semantically mismatched audio is substantially more harmful than noisy but correctly paired audio. This confirms that the model uses cross-modal context in a meaningful way, but it also reveals a calibration failure mode under contradictory modalities. Third, clean audio partially compensates for corrupted text, as the corrupted-text condition performs better than the mismatched-audio condition and remains reasonably calibrated.

These results strengthen the central claim that audio can function as a reliability signal for Bayesian low-rank adaptation. At the same time, they show that the current version of SPECTRA is more responsive at the predictive-distribution level than at the aggregate adapter-posterior-variance level. This distinction is important: a model can become less confident in its class probabilities without necessarily increasing the global posterior standard deviation of the adapter latent. Therefore, future extensions should incorporate explicit cross-modal reliability estimation, layer-wise variance diagnostics, and disagreement-aware posterior regularization to ensure that modality conflict is reflected not only in predictive entropy but also in the Bayesian adapter posterior itself.

\begin{table}[tp]
\centering
\caption{AUC (\%) on a 5-fold cross-validation on segment-level predictions on offspring data.}
\label{app:tbl_offspring}
\setlength\tabcolsep{2.75pt}
\scalebox{0.79}{
\hspace{-1.6cm}
\begin{tabular}{lllcccccccccccc}
\toprule
 & & & \rot{\textbf{sentiment}} & \rot{\textbf{anger}} & \rot{\textbf{fear}} & \rot{\textbf{joy}} & \rot{\textbf{sadness}} & \rot{\textbf{richness}} & \rot{\textbf{coherence}} & \rot{\textbf{rumination}} & \rot{\textbf{worry}} & \rot{\textbf{anxiousness}} & \rot{\textbf{criticism}} & \rot{\textbf{self-criticism}} \\
Method & LLM & Audio Encoder &  &  &  &  &  &  &  &  &  &  &  &  \\
\midrule
LoRA & nRoBERTa & - & 93.88 & 87.36 & 73.03 & 83.36 & 81.42 & 83.99 & 78.82 & 51.05 & 64.98 & 66.81 & 93.12 & 93.82 \\
C-LoRA & nRoBERTa & - & 93.38 & 85.77 & 80.65 & 80.00 & 82.73 & 82.57 & 77.17 & 66.81 & 87.35 & 64.91 & 91.66 & 90.02 \\
\cmidrule[0.05pt](lr){1-3}
\multirow[t]{3}{*}{Transfer} & \multirow[t]{3}{*}{nRoBERTa} & whisperM & 93.19 & \textbf{88.37} & 87.90 & \textbf{86.63} & 84.88 & 84.22 & 82.34 & 83.46 & 87.93 & 80.90 & 92.52 & 92.58 \\
 &  & wav2vec2LFT & 93.03 & 87.48 & 73.82 & 83.59 & 82.31 & 83.54 & 80.51 & 73.44 & 86.72 & 75.24 & 91.31 & 91.70 \\
 &  & hubertLFT & 92.59 & 87.13 & 86.24 & 83.29 & 82.11 & 83.39 & 79.64 & 79.07 & 87.03 & 72.57 & 90.84 & 92.21 \\
\cmidrule[0.05pt](lr){1-3}
\multirow[t]{3}{*}{SPECTRA-G} & \multirow[t]{3}{*}{nRoBERTa} & whisperM & 94.65 & 86.16 & 88.83 & 84.16 & 73.73 & 83.83 & 81.36 & 75.56 & 76.99 & 63.75 & 92.20 & 90.19 \\
 &  & wav2vec2LFT & 94.06 & 86.05 & 80.95 & 81.00 & 83.53 & 83.91 & 78.79 & 76.33 & 87.51 & 70.86 & 91.26 & 92.98 \\
 &  & hubertLFT & 94.10 & 86.95 & 83.55 & 81.32 & 79.61 & 83.25 & 78.36 & 84.15 & 87.50 & 65.72 & 93.27 & 91.05 \\
\cmidrule[0.05pt](lr){1-3}
\multirow[t]{3}{*}{SPECTRA-X} & \multirow[t]{3}{*}{nRoBERTa} & whisperM & 94.27 & 86.21 & \textbf{90.12} & 84.38 & 73.81 & 84.91 & 81.92 & 75.86 & 77.25 & 64.4 & 92.46 & 91.57 \\
 &  & wav2vec2LFT & 93.54 & 86.52 & 81.17 & 81.34 & 84.45 & 84.8 & 79.29 & 77.91 & 88.76 & 71.89 & 91.95 & \textbf{94.31} \\
 &  & hubertLFT & 93.91 & 87.05 & 83.8 & 81.86 & 80.02 & 84.17 & 79.03 & 85.43 & 87.71 & 66.72 & \textbf{94.71} & 91.61 \\
\midrule
LoRA & pTinyBERT & - & 90.92 & 53.15 & 50.25 & 51.92 & 50.30 & 72.80 & 72.50 & 60.17 & 50.03 & 52.86 & 50.00 & 50.00 \\
C-LoRA & pTinyBERT & - & 91.79 & 65.82 & 51.92 & 53.92 & 52.36 & 76.32 & 75.83 & 52.63 & 51.53 & 55.78 & 62.60 & 55.07 \\
\cmidrule[0.05pt](lr){1-3}
\multirow[t]{3}{*}{Transfer} & \multirow[t]{3}{*}{pTinyBERT} & whisperM & 90.86 & 86.08 & 84.29 & 85.66 & 83.69 & 85.09 & 82.22 & 83.73 & 83.40 & 79.47 & 90.32 & 89.98 \\
 &  & wav2vec2LFT & 90.78 & 84.83 & 74.03 & 81.31 & 79.98 & 82.67 & 79.56 & 75.44 & 82.01 & 71.68 & 87.18 & 88.43 \\
 &  & hubertLFT & 90.54 & 84.60 & 70.41 & 81.87 & 78.13 & 80.51 & 80.38 & 78.71 & 84.84 & 72.00 & 88.27 & 86.56 \\
\cmidrule[0.05pt](lr){1-3}
\multirow[t]{3}{*}{SPECTRA-G} & \multirow[t]{3}{*}{pTinyBERT} & whisperM & \textbf{95.06} & 63.41 & 55.11 & 54.37 & 58.71 & 78.99 & 81.53 & 53.00 & 58.51 & 56.63 & 67.66 & 54.78 \\
 &  & wav2vec2LFT & 94.86 & 62.45 & 55.37 & 57.72 & 63.98 & 78.69 & 82.25 & 53.00 & 55.00 & 66.73 & 75.55 & 61.74 \\
 &  & hubertLFT & 94.64 & 79.64 & 53.00 & 56.72 & 64.00 & 79.37 & 81.32 & 54.50 & 53.54 & 60.57 & 75.86 & 56.03 \\
\cmidrule[0.05pt](lr){1-3}
\multirow[t]{3}{*}{SPECTRA-X} & \multirow[t]{3}{*}{pTinyBERT} & whisperM & 94.66 & 64.41 & 56.13 & 54.97 & 59.21 & 79.38 & 82.04 & 53.64 & 59.56 & 57.44 & 68.61 & 54.91 \\
 &  & wav2vec2LFT & 94.11 & 63.59 & 56.06 & 58.45 & 64.32 & 79.59 & 82.92 & 54.03 & 55.54 & 67.06 & 76.25 & 62.47 \\
 &  & hubertLFT & 94.32 & 80.37 & 52.66 & 57.16 & 64.89 & 80.17 & 81.91 & 55.6 & 54.01 & 61.56 & 76.58 & 56.62 \\
\midrule
LoRA & aMiniLM12 & - & 80.51 & 50.00 & 50.00 & 50.00 & 52.97 & 71.40 & 69.07 & 59.62 & 50.20 & 51.42 & 50.00 & 50.00 \\
C-LoRA & aMiniLM12 & - & 91.64 & 52.23 & 53.25 & 51.19 & 51.87 & 74.24 & 72.82 & 51.43 & 50.60 & 51.24 & 51.29 & 51.25 \\
\cmidrule[0.05pt](lr){1-3}
\multirow[t]{3}{*}{Transfer} & \multirow[t]{3}{*}{aMiniLM12} & whisperM & 75.96 & 79.29 & 80.12 & 80.81 & 75.73 & \textbf{85.11} & 81.98 & 68.45 & 75.09 & 79.84 & 79.47 & 83.45 \\
 &  & wav2vec2LFT & 81.76 & 75.10 & 72.52 & 71.75 & 64.75 & 78.89 & 79.10 & 62.52 & 66.45 & 72.02 & 77.42 & 77.76 \\
 &  & hubertLFT & 80.84 & 74.22 & 69.86 & 74.06 & 66.94 & 79.33 & 79.18 & 68.26 & 66.13 & 70.23 & 74.85 & 76.94 \\
\cmidrule[0.05pt](lr){1-3}
\multirow[t]{3}{*}{SPECTRA-G} & \multirow[t]{3}{*}{aMiniLM12} & whisperM & 94.70 & 57.13 & 53.79 & 54.47 & 54.13 & 77.39 & 81.28 & 53.98 & 53.89 & 58.39 & 53.19 & 53.80 \\
 &  & wav2vec2LFT & 94.10 & 53.99 & 53.00 & 55.45 & 54.23 & 77.43 & 83.44 & 53.27 & 58.60 & 64.31 & 53.72 & 53.08 \\
 &  & hubertLFT & 94.63 & 55.76 & 53.37 & 55.27 & 56.06 & 77.52 & 83.06 & 53.69 & 53.00 & 59.03 & 55.67 & 53.43 \\
\cmidrule[0.05pt](lr){1-3}
\multirow[t]{3}{*}{SPECTRA-X} & \multirow[t]{3}{*}{aMiniLM12} & whisperM & 94.94 & 58.73 & 54.85 & 54.77 & 53.76 & 78.03 & 82.29 & 55.24 & 54.2 & 59.28 & 54.07 & 54.58 \\
 &  & wav2vec2LFT & 93.85 & 54.81 & 54.42 & 55.84 & 54.41 & 77.37 & 83.72 & 54.22 & 59.42 & 65.28 & 54.35 & 53.07 \\
 &  & hubertLFT & 94.21 & 56.89 & 53.77 & 56.04 & 57.31 & 78.21 & 83.85 & 53.8 & 53.54 & 59.49 & 55.93 & 54.16 \\
\midrule
LoRA & aDRoBERTa & - & 92.91 & 86.13 & 74.92 & 82.40 & 84.16 & 84.32 & 79.67 & 75.47 & 87.85 & 73.35 & 92.17 & 93.04 \\
C-LoRA & aDRoBERTa & - & 92.58 & 82.76 & 81.89 & 78.76 & 81.86 & 84.13 & 78.73 & 72.91 & 86.68 & 69.79 & 92.01 & 91.36 \\
\cmidrule[0.05pt](lr){1-3}
\multirow[t]{3}{*}{Transfer} & \multirow[t]{3}{*}{aDRoBERTa} & whisperM & 85.62 & 82.10 & 83.84 & 82.06 & 73.47 & 83.85 & 81.77 & 70.05 & 80.54 & \textbf{81.57} & 83.90 & 83.83 \\
 &  & wav2vec2LFT & 84.86 & 77.93 & 75.00 & 73.41 & 69.89 & 80.11 & 79.08 & 61.88 & 71.29 & 71.74 & 80.05 & 81.22 \\
 &  & hubertLFT & 84.47 & 78.35 & 70.11 & 76.10 & 71.19 & 79.01 & 78.18 & 68.18 & 65.96 & 71.80 & 78.71 & 76.86 \\
\cmidrule[0.05pt](lr){1-3}
\multirow[t]{3}{*}{SPECTRA-G} & \multirow[t]{3}{*}{aDRoBERTa} & whisperM & 93.65 & 81.82 & 78.77 & 83.41 & 84.27 & 84.75 & 81.39 & 85.99 & 87.54 & 66.68 & 93.94 & 93.04 \\
 &  & wav2vec2LFT & 93.59 & 85.56 & 76.64 & 80.41 & 83.06 & 84.68 & 82.59 & 75.14 & \textbf{89.28} & 75.99 & 93.93 & 93.32 \\
 &  & hubertLFT & 93.64 & 84.14 & 81.71 & 79.34 & 83.32 & 84.92 & 81.49 & 74.64 & 88.53 & 70.37 & 93.58 & 91.78 \\
\cmidrule[0.05pt](lr){1-3}
\multirow[t]{3}{*}{SPECTRA-X} & \multirow[t]{3}{*}{aDRoBERTa} & whisperM & 94.46 & 82.32 & 79.72 & 84.39 & \textbf{85.12} & 84.33 & 82.39 & \textbf{86.91} & 87.24 & 66.72 & 94.4 & 93.56 \\
 &  & wav2vec2LFT & 93.34 & 86.13 & 77.46 & 80.43 & 83.89 & 83.15 & \textbf{83.93} & 75.86 & 88.66 & 76.48 & 93.62 & 94.0 \\
 &  & hubertLFT & 94.39 & 84.87 & 82.58 & 80.3 & 83.52 & 84.43 & 82.53 & 75.58 & 88.9 & 71.55 & 94.44 & 92.28 \\
\bottomrule
\end{tabular}
}
\end{table}

Table~\ref{app:tbl_offspring} shows the evaluation results on the offspring data. Results on offspring data reveal a more heterogeneous pattern than the parent setting, but overall it still supports the value of multimodal contextualization, especially for the stronger text backbones. For nliRoBERTa and allDistillRoBERTa, SPECTRA remains consistently competitive and often achieves the best results on several clinically relevant tasks, including fear, self-criticism, criticism, sadness, rumination, and coherence. In particular, SPECTRA-X yields the top performance on 6 tasks, indicating that token-level cross-modal conditioning can again improve over both text-only PEFT and simple transfer-based fusion when the textual backbone is sufficiently expressive. At the same time, the offspring results also highlight a sharper dependence on backbone quality and task difficulty. For weaker encoders such as paraTinyBERT and allMiniLM12, SPECTRA substantially boosts sentiment into the mid-90s, but often degrades performance on sparse affective and psychological targets relative to the transfer baselines, suggesting that the additional flexibility of contextual Bayesian conditioning is harder to exploit when the underlying text representation is weak and the labels are extremely imbalanced. In these settings, strong transfer baselines with whisperMedium remain highly competitive. Overall, the offspring results suggest that SPECTRA is most effective when paired with sufficiently strong language encoders and when the task benefits from localized cross-modal reliability cues, whereas for weaker backbones or highly sparse targets, simpler multimodal fusion may sometimes provide a more stable inductive bias.

\begin{figure}[tp]
\centering
\makebox[\textwidth][c]{%
\includegraphics[width=\textwidth]{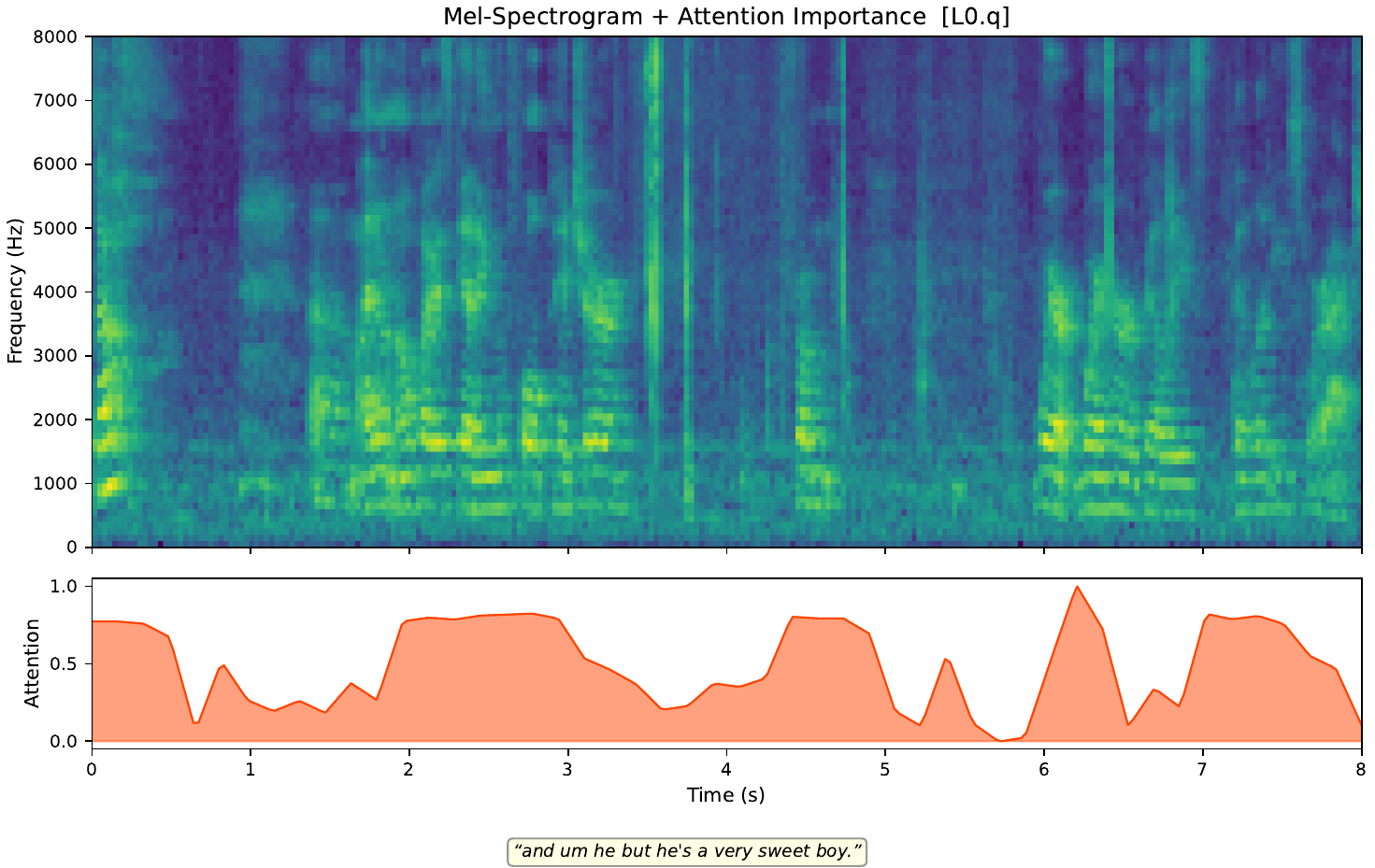}
}
\caption{\label{app:fig_spec_attn}Mel-spectrogram of a random datapoint with token-level cross-modal attention over time.}
\end{figure}

Figure \ref{app:fig_spec_attn} visualizes how the model allocates cross-modal attention over an input utterance. The upper panel shows the mel-spectrogram of the speech signal, while the lower panel shows the corresponding token-level attention importance across time. Peaks in the attention curve indicate temporal regions of the audio that the model considers more informative when conditioning the text representation. This illustrates that the proposed cross-attention mechanism is able to focus selectively on localized acoustic segments rather than relying on a single global audio summary.

\begin{figure}[tp]
\centering
\makebox[\textwidth][c]{%
\includegraphics[width=\textwidth]{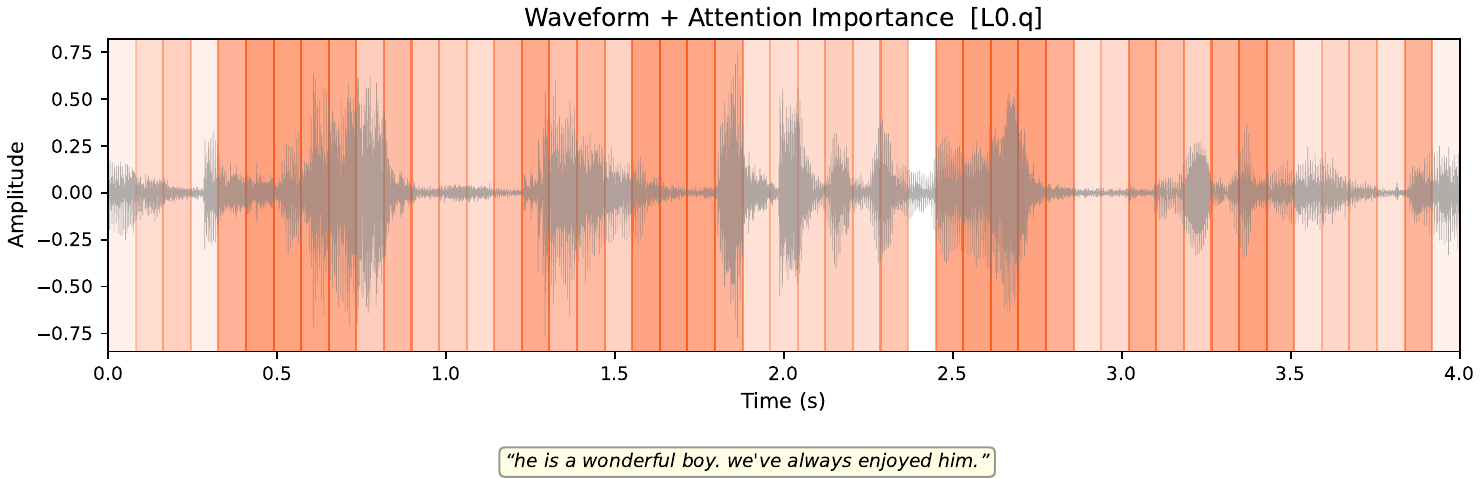}
}
\caption{\label{app:fig_wave_attn}Waveform of a random datapoint with token-level cross-modal attention over time.}
\end{figure}

Figure \ref{app:fig_wave_attn} shows the temporal alignment between the speech waveform and the model's token-level cross-modal attention. The gray trace represents the input audio waveform, while the shaded orange regions indicate the relative attention assigned to different time intervals. Stronger shading highlights portions of the utterance that the model considers more informative when conditioning the text representation. This visualization suggests that the proposed cross-attention mechanism selectively emphasizes localized acoustic regions, supporting the claim that SPECTRA uses fine-grained audio context rather than a single pooled summary.

\section{Complementary Information on FORBOW Dataset}
\label{sec_supp_data}

In this section, we provide more details on the collected data. As mentioned in the main manuscript, we transcribed and broke down each sample into multiple segments based on changes in emotion, sentiment, objectivity/subjectivity, etc. Average word count in a segment is 17 and average audio length for a segment is 6.47 seconds. Six multidisciplinary researchers rated these segments for subjectivity, emotions, sentiment, criticism, etc. 5,818 segments were rated by two or more researchers and the intraclass correlation for ratings of different researchers was high showing strong agreement in the labeling. Table \ref{app:tbl_dataset_info} show detailed information about the dataset.

\begin{table}[tp]
\centering
\caption{Dataset information for the segment-level tasks in parent and offspring data.}
\label{app:tbl_dataset_info}
\setlength\tabcolsep{2pt}
\scalebox{0.8}{
\begin{tabular}{llcccccccc}
\toprule
 & \textbf{Task} & \textbf{\# Samples} & \textbf{\# Classes} & \textbf{Imbalanced (\%)} & \textbf{Label 0} & \textbf{Label 1} & \textbf{Label 2} & \textbf{Label 3} \\
\midrule
Parent & objective & 23693 & 2 & 19.08 & 13096 & 10597 & - & - \\
Parent & sentiment & 23479 & 3 & 57.39 & 4725 & 11088 & 7666 & - \\
Parent & anger & 23694 & 2 & 92.22 & 21984 & 1710 & - & - \\
Parent & fear & 23694 & 2 & 95.29 & 22628 & 1066 & - & - \\
Parent & joy & 23690 & 2 & 66.58 & 17756 & 5934 & - & - \\
Parent & sadness & 23694 & 2 & 92.70 & 22081 & 1613 & - & - \\
Parent & neutral & 23693 & 2 & 16.28 & 10797 & 12896 & - & - \\
Parent & cohesion & 23694 & 2 & 81.32 & 19964 & 3730 & - & - \\
Parent & rumination & 23694 & 2 & 98.95 & 23447 & 247 & - & - \\
Parent & overinclusive & 23694 & 2 & 97.20 & 23049 & 645 & - & - \\
Parent & worry & 23694 & 2 & 91.76 & 21890 & 1804 & - & - \\
Parent & criticism & 23694 & 2 & 88.42 & 21235 & 2459 & - & - \\
\midrule
Offspring & objective & 10319 & 2 & 1.90 & 5209 & 5110 & - & - \\
Offspring & sentiment & 10222 & 3 & 52.51 & 2404 & 5062 & 2756 & - \\
Offspring & richness & 10273 & 3 & 97.08 & 1366 & 8654 & 253 & - \\
Offspring & reference & 10328 & 4 & 77.89 & 2072 & 4970 & 2187 & 1099 \\
Offspring & irrelevance & 10301 & 2 & 87.98 & 9196 & 1105 & - & - \\
Offspring & anger & 10329 & 2 & 95.55 & 9889 & 440 & - & - \\
Offspring & fear & 10329 & 2 & 99.04 & 10231 & 98 & - & - \\
Offspring & joy & 10329 & 2 & 94.49 & 9790 & 539 & - & - \\
Offspring & sadness & 10329 & 2 & 94.62 & 9802 & 527 & - & - \\
Offspring & neutral & 10328 & 2 & 80.92 & 1655 & 8673 & - & - \\
Offspring & coherence & 10329 & 2 & 68.35 & 7846 & 2483 & - & - \\
Offspring & rumination & 10329 & 2 & 99.01 & 10228 & 101 & - & - \\
Offspring & worry & 10329 & 2 & 99.13 & 10240 & 89 & - & - \\
Offspring & anxiousness & 10317 & 2 & 96.51 & 9969 & 348 & - & - \\
Offspring & aggression & 2182 & 2 & 99.49 & 2171 & 11 & - & - \\
Offspring & criticism & 8090 & 2 & 95.04 & 7708 & 382 & - & - \\
Offspring & self-criticism & 10329 & 2 & 96.77 & 10006 & 323 & - & - \\
\bottomrule
\end{tabular}
}
\end{table}

Researchers also rated affect, warmth, overprotection, cohesion, criticism, and worry at the document-level (i.e. for each audio sample). Document-level assessments are provided as nominal ratings. Table \ref{app:tbl_dataset_info2} shows the statistics of the dataset, all the tasks, and the distribution of labels. For certain tasks, some documents were missing the corresponding label, or some labels were extremely imbalanced such there were only 1 or 2 examples from one category, and we removed those cases. This is the reason for the differences in the number of samples in Table \ref{app:tbl_dataset_info2}. The imbalanced \% is calculated as $100 * (\max_{c\in C}n_c - \min_{c\in C}n_c) / \max_{c\in C}n_c$. As we can see most of the tasks are quite imbalanced.

In addition to the segment-level and document-level tasks that were labeled by the researchers, there are also psychological and cognitive tasks that are labeled by psychiatrists. The `spectrum' task is a 4-class categorization of normal/control, depression, Bipolar, and Schizophrenia. In addition to that, we have introvert, extrovert, ADHD, anxiety, and separate binary depression labels. The statistics of these tasks, the distribution of labels and imbalancedness are also shown in Table \ref{app:tbl_dataset_info2}.

\begin{table}[!t]
\centering
\caption{Dataset information for the document-level and psychological and cognitive prediction tasks in parent and offspring data.}
\label{app:tbl_dataset_info2}
\setlength\tabcolsep{3pt}
\scalebox{0.8}{
\begin{tabular}{lllcccccccc}
\toprule
& \textbf{Level} & \textbf{Task} & \rot{\textbf{Num Samples}} & \rot{\textbf{Num Classes}} & \rot{\textbf{Imbalanced (\%)}} & \rot{\textbf{Label 0}} & \rot{\textbf{Label 1}} & \rot{\textbf{Label 2}} & \rot{\textbf{Label 3}} & \rot{\textbf{Label 4}} \\
\midrule
Parent & Document & affect & 484 & 3 & 93.18 & 23 & 124 & 337 & - & - \\
Parent & Document & warmth & 485 & 3 & 39.69 & 117 & 194 & 174 & - & - \\
Parent & Document & overprotection & 487 & 3 & 85.77 & 267 & 182 & 38 & - & - \\
Parent & Document & cohesion & 492 & 5 & 92.63 & 14 & 14 & 132 & 190 & 142 \\
Parent & Document & criticism & 492 & 4 & 89.22 & 232 & 166 & 69 & 25 & - \\
Parent & Document & worry & 381 & 4 & 93.29 & 164 & 147 & 59 & 11 & - \\
\midrule
Parent & Cognitive & spectrum & 363 & 4 & 87.25 & 129 & 149 & 66 & 19 & - \\
Parent & Cognitive & introvert & 369 & 2 & 49.39 & 245 & 124 & - & - & - \\
Parent & Cognitive & extrovert & 369 & 2 & 63.84 & 271 & 98 & - & - & - \\
Parent & Cognitive & ADHD & 369 & 2 & 67.74 & 279 & 90 & - & - & - \\
Parent & Cognitive & anxiety & 369 & 2 & 51.81 & 249 & 120 & - & - & - \\
Parent & Cognitive & depression & 369 & 2 & 89.19 & 333 & 36 & - & - & - \\
\midrule
Offspring & Document & affect & 621 & 4 & 97.73 & 16 & 11 & 109 & 485 & - \\
Offspring & Document & coherence & 616 & 5 & 95.83 & 11 & 33 & 109 & 264 & 199 \\
Offspring & Document & richness & 614 & 3 & 96.70 & 82 & 515 & 17 & - & - \\
\midrule
Offspring & Cognitive & spectrum & 85 & 4 & 86.11 & 36 & 27 & 17 & 5 & - \\
Offspring & Cognitive & introvert & 85 & 2 & 39.62 & 53 & 32 & - & - & - \\
Offspring & Cognitive & extrovert & 85 & 2 & 53.45 & 58 & 27 & - & - & - \\
Offspring & Cognitive & ADHD & 85 & 2 & 53.45 & 58 & 27 & - & - & - \\
Offspring & Cognitive & anxiety & 85 & 2 & 39.62 & 53 & 32 & - & - & - \\
Offspring & Cognitive & depression & 85 & 2 & 83.56 & 73 & 12 & - & - & - \\
\bottomrule
\end{tabular}}
\end{table}




The summary of information and statistics about the segment-level, document-level, and psychological and cognitive tasks on parent and offspring data are shown in Tables \ref{app:tbl_dataset_info_parent} and \ref{app:tbl_dataset_info_offspring}, respectively

\begin{table*}[!t]
\centering
\caption{Summary statistics of the parent dataset used in this research. The table reports the number of samples, number of classes, class distribution, and degree of imbalance for each task across three levels: segment-level (fine-grained annotations per speech segment), document-level (ratings per full interview), and cognitive/psychological labels (diagnostic categories assigned by clinicians). The imbalance percentage is computed as $100 \times (\max_c n_c - \min_c n_c) / \max_c n_c$, where $n_c$ denotes the number of samples in class $c$. Missing or extremely sparse labels have been removed, resulting in varying sample sizes across tasks.}
\label{app:tbl_dataset_info_parent}
\scalebox{0.9}{
\begin{tabular}{llcccccccc}
\toprule
\textbf{Level} & \textbf{Task} & \rot{\textbf{Num Samples}} & \rot{\textbf{Num Classes}} & \rot{\textbf{Imbalanced (\%)}} & \rot{\textbf{Label 0}} & \rot{\textbf{Label 1}} & \rot{\textbf{Label 2}} & \rot{\textbf{Label 3}} & \rot{\textbf{Label 4}} \\
\midrule
Segment & objective & 23693 & 2 & 19.08 & 13096 & 10597 & - & - & - \\
Segment & sentiment & 23479 & 3 & 57.39 & 4725 & 11088 & 7666 & - & - \\
Segment & anger & 23694 & 2 & 92.22 & 21984 & 1710 & - & - & - \\
Segment & fear & 23694 & 2 & 95.29 & 22628 & 1066 & - & - & - \\
Segment & joy & 23690 & 2 & 66.58 & 17756 & 5934 & - & - & - \\
Segment & sadness & 23694 & 2 & 92.70 & 22081 & 1613 & - & - & - \\
Segment & neutral & 23693 & 2 & 16.28 & 10797 & 12896 & - & - & - \\
Segment & cohesion & 23694 & 2 & 81.32 & 19964 & 3730 & - & - & - \\
Segment & rumination & 23694 & 2 & 98.95 & 23447 & 247 & - & - & - \\
Segment & overinclusive & 23694 & 2 & 97.20 & 23049 & 645 & - & - & - \\
Segment & worry & 23694 & 2 & 91.76 & 21890 & 1804 & - & - & - \\
Segment & criticism & 23694 & 2 & 88.42 & 21235 & 2459 & - & - & - \\
\midrule
Document & affect & 484 & 3 & 93.18 & 23 & 124 & 337 & - & - \\
Document & warmth & 485 & 3 & 39.69 & 117 & 194 & 174 & - & - \\
Document & overprotection & 487 & 3 & 85.77 & 267 & 182 & 38 & - & - \\
Document & cohesion & 492 & 5 & 92.63 & 14 & 14 & 132 & 190 & 142 \\
Document & criticism & 492 & 4 & 89.22 & 232 & 166 & 69 & 25 & - \\
Document & worry & 381 & 4 & 93.29 & 164 & 147 & 59 & 11 & - \\
\midrule
Cognitive & spectrum & 363 & 4 & 87.25 & 129 & 149 & 66 & 19 & - \\
Cognitive & introvert & 369 & 2 & 49.39 & 245 & 124 & - & - & - \\
Cognitive & extrovert & 369 & 2 & 63.84 & 271 & 98 & - & - & - \\
Cognitive & ADHD & 369 & 2 & 67.74 & 279 & 90 & - & - & - \\
Cognitive & anxiety & 369 & 2 & 51.81 & 249 & 120 & - & - & - \\
Cognitive & depression & 369 & 2 & 89.19 & 333 & 36 & - & - & - \\
\bottomrule
\end{tabular}
}
\end{table*}

\begin{figure*}[!t]
	\centering
    \makebox[\linewidth][c]{%
    	\begin{subfigure}[b]{0.5\columnwidth}
    		\centering
    		\includegraphics[width=\linewidth]{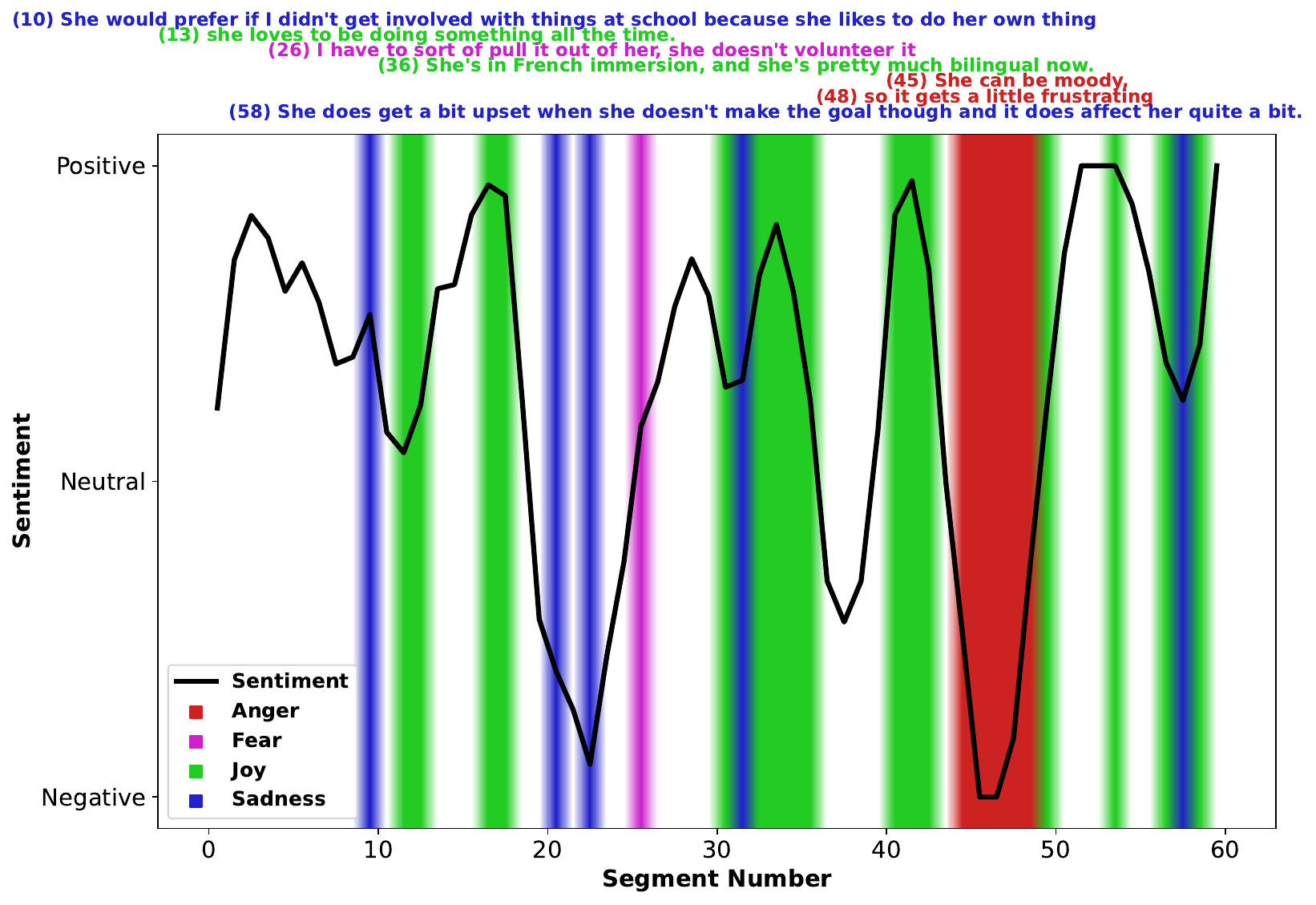}
            \caption{Sample 1}
    	\end{subfigure}
    	\begin{subfigure}[b]{0.5\columnwidth}
    		\centering
    		\includegraphics[width=\linewidth]{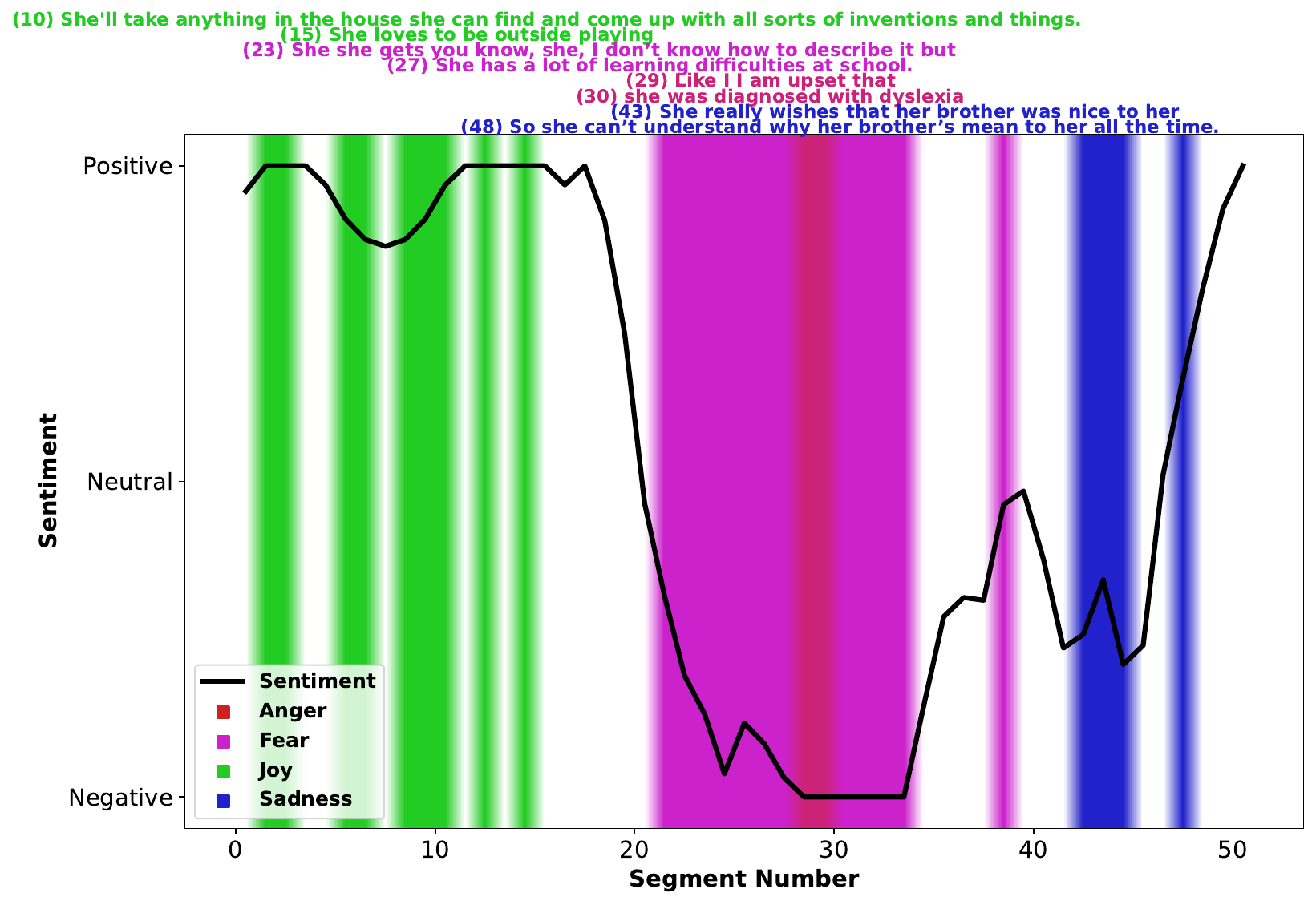}
            \caption{Sample 2}
    	\end{subfigure}
    }\\
    \makebox[\linewidth][c]{%
    	\begin{subfigure}[b]{0.5\columnwidth}
    		\centering
    		\includegraphics[width=\linewidth]{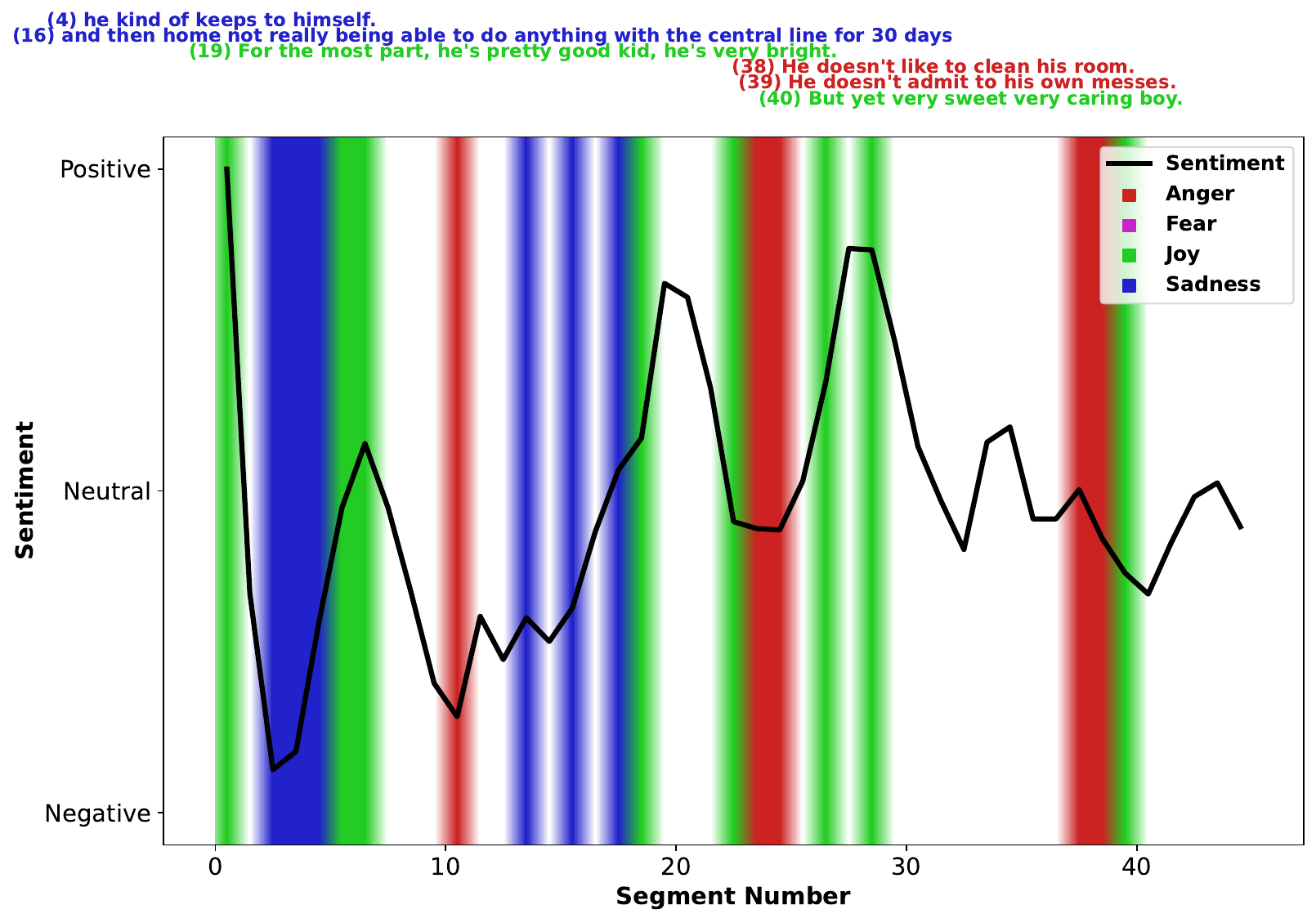}
            \caption{Sample 3}
    	\end{subfigure}
    	\begin{subfigure}[b]{0.5\columnwidth}
    		\centering
    		\includegraphics[width=\linewidth]{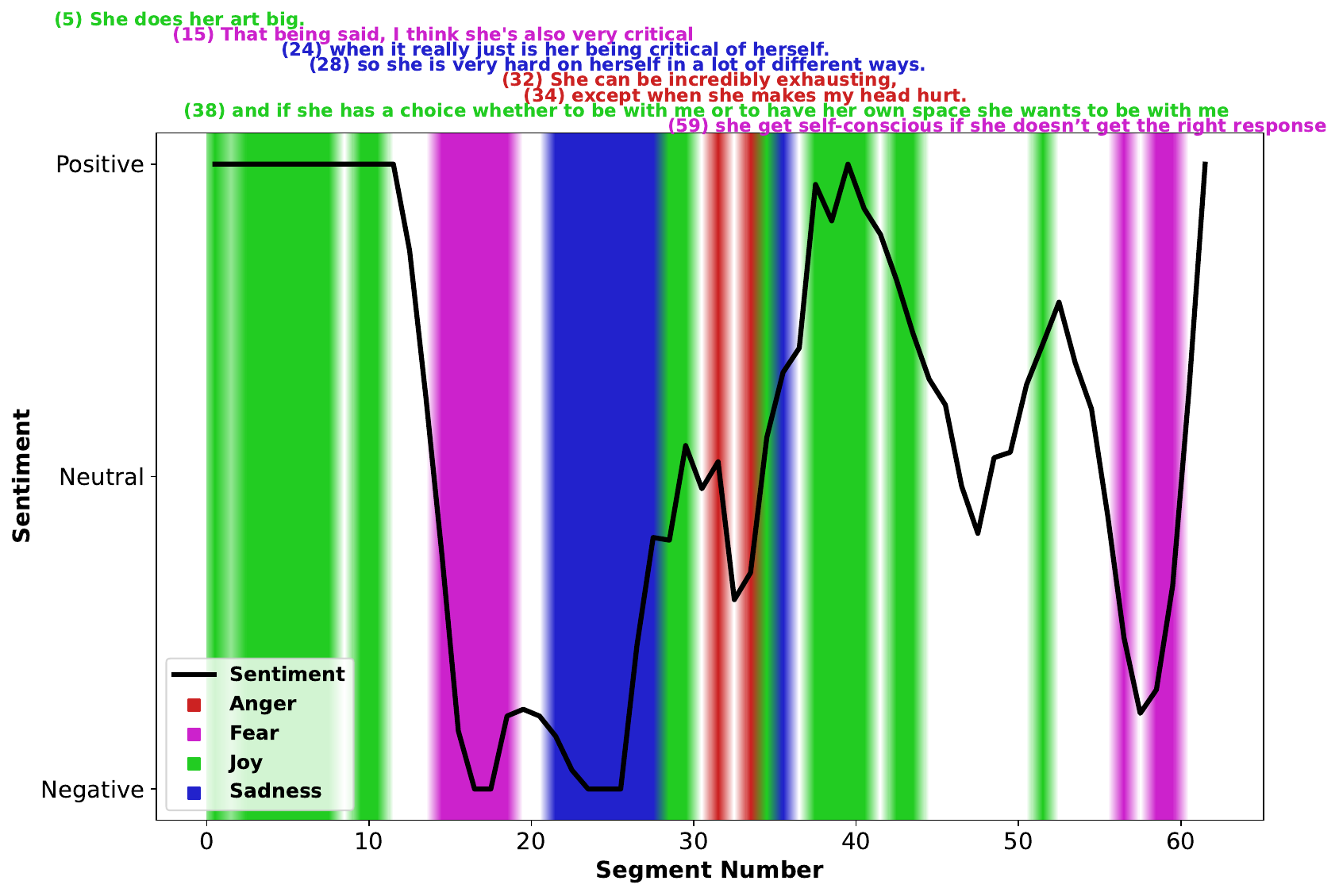}
            \caption{Sample 4}
    	\end{subfigure}
    }
    \caption{\label{fig_mood_parent}Emotion and sentiment dynamics for four representative parent interview samples from the FORBOW dataset. Each subplot corresponds to a different interview. The horizontal axis represents time (segmented speech units), and the vertical axis represents normalized sentiment scores. The black curve shows segment-level sentiment trajectories, while colored vertical bands indicate discrete emotion labels (anger, fear, joy, sadness, or neutral) assigned to each segment. Segments are obtained via content-aware partitioning based on changes in affective and linguistic cues. Representative text excerpts are displayed above each plot to illustrate the correspondence between raw speech content, segmentation boundaries, and assigned annotations.}
\end{figure*}

Figure \ref{fig_mood_parent} illustrates the changes in mood, emotions, and sentiment in four random samples in the parent data. We can also see sample segments in those files shown at the top of the figures which gives an idea of how the segmentation and the labeling is done. To protect participant privacy, all text excerpts shown throughout this paper are paraphrased to remove potentially identifying details. These modifications do not affect the underlying segmentation, emotion labels, or sentiment trajectories used in the analysis.

\begin{figure}[tp]
	\centering
    \makebox[\linewidth][c]{%
    	\begin{subfigure}[b]{0.9\textwidth}
    		\centering
    		\includegraphics[width=0.99\textwidth]{./figs/dataset/mood_parent_800-0036-060_M1_2017_CH_ch.pdf}
            \caption{Ground truth emotion and sentiment for a random sample}
    	\end{subfigure}
    }\\
    \makebox[\linewidth][c]{%
    	\begin{subfigure}[b]{0.9\textwidth}
    		\centering
    		\includegraphics[width=0.99\textwidth]{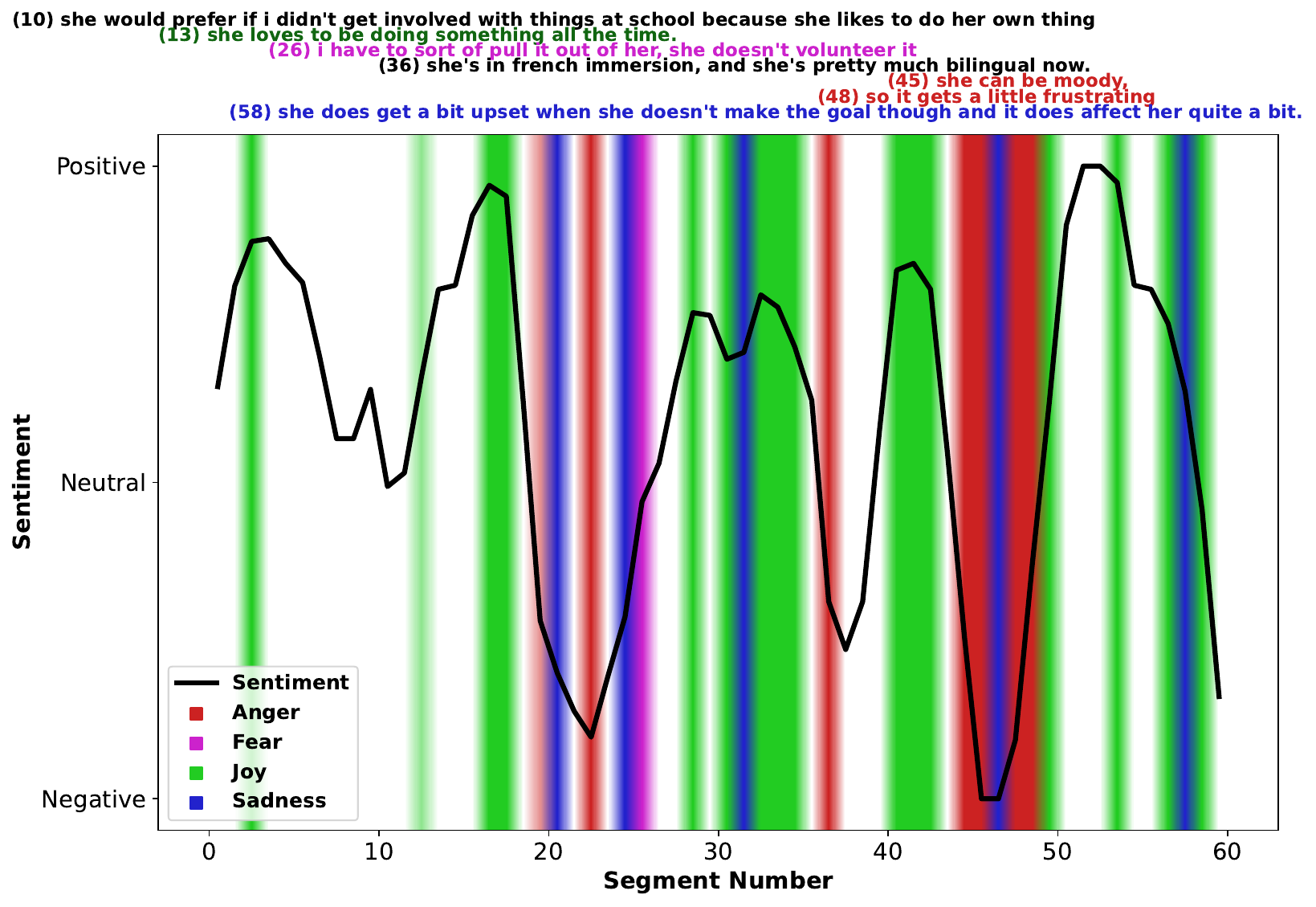}
            \caption{Predicted emotion and sentiment for the same random sample}
    	\end{subfigure}
    }
	\caption{\label{fig_mood_parent_predicted}Comparison of ground-truth and predicted segment-level sentiment and emotion trajectories for a representative parent interview using sentenceT5XL.}
\end{figure}

A comparison between the ground-truth and predicted sentiment-emotion trajectories for sentenceT5XL (the best-performing text-based model) on a random sample is illustrated in Figure \ref{fig_mood_parent_predicted}. The model captures the overall sentiment patterns across segments effectively, including major polarity shifts. The predicted emotion signals are generally consistent with the underlying semantic content, although some variations in fine-grained emotional labeling can be observed. This is expected given that predictions are made independently at the segment level using only textual information. Overall, the results demonstrate that text-based transfer learning provides a strong baseline for segment-level affect modeling, while leaving room for potential improvements through the incorporation of additional modalities such as audio.

\begin{figure*}[!t]
	\centering
    \hspace{-1cm}
    \makebox[\linewidth][c]{
	   \includegraphics[width=0.9\textwidth]{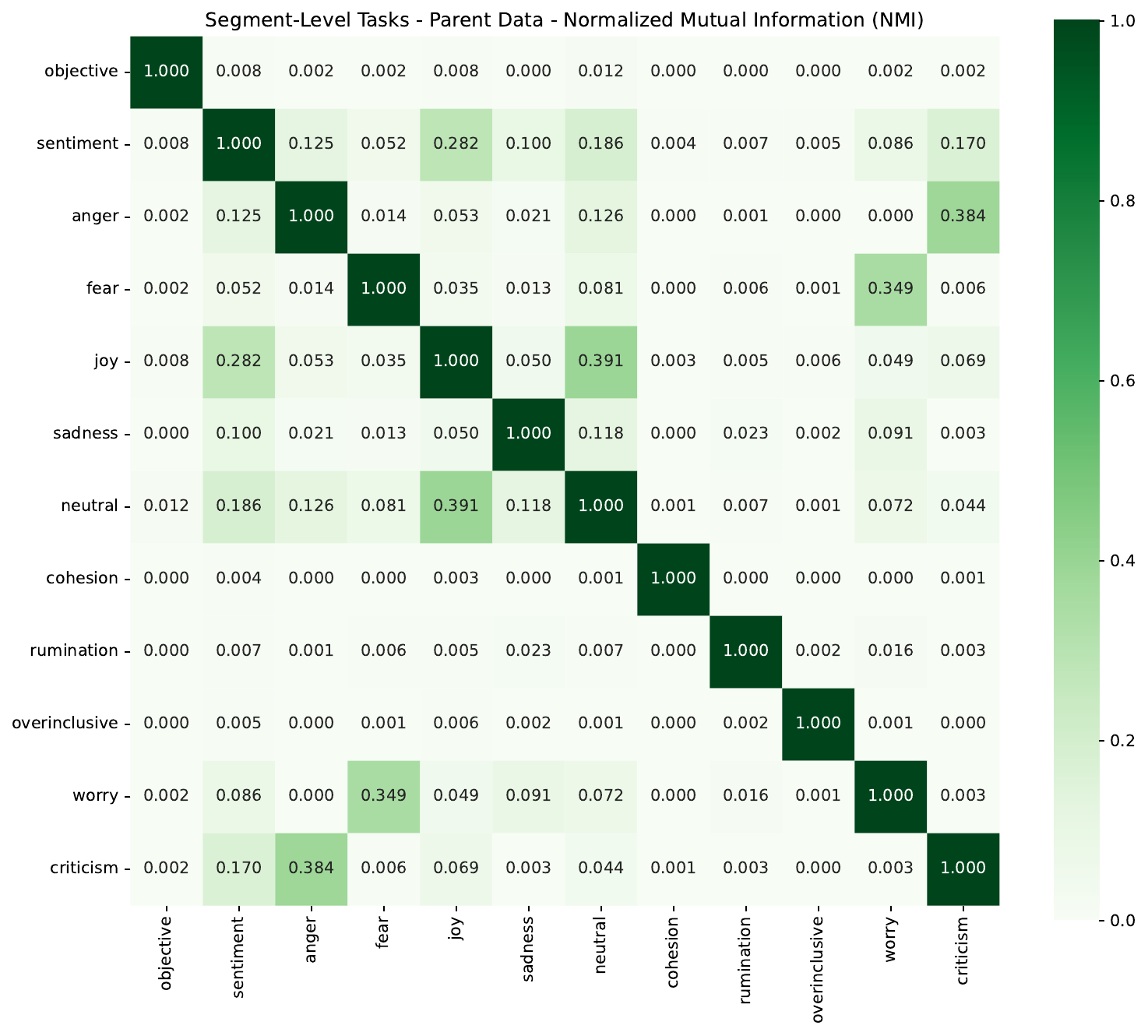}
    }
    \caption{Pairwise Normalized Mutual Information (NMI) between segment-level annotation tasks for parent interview data. Each cell represents the NMI value between two tasks, quantifying the degree of statistical dependency between their label distributions. Higher values indicate stronger relationships between tasks (e.g., co-occurrence or conceptual overlap), while lower values indicate more independent labeling behavior. All values are computed using segment-level annotations across the full parent dataset.}
	\label{fig_dataset_nmi_segment_parent}
\end{figure*}

Figure \ref{fig_dataset_nmi_segment_parent} shows the pairwise Normalized Mutual Information (NMI) between all the segment-level tasks on parent data. As we can see there are some labels that are correlated and have a fairly high mutual information in the parent data. For instance, `joy' and `sentiment` have an NMI of 0.282, `anger' and `criticism' have an NMI of 0.384, and `fear' and `worry' have an NMI of 0.349 which all makes sense as those tasks are intuitively correlated.

\begin{figure*}[!t]
	\centering
    \includegraphics[width=0.7\textwidth]{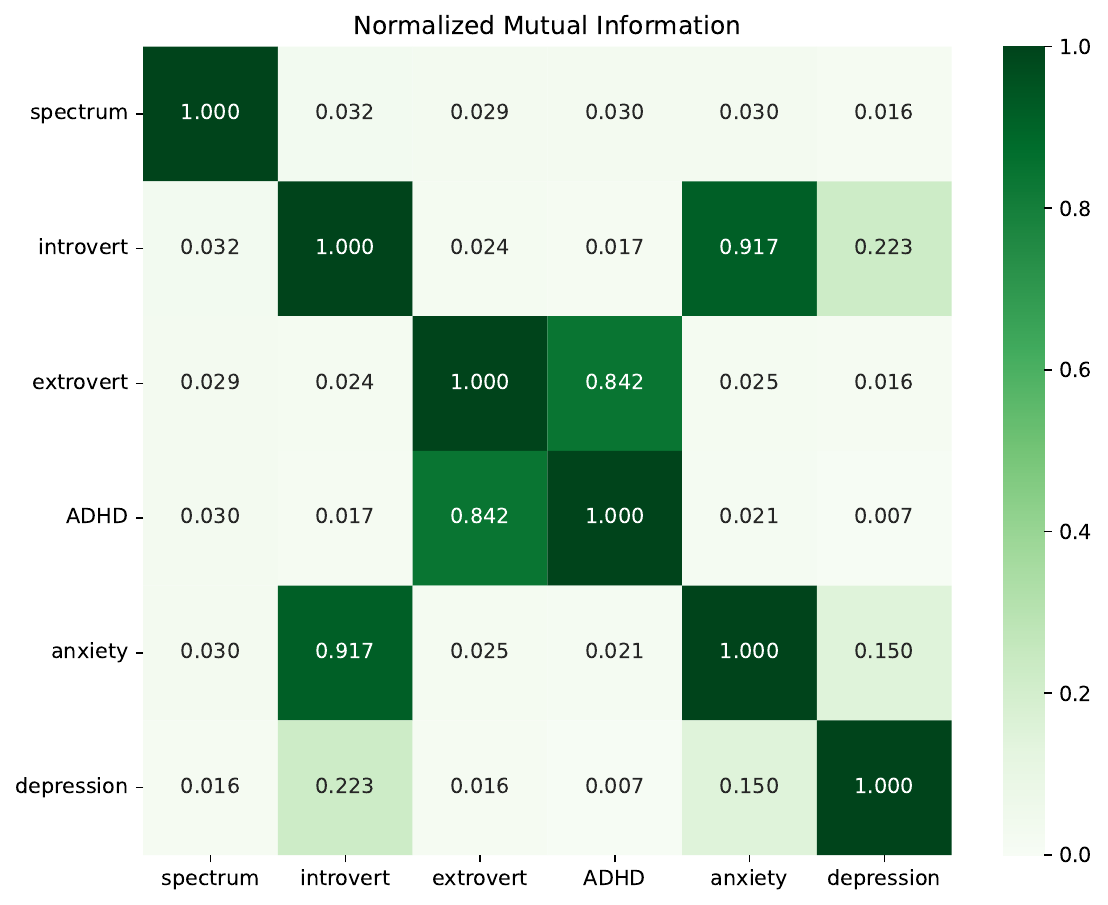}
    \caption{Pairwise Normalized Mutual Information (NMI) between psychological and cognitive labels derived from parent-level annotations. Each cell represents the dependency between two tasks, computed from subject-level labels. High NMI values indicate strong relationships or shared underlying constructs (e.g., introversion and anxiety), while low values indicate weak or independent associations. This matrix highlights redundancy and interdependence among higher-level diagnostic labels.}
	\label{fig_dataset_nmi_final}
\end{figure*}

Figure \ref{fig_dataset_nmi_final} illustrates the pairwise Normalized Mutual Information (NMI) between all the psychological and cognitive tasks. As we can see there are some tasks with very high mutual information. Particularly, `anxiety' and `introvert' have an NMI of 0.917 which is close to perfect mutual information. Similarly, `ADHD' and `extrovert' also have an NMI of 0.842 which is very high. `depression' and `introvert' have an NMI of 0.223 which is fairly high but not too strong. We can also observe that the `spectrum' task has a very low NMI with any other task which show there are many other factors into play and it is a complex task.

\begin{figure*}[!t]
	\centering
    \makebox[\linewidth][c]{%
    	\begin{subfigure}[b]{0.5\columnwidth}
    		\centering
    		\includegraphics[width=\linewidth]{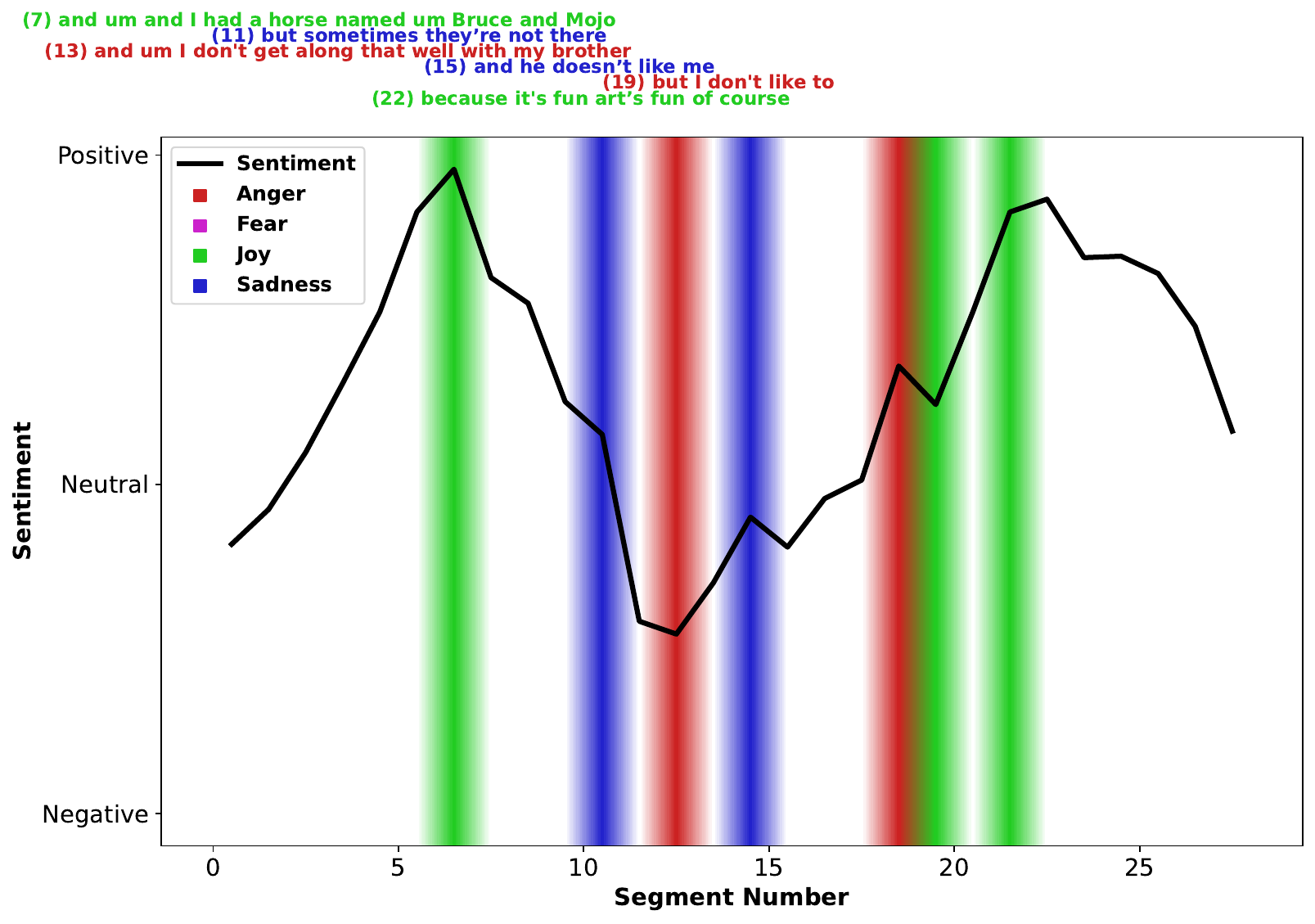}
            \caption{General interview, random sample 1}
    	\end{subfigure}
    	\begin{subfigure}[b]{0.5\columnwidth}
    		\centering
    		\includegraphics[width=\linewidth]{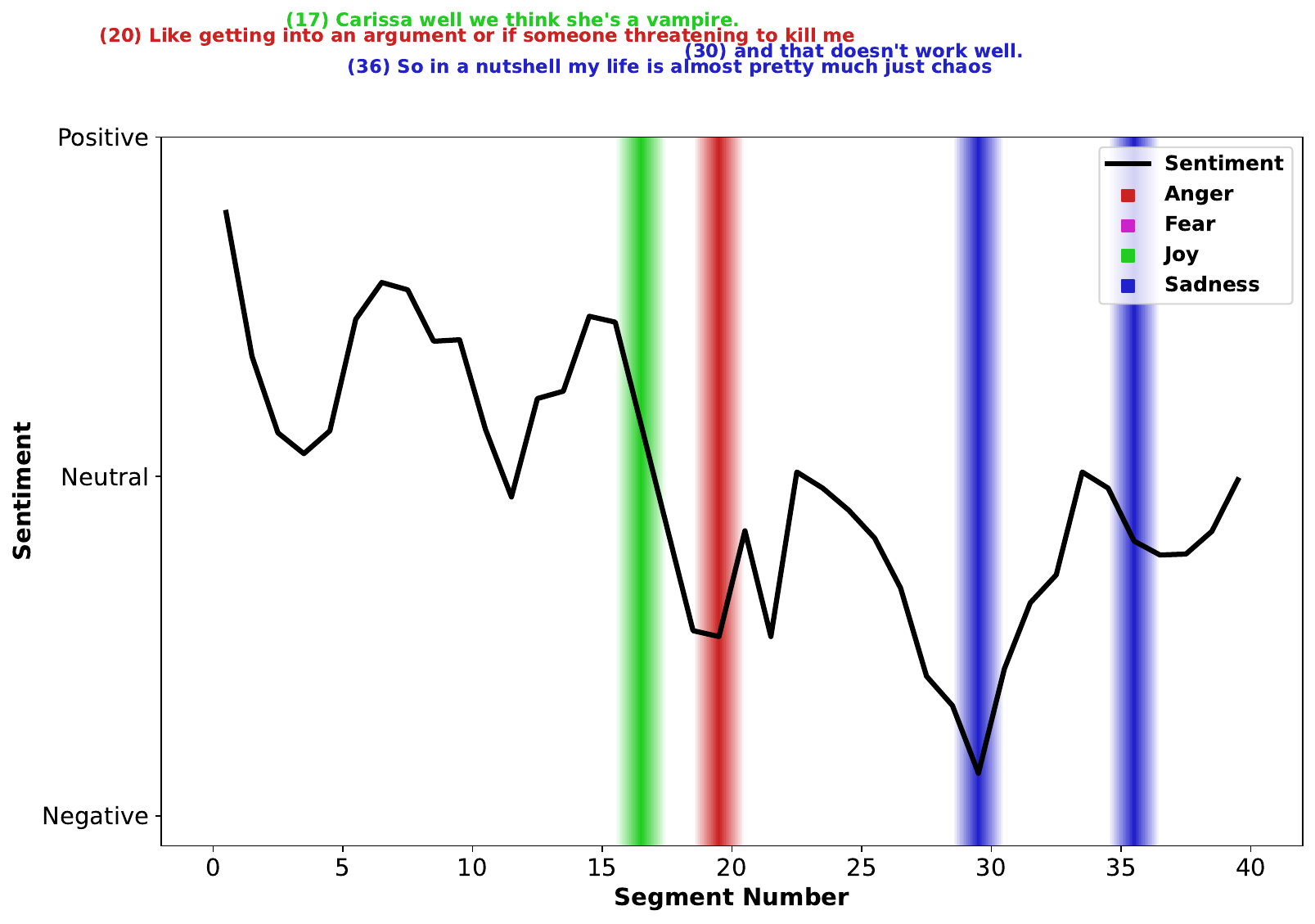}
            \caption{General interview, random sample 2}
    	\end{subfigure}
    }\\
    \makebox[\linewidth][c]{%
    	\begin{subfigure}[b]{0.5\columnwidth}
    		\centering
    		\includegraphics[width=\linewidth]{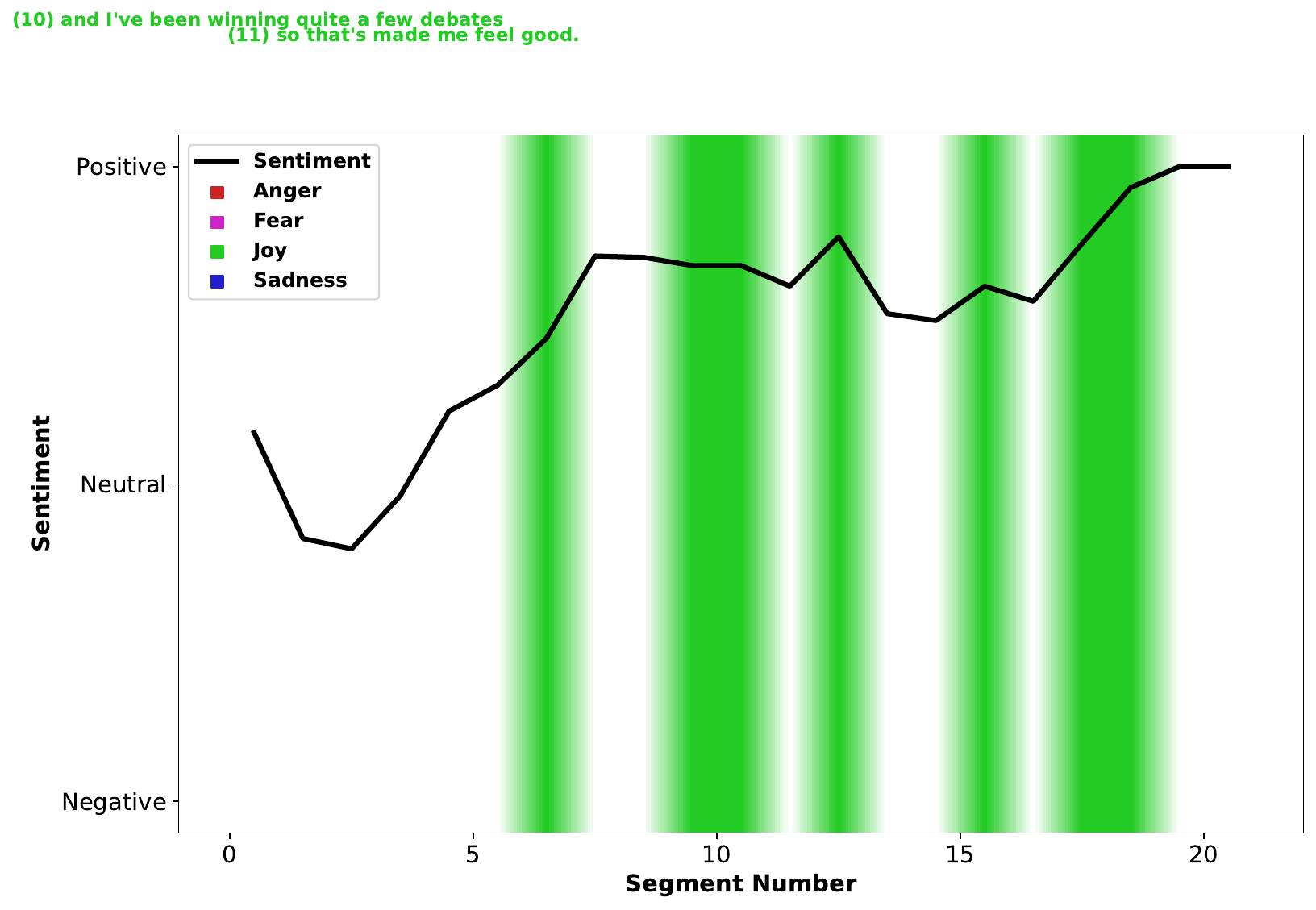}
            \caption{Positive experience interview, random sample 3}
    	\end{subfigure}
    	\begin{subfigure}[b]{0.5\columnwidth}
    		\centering
    		\includegraphics[width=\linewidth]{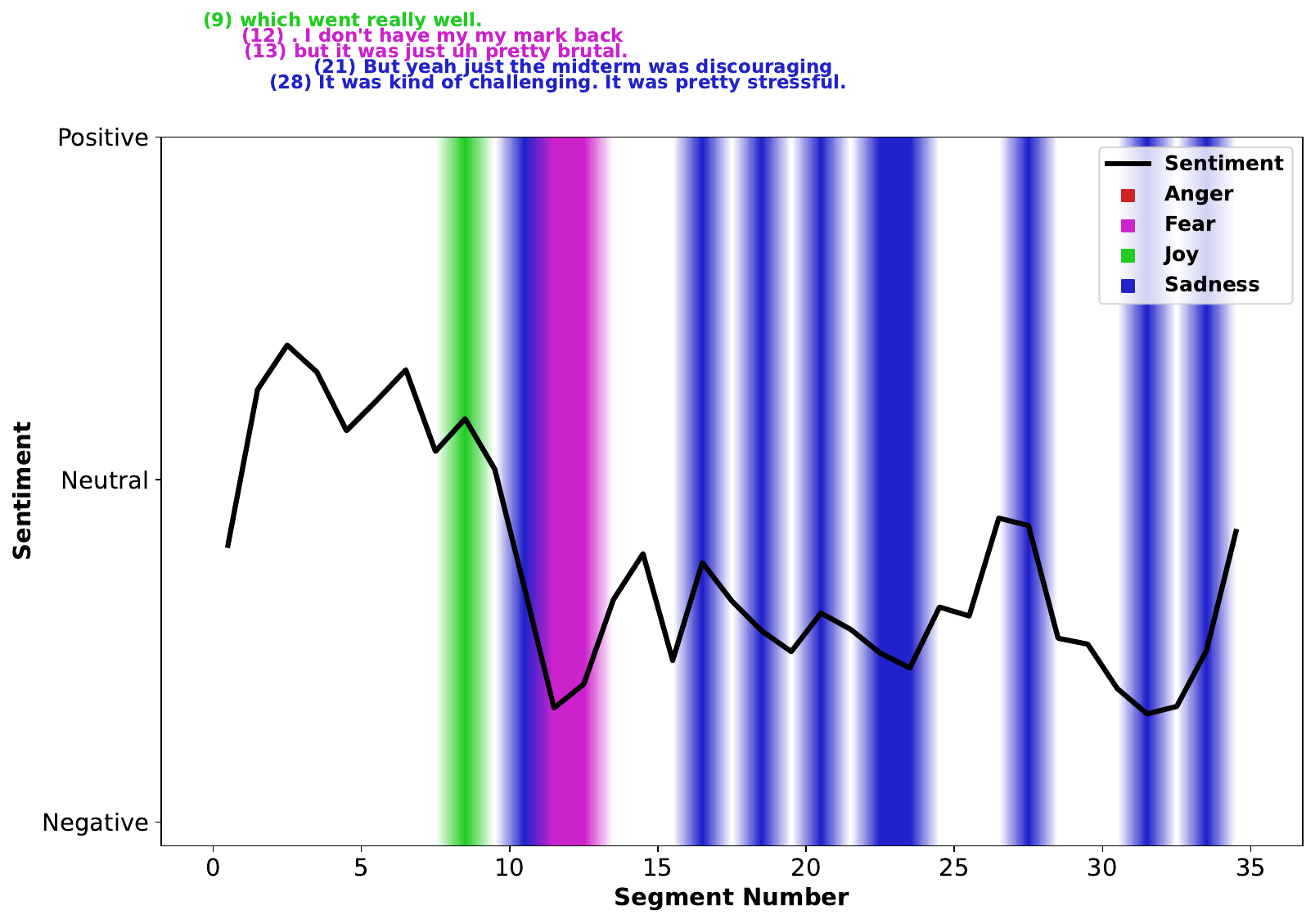}
            \caption{Negative experience interview, random sample 4}
    	\end{subfigure}
    }
    \caption{\label{fig_mood_offspring}Emotion and sentiment dynamics for representative offspring (child) interview samples. The top row shows segments from general interviews, while the bottom row shows segments from positive and negative experience interviews. The horizontal axis represents segmented time, and the vertical axis represents sentiment scores. The black curve shows sentiment trajectories, while colored vertical bands indicate discrete emotion labels. Segments are derived using content-aware partitioning, and representative text excerpts are displayed above each plot to illustrate alignment between raw speech, segmentation, and annotations.}
\end{figure*}

Figure \ref{fig_mood_offspring} illustrates the changes in mood, emotions, and sentiment in four random samples in offspring data. We can also see sample segments in those files shown at the top of the figures which gives an idea of how the segmentation and the labeling is done. The first two figures on top are random samples from the three minutes general interviews, and the bottom figures are random samples from a positive experience and a negative experience. As we can see in the positive experience interview, the dominant emotion is `joy' and in the negative experience, `sadness' is the dominant emotion. Often times `anger' is the dominant emotion in the negative experience interviews.

\begin{figure*}[!t]
	\centering
    \hspace{-1cm}
    \makebox[\linewidth][c]{
	   \includegraphics[width=0.97\textwidth]{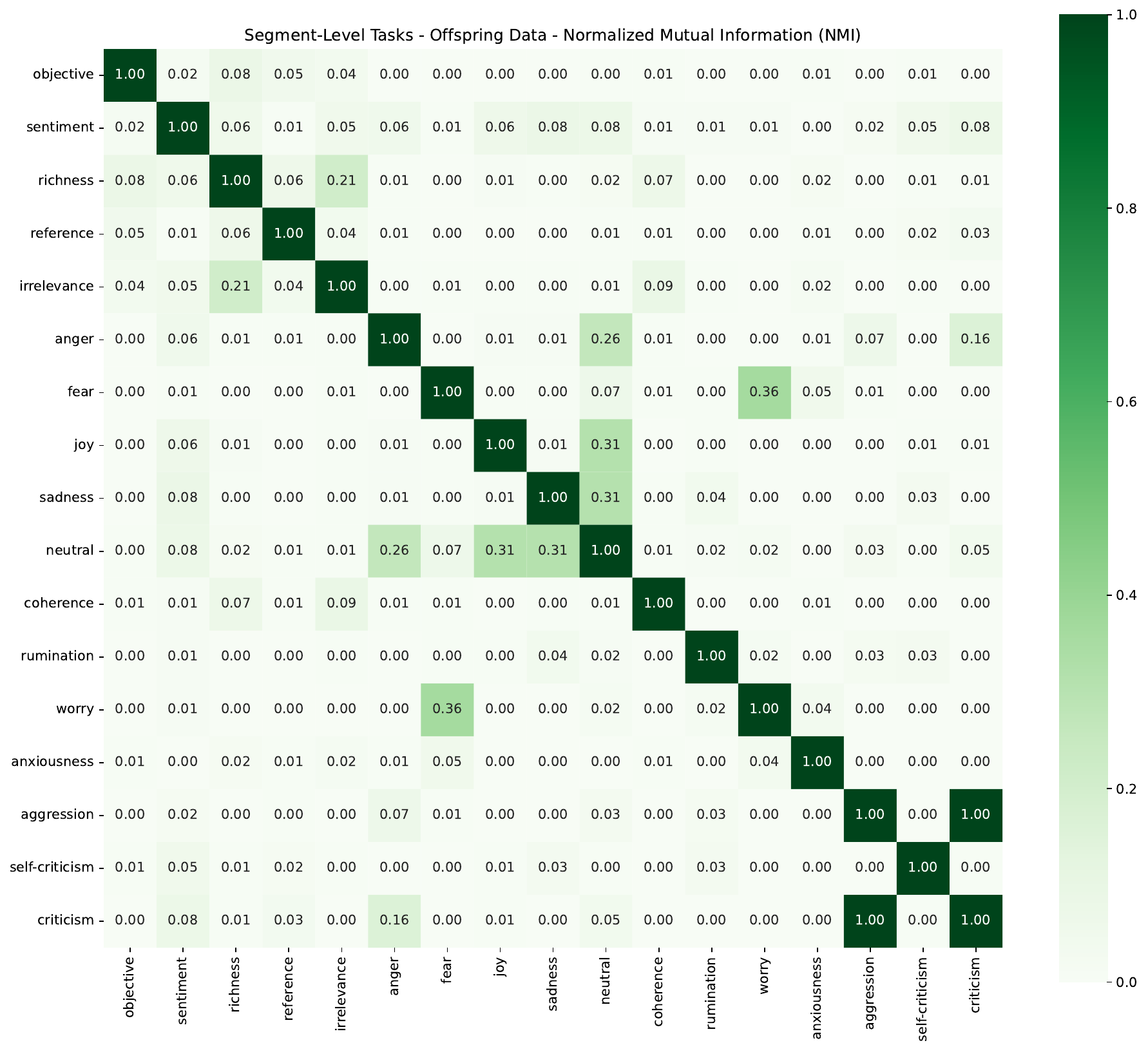}
    }
    \caption{Pairwise Normalized Mutual Information (NMI) between segment-level annotation tasks for offspring (child) interview data. Each cell represents the dependency between two tasks based on their label distributions. Higher values indicate stronger relationships between tasks (e.g., fear and worry), while lower values indicate weaker or independent labeling patterns. All values are computed using segment-level annotations across the offspring dataset.}
	\label{fig_dataset_nmi_segment_offspring}
\end{figure*}

\begin{table*}[!t]
\centering
\caption{Summary statistics of the offspring (child) dataset. The table reports the number of samples, number of classes, class distribution, and degree of imbalance for each task across segment-level, document-level, and cognitive/psychological categories. Segment-level tasks include additional annotations (e.g., self-criticism, aggression, anxiousness) specific to child interviews. Cognitive labels are based on clinical diagnoses and are available for a subset of subjects. Imbalance percentage is computed as $100 \times (\max_c n_c - \min_c n_c) / \max_c n_c$. Differences in sample counts reflect missing or filtered labels.}
\label{app:tbl_dataset_info_offspring}
\scalebox{0.85}{
\begin{tabular}{llcccccccc}
\toprule
\textbf{Level} & \textbf{Task} & \rot{\textbf{Num Samples}} & \rot{\textbf{Num Classes}} & \rot{\textbf{Imbalanced (\%)}} & \rot{\textbf{Label 0}} & \rot{\textbf{Label 1}} & \rot{\textbf{Label 2}} & \rot{\textbf{Label 3}} & \rot{\textbf{Label 4}} \\
\midrule
Segment & objective & 10319 & 2 & 1.90 & 5209 & 5110 & - & - & - \\
Segment & sentiment & 10222 & 3 & 52.51 & 2404 & 5062 & 2756 & - & - \\
Segment & richness & 10273 & 3 & 97.08 & 1366 & 8654 & 253 & - & - \\
Segment & reference & 10328 & 4 & 77.89 & 2072 & 4970 & 2187 & 1099 & - \\
Segment & irrelevance & 10301 & 2 & 87.98 & 9196 & 1105 & - & - & - \\
Segment & anger & 10329 & 2 & 95.55 & 9889 & 440 & - & - & - \\
Segment & fear & 10329 & 2 & 99.04 & 10231 & 98 & - & - & - \\
Segment & joy & 10329 & 2 & 94.49 & 9790 & 539 & - & - & - \\
Segment & sadness & 10329 & 2 & 94.62 & 9802 & 527 & - & - & - \\
Segment & neutral & 10328 & 2 & 80.92 & 1655 & 8673 & - & - & - \\
Segment & coherence & 10329 & 2 & 68.35 & 7846 & 2483 & - & - & - \\
Segment & rumination & 10329 & 2 & 99.01 & 10228 & 101 & - & - & - \\
Segment & worry & 10329 & 2 & 99.13 & 10240 & 89 & - & - & - \\
Segment & anxiousness & 10317 & 2 & 96.51 & 9969 & 348 & - & - & - \\
Segment & aggression & 2182 & 2 & 99.49 & 2171 & 11 & - & - & - \\
Segment & criticism & 8090 & 2 & 95.04 & 7708 & 382 & - & - & - \\
Segment & self-criticism & 10329 & 2 & 96.77 & 10006 & 323 & - & - & - \\
\midrule
Document & affect & 621 & 4 & 97.73 & 16 & 11 & 109 & 485 & - \\
Document & coherence & 616 & 5 & 95.83 & 11 & 33 & 109 & 264 & 199 \\
Document & richness & 614 & 3 & 96.70 & 82 & 515 & 17 & - & - \\
\midrule
Cognitive & spectrum & 85 & 4 & 86.11 & 36 & 27 & 17 & 5 & - \\
Cognitive & introvert & 85 & 2 & 39.62 & 53 & 32 & - & - & - \\
Cognitive & extrovert & 85 & 2 & 53.45 & 58 & 27 & - & - & - \\
Cognitive & ADHD & 85 & 2 & 53.45 & 58 & 27 & - & - & - \\
Cognitive & anxiety & 85 & 2 & 39.62 & 53 & 32 & - & - & - \\
Cognitive & depression & 85 & 2 & 83.56 & 73 & 12 & - & - & - \\
\bottomrule
\end{tabular}
}
\end{table*}

Figure \ref{fig_dataset_nmi_segment_offspring} illustrates the pairwise Normalized Mutual Information (NMI) between all the segment-level tasks on offspring data. As we can see there are some labels that are correlated and have a fairly high mutual information in the offspring data. For instance, `irrelevance' and `richness` have an NMI of 0.21 which makes sense as they are fairly opposite of each other. The two tasks of `fear' and `worry' also have an NMI of 0.36 which again makes sense as those tasks are intuitively correlated.

In terms of document-level tasks there are only three tasks of `affect', `coherence', and `richness' for the offspring data. These are labeled for each of the three interview files of children separately. In terms of psychological and cognitive labels we have the same labels as we saw in the parent data, as those are indeed the children's conditions labeled by psychiatrists. The information and statistics about the tasks on offspring data is shown in Table \ref{tbl_dataset_info_offspring}. Here again for certain tasks, some segments or documents were missing the corresponding label, or some labels were extremely imbalanced such there were only 1 or 2 examples from one category, and we removed those cases. This is the reason for the differences in the number of samples in Table \ref{tbl_dataset_info_offspring}. The imbalanced \% is calculated similarly to the parent data, as $100 * (\max_{c\in C}n_c - \min_{c\in C}n_c) / \max_{c\in C}n_c$. As we can see most of the tasks are quite imbalanced.

For predicting the psychological and cognitive labels, we concatenate the three audio files of each child and use the three interviews together to predict the final cognitive labels. In contrary, for segment-level and document-level we use them separately as they are labeled separately in the first place. We can see from Table \ref{tbl_dataset_info_offspring} that we only have 85 samples for the cognitive tasks where we have all 3 audio files, and we have the psychiatrist label. This is a low data setting and poses a lot of challenges.

\begin{table*}[h]
\centering
\caption{Inter-rater agreement statistics for segment-level annotation tasks in parent interview data. Each value is reported as mean $\pm$ standard deviation across subjects. Metrics include three forms of Intraclass Correlation Coefficient (ICC1: one-way random effects; ICC2: two-way random effects with absolute agreement; ICC3: two-way mixed effects with consistency), Krippendorff’s $\alpha$ for interval data, and average pairwise Pearson correlation. These metrics collectively quantify agreement reliability, accounting for both consistency and chance-corrected agreement across multiple annotators.}
\label{tbl_icc_parent}
\hspace*{-0.5cm}
\scalebox{0.8}{
\begin{tabular}{llllll}
\toprule
 & \textbf{ICC1} & \textbf{ICC2} & \textbf{ICC3} & \textbf{Krippendorff} & \textbf{Pearson} \\
 & & & & \textbf{Alpha} & \textbf{Correlation} \\
\textbf{Task} &  &  &  &  &  \\
\midrule
\textbf{Objective} & 0.9954 $\pm$ 0.0174 & 0.9954 $\pm$ 0.0174 & 0.9955 $\pm$ 0.0173 & 0.9954 $\pm$ 0.0176 & 0.9957 $\pm$ 0.0166 \\
\textbf{Sentiment} & 0.7019 $\pm$ 0.1725 & 0.7098 $\pm$ 0.1451 & 0.7229 $\pm$ 0.1281 & 0.6996 $\pm$ 0.1726 & 0.7332 $\pm$ 0.1256 \\
\textbf{Criticism} & 0.5115 $\pm$ 0.2537 & 0.523 $\pm$ 0.2409 & 0.5345 $\pm$ 0.2336 & 0.5093 $\pm$ 0.2536 & 0.5549 $\pm$ 0.2273 \\
\textbf{Worry} & 0.4726 $\pm$ 0.2814 & 0.4831 $\pm$ 0.2723 & 0.496 $\pm$ 0.2694 & 0.4706 $\pm$ 0.2811 & 0.5123 $\pm$ 0.2745 \\
\textbf{Anger} & 0.4684 $\pm$ 0.3185 & 0.4745 $\pm$ 0.3135 & 0.4798 $\pm$ 0.3116 & 0.4666 $\pm$ 0.3181 & 0.4933 $\pm$ 0.3132 \\
\textbf{Sadness} & 0.4609 $\pm$ 0.2678 & 0.4656 $\pm$ 0.2631 & 0.4693 $\pm$ 0.2623 & 0.459 $\pm$ 0.2674 & 0.4878 $\pm$ 0.2667 \\
\textbf{Joy} & 0.4152 $\pm$ 0.2778 & 0.4472 $\pm$ 0.2414 & 0.4713 $\pm$ 0.2311 & 0.4136 $\pm$ 0.2766 & 0.4878 $\pm$ 0.2251 \\
\textbf{Rumination} & 0.4331 $\pm$ 0.0946 & 0.4323 $\pm$ 0.0957 & 0.4306 $\pm$ 0.0981 & 0.4302 $\pm$ 0.0953 & 0.3579 $\pm$ 0.0347 \\
\textbf{Cohesion} & 0.3994 $\pm$ 0.2379 & 0.4101 $\pm$ 0.2272 & 0.4184 $\pm$ 0.2276 & 0.3973 $\pm$ 0.2367 & 0.4305 $\pm$ 0.2305 \\
\textbf{Fear} & 0.3742 $\pm$ 0.277 & 0.3839 $\pm$ 0.2727 & 0.3945 $\pm$ 0.2717 & 0.3723 $\pm$ 0.2764 & 0.4088 $\pm$ 0.28 \\
\textbf{Neutral} & 0.3022 $\pm$ 0.2549 & 0.3594 $\pm$ 0.1888 & 0.3978 $\pm$ 0.1783 & 0.3007 $\pm$ 0.253 & 0.399 $\pm$ 0.1787 \\
\textbf{Overinclusive} & 0.3463 $\pm$ 0.2901 & 0.3506 $\pm$ 0.2886 & 0.3543 $\pm$ 0.2884 & 0.3448 $\pm$ 0.2895 & 0.3554 $\pm$ 0.2998 \\
\bottomrule
\end{tabular}
}
\end{table*}

\begin{table*}[h]
\centering
\caption{Inter-rater agreement statistics for segment-level annotation tasks in offspring (children) interview data. Values are reported as mean $\pm$ standard deviation across subjects. Metrics include ICC1, ICC2, ICC3, Krippendorff’s $\alpha$, and average Pearson correlation. Compared to parent data, these values reflect annotation consistency under increased task difficulty and data variability in child speech.}
\label{tbl_icc_offspring}
\scalebox{0.8}{
\begin{tabular}{llllll}
\toprule
 & \textbf{ICC1} & \textbf{ICC2} & \textbf{ICC3} & \textbf{Krippendorff} & \textbf{Pearson} \\
 & & & & \textbf{Alpha} & \textbf{Correlation} \\
\textbf{Task} &  &  &  &  &  \\
\midrule
\textbf{Objective} & 0.9902 $\pm$ 0.0886 & 0.9913 $\pm$ 0.077 & 0.9928 $\pm$ 0.0626 & 0.9902 $\pm$ 0.0888 & 0.9916 $\pm$ 0.074 \\
\textbf{Sentiment} & 0.7104 $\pm$ 0.1999 & 0.7156 $\pm$ 0.1928 & 0.7297 $\pm$ 0.1836 & 0.7031 $\pm$ 0.2023 & 0.7391 $\pm$ 0.1827 \\
\textbf{Criticism} & 0.6758 $\pm$ 0.3417 & 0.69 $\pm$ 0.3213 & 0.7034 $\pm$ 0.3058 & 0.6723 $\pm$ 0.3432 & 0.7184 $\pm$ 0.2972 \\
\textbf{Fear} & 0.5188 $\pm$ 0.2678 & 0.5302 $\pm$ 0.2503 & 0.5404 $\pm$ 0.2377 & 0.5125 $\pm$ 0.2696 & 0.5132 $\pm$ 0.2775 \\
\textbf{Worry} & 0.5012 $\pm$ 0.3936 & 0.4972 $\pm$ 0.4029 & 0.5028 $\pm$ 0.3992 & 0.4975 $\pm$ 0.3928 & 0.4667 $\pm$ 0.4128 \\
\textbf{Joy} & 0.4898 $\pm$ 0.3245 & 0.4932 $\pm$ 0.3229 & 0.4992 $\pm$ 0.3216 & 0.4841 $\pm$ 0.3239 & 0.4952 $\pm$ 0.3331 \\
\textbf{Anger} & 0.4526 $\pm$ 0.3134 & 0.4592 $\pm$ 0.3104 & 0.4693 $\pm$ 0.3049 & 0.4469 $\pm$ 0.312 & 0.4379 $\pm$ 0.3263 \\
\textbf{Neutral} & 0.4354 $\pm$ 0.2873 & 0.4396 $\pm$ 0.2883 & 0.45 $\pm$ 0.2874 & 0.4287 $\pm$ 0.2851 & 0.4531 $\pm$ 0.2925 \\
\textbf{Sadness} & 0.4208 $\pm$ 0.2981 & 0.4172 $\pm$ 0.3095 & 0.425 $\pm$ 0.3064 & 0.4146 $\pm$ 0.2956 & 0.4179 $\pm$ 0.3195 \\
\textbf{Rumination} & 0.3169 $\pm$ 0.3033 & 0.3254 $\pm$ 0.3008 & 0.3353 $\pm$ 0.3005 & 0.3113 $\pm$ 0.301 & 0.2845 $\pm$ 0.2988 \\
\bottomrule
\end{tabular}
}
\end{table*}

\begin{figure*}[!t]
	\centering
    \includegraphics[width=0.7\textwidth]{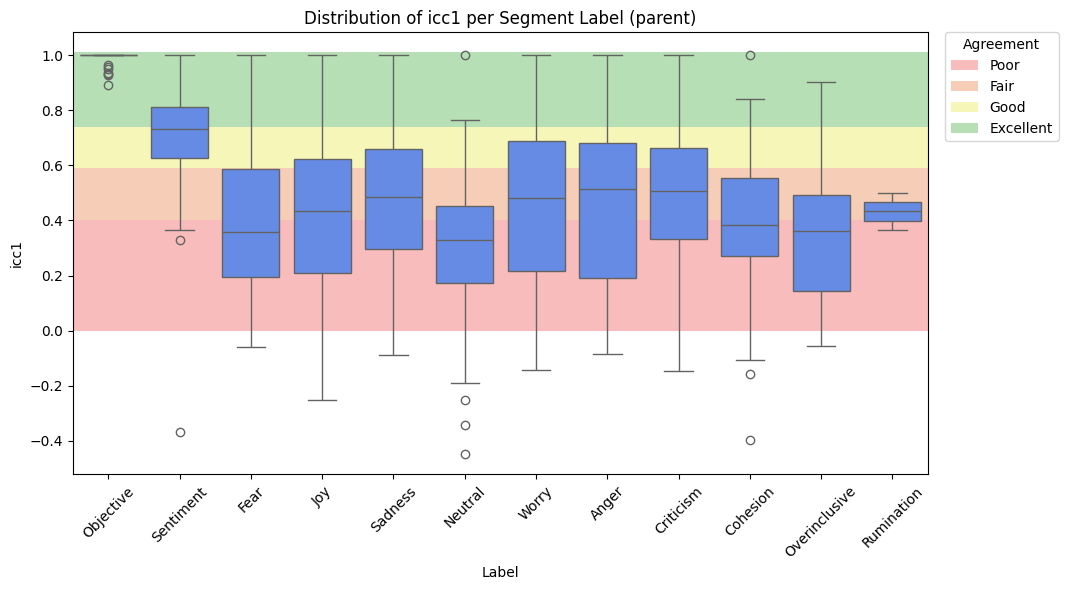}
    \caption{Distribution of Intraclass Correlation Coefficient (ICC1) values across subjects for each segment-level task in parent data. Each boxplot summarizes variability in inter-rater agreement across subjects, showing median, quartiles, and outliers. Higher ICC values indicate stronger agreement among raters, while wider distributions reflect variability in annotation consistency across subjects and tasks.}
	\label{fig_dataset_icc_parent}
\end{figure*}

\begin{figure*}[!t]
	\centering
    \includegraphics[width=0.7\textwidth]{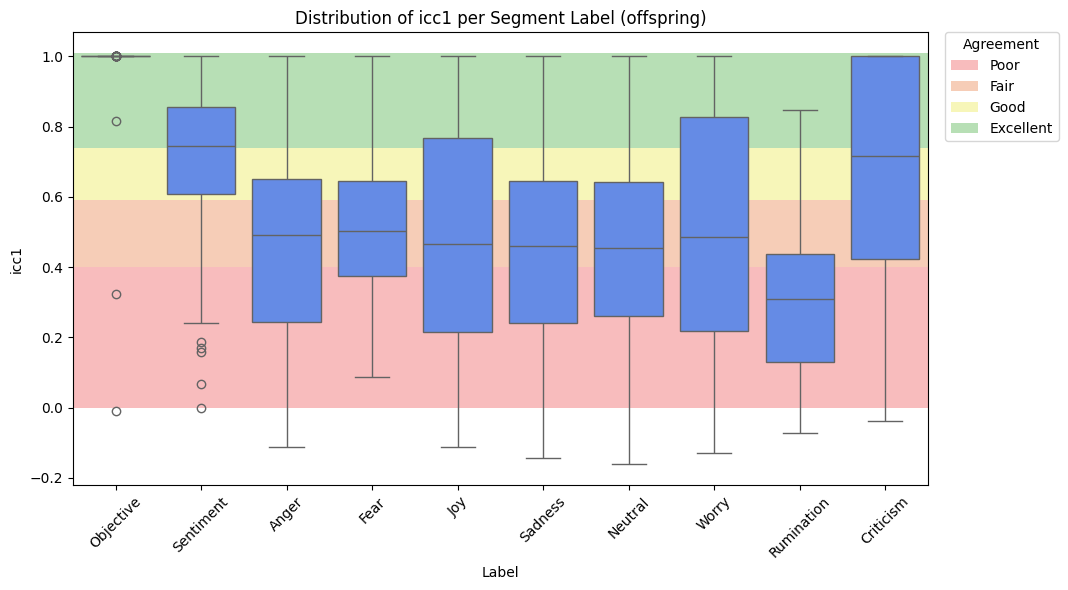}
    \caption{Distribution of Intraclass Correlation Coefficient (ICC1) values across subjects for each segment-level task in offspring data. Boxplots illustrate variability in agreement across annotators, highlighting differences in reliability across tasks. Lower ICC values and wider spreads indicate increased annotation uncertainty, particularly for subjective or low-frequency labels.}
	\label{fig_dataset_icc_offspring}
\end{figure*}

Inter-rater agreement at the segment level is evaluated on a per-subject basis by first constructing, for each subject and each task, a ratings matrix in which rows represent segments belonging to that subject and columns represent the raters who annotated those segments. Reliability is quantified using multiple complementary statistics. Three intraclass correlation coefficients, ICC(1), ICC(2), and ICC(3) are derived from an ANOVA decomposition of between-segment, between-rater, and residual variance components, corresponding respectively to a one-way random-effects model, a two-way random-effects model assessing absolute agreement, and a two-way mixed-effects model assessing consistency. Chance-corrected reliability is captured through Krippendorff’s $\alpha$ for interval data, computed by contrasting the observed disagreement among raters with the disagreement expected under statistical independence. In addition, average pairwise Pearson correlations are calculated across all rater pairs. This full set of agreement metrics is computed independently for each subject, producing one reliability estimate per metric per subject. Finally, to characterize the overall reliability of each task across the entire dataset, these subject-level values are aggregated by computing their mean and standard deviation, yielding the summary statistics reported in Table \ref{tbl_icc_parent} and Table \ref{tbl_icc_offspring} for parent and offspring data, respectively. Figures \ref{fig_dataset_icc_parent} and \ref{fig_dataset_icc_offspring} illustrate the distribution of ICC1 scores across subjects for the different tasks on parent and offspring data.

To contextualize these coefficients, ICC values below 0.40 are typically interpreted as indicating poor agreement, values between 0.40-0.59 as fair, 0.60-0.74 as good, and values of 0.75 or higher as excellent reliability \cite{muller1994critical_icc}. Under these qualitative bands, the Objective task exhibits excellent reliability for both parent and offspring speech, reflecting its relatively well-defined operationalization. In contrast, several affective labels fall into the fair and occasionally poor ranges, including Overinclusive, Cohesion, Rumination, Fear, Neutral, Joy, and Sadness in the parent data, and Rumination, Sadness, Neutral, and Anger in offspring annotations. These labels tend to capture nuanced, introspective, or context-dependent psychosocial constructs, which naturally yield greater rater variability. An additional contributing factor is label imbalance across subjects and tasks. When only a small fraction of segments are assigned a given label, variance across raters becomes dominated by a few sparse decisions, which inflates uncertainty and depresses reliability estimates such as ICC and Krippendorff's $\alpha$. This effect is particularly evident for rare categories like Overinclusive and Rumination. Consequently, findings derived from these low-frequency, lower-reliability labels should be interpreted with caution in subsequent modeling analyses, as downstream performance will inherently reflect the noisier ground truth associated with them.

\section{Societal Impacts}
\label{sec_societal_impact}

In terms of environmental impact, training large models is costly. In this work we used relatively small datasets and small network architecture, so that all the experiments were run on a high-end PC with Titan RTX GPU. However, if someone wants to train a larger version of SPECTRA on larger datasets it could become costly.

As a positive societal impact, this project has great positive impacts in helping psychiatrists in early diagnosis of mental disorders. The goal is to provide decision-support signals and tool that may complement clinical judgment, and not to replace clinicians and human judgment. We advocate for the responsible use of SPECTRA, particularly for potential clinical applications, ensuring consent, de-identification, and human-in-the-loop deployment. Our clinical speech analysis work was approved by the government's Research Ethics Board (file number: 100266).

\section{Source Code}
To facilitate reproducibility, the Python implementation of the proposed SPECTRA framework is included in the supplementary material
(spectra.py).



\clearpage
\section*{NeurIPS Paper Checklist}

\begin{enumerate}

\item {\bf Claims}
    \item[] Question: Do the main claims made in the abstract and introduction accurately reflect the paper's contributions and scope?
    \item[] Answer: \answerYes{}
    \item[] Justification: The claims are supported by the empirical results and analysis in both section \ref{sec_experiments} and appendix.

\item {\bf Limitations}
    \item[] Question: Does the paper discuss the limitations of the work performed by the authors?
    \item[] Answer: \answerYes{} 
    \item[] Justification: Monte Carlo sampling could be slow and the method generally addresses uncertainty rather than directly targeting accuracy.

\item {\bf Theory assumptions and proofs}
    \item[] Question: For each theoretical result, does the paper provide the full set of assumptions and a complete (and correct) proof?
    \item[] Answer: \answerNA{} 
    \item[] Justification: Our paper does not include theoretical claims.

    \item {\bf Experimental result reproducibility}
    \item[] Question: Does the paper fully disclose all the information needed to reproduce the main experimental results of the paper to the extent that it affects the main claims and/or conclusions of the paper (regardless of whether the code and data are provided or not)?
    \item[] Answer: \answerYes{} 
    \item[] Justification: We have used a combination of public and private datasets. The details of datasets and experiments including hyperparameters are provided in \ref{sec_experiments}. And the code is provided in supplementary material. But we are not allowed to publicly share the mental health data.

\item {\bf Open access to data and code}
    \item[] Question: Does the paper provide open access to the data and code, with sufficient instructions to faithfully reproduce the main experimental results, as described in supplemental material?
    \item[] Answer: \answerYes{} 
    \item[] Justification: Other than the mental health dataset, the other datasets and the source code are provided. 

\item {\bf Experimental setting/details}
    \item[] Question: Does the paper specify all the training and test details (e.g., data splits, hyperparameters, how they were chosen, type of optimizer) necessary to understand the results?
    \item[] Answer: \answerYes{} 
    \item[] Justification: All the training details and experimental setup is provided at \ref{sec_experiments}.

\item {\bf Experiment statistical significance}
    \item[] Question: Does the paper report error bars suitably and correctly defined or other appropriate information about the statistical significance of the experiments?
    \item[] Answer: \answerYes{} 
    \item[] Justification: We have provided the standard deviations of errors for the results on public dataset. However, for the mental health data experiments, we did not include error bars as the data is relatively small.

\item {\bf Experiments compute resources}
    \item[] Question: For each experiment, does the paper provide sufficient information on the computer resources (type of compute workers, memory, time of execution) needed to reproduce the experiments?
    \item[] Answer: \answerYes{} 
    \item[] Justification: We have provided the GPU type and memory information in section \ref{sec_experiments}, however we have not measured the runtime for each of the experiments.
    
\item {\bf Code of ethics}
    \item[] Question: Does the research conducted in the paper conform, in every respect, with the NeurIPS Code of Ethics \url{https://neurips.cc/public/EthicsGuidelines}?
    \item[] Answer: \answerYes{} 
    \item[] Justification: This research conforms with code of ethics and does not have harmful impacts. The societal impacts are discussed in \ref{sec_societal_impact}. 

\item {\bf Broader impacts}
    \item[] Question: Does the paper discuss both potential positive societal impacts and negative societal impacts of the work performed?
    \item[] Answer: \answerYes{} 
    \item[] Justification: We have discussed positive and negative societal impacts in section \ref{sec_societal_impact}. 
    
\item {\bf Safeguards}
    \item[] Question: Does the paper describe safeguards that have been put in place for responsible release of data or models that have a high risk for misuse (e.g., pre-trained language models, image generators, or scraped datasets)?
    \item[] Answer: \answerNA{} 
    \item[] Justification: Our paper poses no such risks.

\item {\bf Licenses for existing assets}
    \item[] Question: Are the creators or original owners of assets (e.g., code, data, models), used in the paper, properly credited and are the license and terms of use explicitly mentioned and properly respected?
    \item[] Answer: \answerYes{} 
    \item[] Justification: We have provided citations for the datasets and models we compared with. And we are the original owner of our work.

\item {\bf New assets}
    \item[] Question: Are new assets introduced in the paper well documented and is the documentation provided alongside the assets?
    \item[] Answer: \answerYes{} 
    \item[] Justification: We have provided complete information on the mental health dataset in section \ref{sec_supp_data}

\item {\bf Crowdsourcing and research with human subjects}
    \item[] Question: For crowdsourcing experiments and research with human subjects, does the paper include the full text of instructions given to participants and screenshots, if applicable, as well as details about compensation (if any)? 
    \item[] Answer: \answerNA{} 
    \item[] Justification: Our research involves human subjects. However, we have not collected the dataset ourselves, we have acquired that through collaboration with Psychiatry department and they have followed all the rules and requirements for compensation. 

\item {\bf Institutional review board (IRB) approvals or equivalent for research with human subjects}
    \item[] Question: Does the paper describe potential risks incurred by study participants, whether such risks were disclosed to the subjects, and whether Institutional Review Board (IRB) approvals (or an equivalent approval/review based on the requirements of your country or institution) were obtained?
    \item[] Answer: \answerYes{} 
    \item[] Justification: The paper does not pose such risks. And the dataset was collected with ethics board approvals as explained in section \ref{sec_societal_impact}.

\item {\bf Declaration of LLM usage}
    \item[] Question: Does the paper describe the usage of LLMs if it is an important, original, or non-standard component of the core methods in this research? Note that if the LLM is used only for writing, editing, or formatting purposes and does \emph{not} impact the core methodology, scientific rigor, or originality of the research, declaration is not required.
    \item[] Answer: \answerNA{} 
    \item[] Justification: LLMs were used for final edits and grammar checks on the manuscript.

\end{enumerate}

\end{document}